\documentclass[final,1p,times]{elsarticle}
\usepackage[utf8]{inputenc}
\usepackage{subcaption}
\usepackage{tabu}
\usepackage{graphbox}
\usepackage[numbers]{natbib}
\usepackage{amsmath}

\begin{document}

\begin{frontmatter}

\title{Explainable Artificial Intelligence: a Systematic Review}

\author{Giulia Vilone}
\author{Luca Longo}

\address{School of Computer Science, College of Science and Health,\\ Technological University Dublin, Dublin, Republic of Ireland}

\begin{abstract}
Explainable Artificial Intelligence (XAI) has experienced a significant growth over the last few years. This is due to the widespread application of machine learning, particularly deep learning, that has led to the development of highly accurate models but lack explainability and interpretability. A plethora of methods to tackle this problem have been proposed, developed and tested. 
This systematic review contributes to the body of knowledge by clustering these methods with a hierarchical classification system with four main clusters: review articles, theories and notions, methods and their evaluation. It also summarises the state-of-the-art in XAI and recommends future research directions.

\end{abstract}


\end{frontmatter}

\section{Introduction}
The number of scientific articles, conferences and symposia around the world in eXplainable Artificial Intelligence (XAI) has significantly increased over the last decade \cite{adadi2018peeking,preece2018asking}. This has led to the development of a plethora of domain-dependent and context-specific methods for dealing with the interpretation of machine learning (ML) models and the formation of explanations for humans. Unfortunately, this trend is far from being over, with an abundance of knowledge in the field which is scattered and needs organisation.
The goal of this article is to systematically review research works in the field of XAI and to try to define some boundaries in the field.
From several hundreds of research articles focused on the concept of explainability, about 350 have been considered for review by using the following search methodology. In a first phase, Google Scholar was queried to find papers related to ``explainable artificial intelligence'', ``explainable machine learning'' and ``interpretable machine learning''. Subsequently, the bibliographic section of these articles was thoroughly examined to retrieve further relevant scientific studies. The first noticeable thing, as shown in figure \ref{fig:tree_root} (a), is the distribution of the publication dates of selected research articles: sporadic in the 70s and 80s, receiving preliminary attention in the 90s, showing raising interest in 2000 and becoming a recognised body of knowledge after 2010. The first research concerned the development of an explanation-based system and its integration in a computer program designed to help doctors make diagnoses \cite{shortliffe1975computer}. Some of the more recent papers focus on work devoted to the clustering of methods for explainability, motivating the need for organising the XAI literature \cite{van2013research,dodge2018should,vellido2012making}. 
The upturn in the XAI research outputs of the last decade is prominently due to the fast increase in the popularity of ML and in particular of deep learning (DL), with many applications in several business areas, spanning from e-commerce \cite{wang2007recommendation} to games \cite{lapuschkin2019unmasking} and including applications in criminal justice \cite{rudin2014algorithms,rudin2019stop}, healthcare \cite{fellous2019explainable}, computer vision \cite{rudin2019stop} and battlefield simulations \cite{fox2017explainable}, just to mention a few. Unfortunately, most of the models that have been built with ML and deep learning have been labelled `black-box' by scholars because their underlying structures are complex, non-linear and extremely difficult to be interpreted and explained to laypeople. This opacity has created the need for XAI architectures that is motivated mainly by three reasons, as suggested by \cite{fox2017explainable,dovsilovic2018explainable}: i) the demand to produce more transparent models; ii) the need of techniques that enable humans to interact with them; iii) the requirement of trustworthiness of their inferences. 
Additionally, as proposed by many scholars \cite{dovsilovic2018explainable,thelisson2017regulatory} \cite{thelisson2017towards,wachter2017transparent}, models induced from data must be liable as liability will likely soon become a legal requirement. Article 22 of the General Data Protection Regulation (GDPR)
sets out the rights and obligations of the use of automated decision making. Noticeably, it introduces the \textit{right of explanation} by giving individuals the right to obtain an explanation of the inference/s automatically produced by a model, confront and challenge an associated recommendation, particularly when it might negatively affect an individual legally, financially, mentally or physically. By approving this GDPR article, the European Parliament attempted to tackle the problem related to the propagation of potentially biased inferences to society, that a computational model might have learnt from biased and unbalanced data.\\

\begin{figure}[htbp]
    \centering
    \includegraphics[scale=0.50]{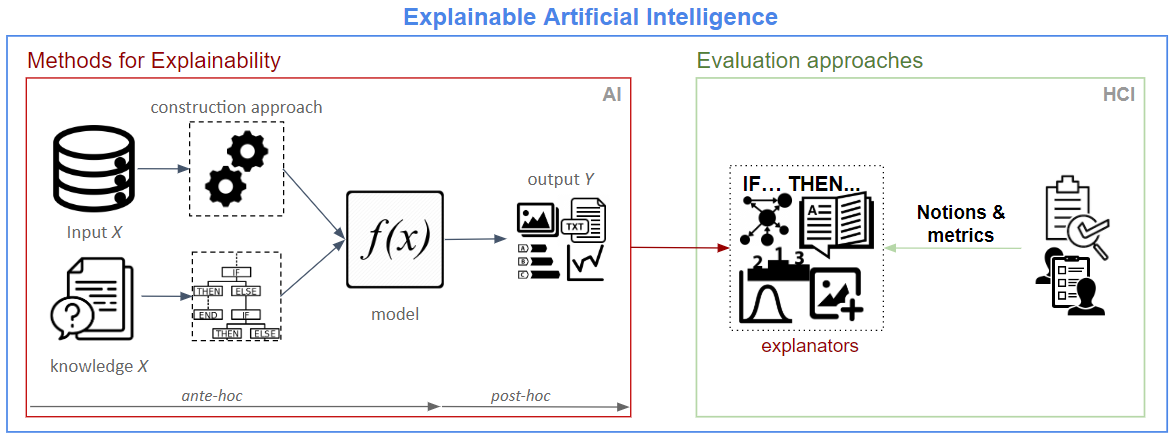}
    \caption{Diagrammatic view of Explainable Artificial Intelligence as a sub-field at the intersection of  Artificial Intelligence and Human-Computer Interaction}
    \label{fig:XAI}
\end{figure}

Many authors surveyed scientific articles surrounding explainability within Artificial Intelligence (AI) in specific sub-domains, motivating the need for literature organisation. For instance, \cite{samek2019towards,lacave2002review} respectively reviewed the methods for explanations with neural and bayesian networks while \cite{martens2007comprehensible} clustered the scientific contributions devoted to extracting rules from models trained with Support Vector Machines (SVMs). The goal was, and in general is, to create rules highly interpretable by humans while maintaining a degree of accuracy offered by trained models. \cite{choo2018visual} carried out a literature review of all the methods focused on the production of visual representations of the inferential process of deep learning techniques, such as heat-maps. Only a few scholars attempted to make a more comprehensive survey and organization of the methods for explainability as a whole \cite{adadi2018peeking,guidotti2018survey}.  
This paper builds on these efforts to organise the vast knowledge surrounding explanations and XAI as a discipline, and it aims at defining a classification system of a larger scope.
The conceptual framework at the basis of the proposed system is represented in Figure \ref{fig:XAI}. Most of the methods for explainability focus on interpreting and making the entire process of building an AI system transparent, from the inputs to the outputs via the application of a learning approach to generate a model. The outcome of these methods are explanations that can be of different formats, such as rules, numerical, textual or visual information, or a combination of the former ones. These explanations can be theoretically evaluated according to a set of notions that can be formalised as metrics, usually borrowed from the discipline of Human-Computer Interaction (HCI) \cite{miller2017explanation}.\\

The remainder of this paper is organised as it follows. Section \ref{research_met} provides a detailed description of the research methods employed for searching for relevant research articles. Section \ref{classification} proposes a classification structure of XAI describing top branches while Sections \ref{notions}-\ref{xaisurveys} expand this structure. Eventually, section \ref{conclusions} concludes this systematic review by trying to define the boundaries of the discipline of XAI, as well as suggesting future research work and challenges.\\

\section{Research methods}\label{research_met}
Organizing the literature of explainability within AI in a precise and indisputable way as well as setting clear boundaries is far from being an easy task. This is due to the multidisciplinarity surroundings of this new fascinating field of research spanning from Computer Science to Mathematics, from Psychology to Human Factors, from Philosophy to Ethics. The development of computational models from data belongs mainly to Computer Science, Statistics and Mathematics, whereas the study of explainability belongs more to Human Factors and Psychology since humans are involved. Reasoning over the notion of explainability touches Ethics and Philosophy. Therefore, some constraints had to be set, and the following publication types were excluded:
\begin{itemize}
\item scientific studies discussing the notion of explainability in different contexts than AI and Computer Science, such as Philosophy or Psychology;
\item articles or technical reports that have not gone through a peer-review process;
\item methods that could be employed for enhancing the explainability of AI techniques but that were not designed specifically for this purposes. For example, the scientific literature contains a considerable amount of articles related to methods designed for improving data visualization or feature selection. These methods can indeed help researchers to gain deeper insights into computational models, but they were not specifically designed for producing explanations. In other words, those methods developed only for enhancing model transparency but not directly focused on explanation were discarded.
\end{itemize}

\noindent Taking into account the above constraints, this systematic review was carried out in two phases:
\begin{enumerate}
\item papers discussing explainability were searched by using Google Scholar and the following terms: 
\emph{`explainable artificial intelligence'},
\emph{`explainable machine learning'}, 
\emph{`interpretable machine learning'}. 
The queries returned several thousands of results, but it became immediately clear that only the first ten pages could contain relevant articles. Altogether, these searches provided a basis of almost two hundred peer-reviewed publications;
\item the bibliographic section of the articles found in phase one was checked thoroughly. This led to the selection of one hundred articles whose bibliographic section was recursively analysed. This process was iterated until it converged and no more  articles were found.
\end{enumerate}

\section{Classification of scientific articles on explainability}\label{classification}
\begin{figure}[!ht]
\begin{minipage}{\textwidth}
\centering
  \subcaptionbox{}
    {\includegraphics[scale=0.27, align=c]{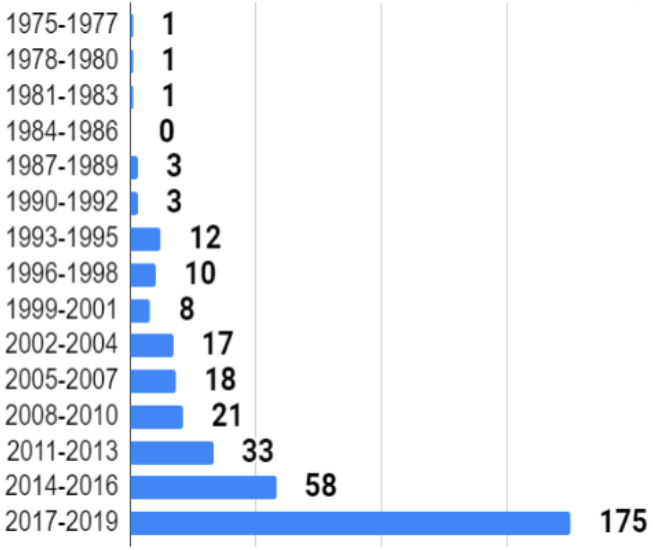}}
  \subcaptionbox{}
    {\includegraphics[scale=0.34, align=c]{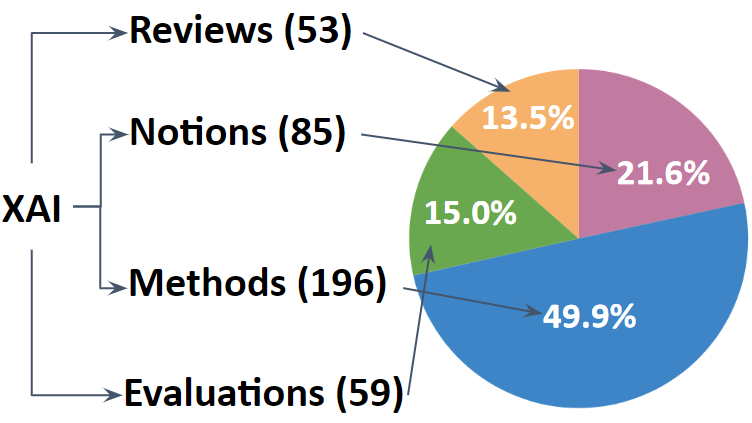}}
  \subcaptionbox{}
    {\includegraphics[scale=0.34, align=c]{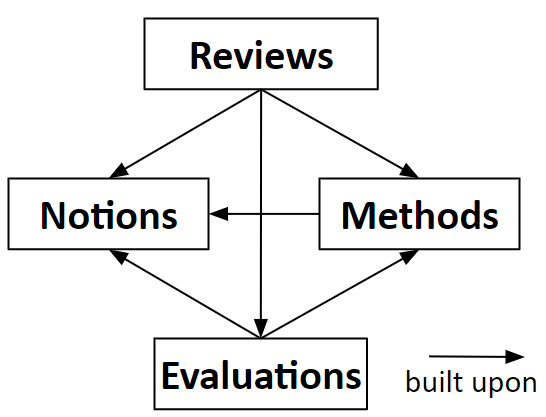}}
  \caption{Proposed classification of the XAI literature with (a) the distribution  of published  scientific articles over time, (b) the root of our hierarchical classification system representing the main four categories and the percentage of articles in each , and (c) the salient relations between these categories that have emerged.}
  \label{fig:tree_root}
\end{minipage}
\end{figure}

After a thorough analysis of all the selected articles, four main categories were extracted as depicted in Fig. \ref{fig:tree_root} and as listed below:
\begin{itemize}
\item {\verb|reviews on methods for explainability|} - it includes either literature or systematic reviews of those methods devoted to the proposal and/or testing of solutions for the explainability of data- and knowledge-driven models;
\item {\verb|notions related to the concept of explainability|} - it includes studies focused on the definition of those notions related to the concept of explainability and on the determination of the main characteristics as well as the requirements of an effective explanation; 
\item {\verb|development of new methods for explainability|} - it includes articles that propose novel and original methods for enhancing the explainability of data/knowledge-driven models;
\item {\verb|evaluation of methods for explainability|} - it includes articles reporting the results of scientific studies aiming at evaluating the performance of different methods for explainability.
\end{itemize}

Following the proposed classification, it was possible to design a map of the XAI literature in form of a tree whose root contains the above four categories (figure \ref{fig:tree_root}, part b). This tree expands into branches of different depth where leaves represent scientific articles. Figure \ref{fig:tree_root}, part b, also shows the percentage of articles grouped by each category, clearly highlighting the distribution of the research efforts towards the development of methods for explainability. Note that, a paper might appear in multiple branches of this classification, as it might cover multiple dimensions.
Figure \ref{fig:tree_root}, part c, depicts the dependencies of the main four categories. In general, scholars would not be able to carry out reviews of the XAI literature without the existence and consideration of relevant notions and methods for explainability as well as the approaches for evaluating the performances of these methods. Evaluation approaches naturally followed the creation of methods for explainability which have been engineered to meet as many requirements of an effective explanation as possible. 

\section{Reviews of the XAI literature}\label{xaisurveys}
\begin{figure}[!ht]
  \includegraphics[scale=0.35]{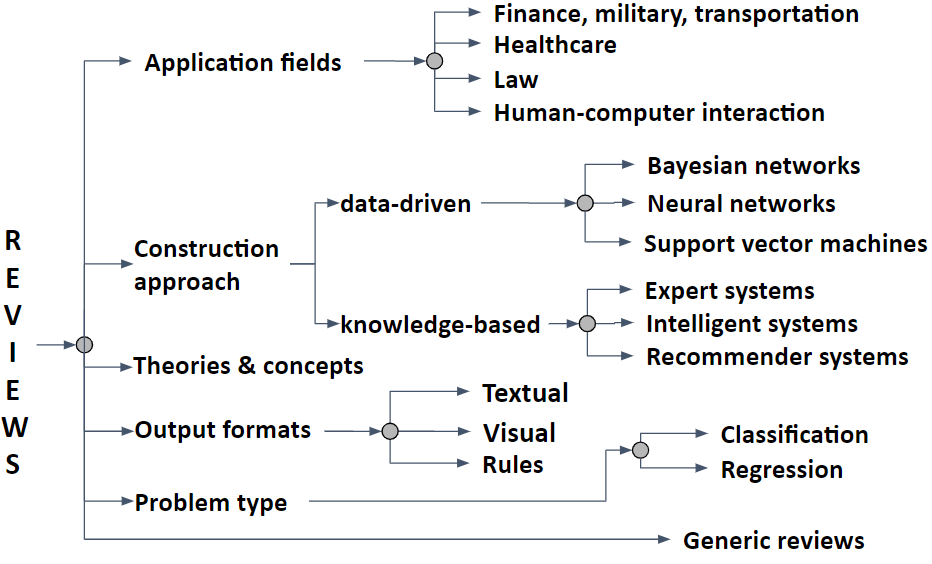}
  \includegraphics[scale=0.36]{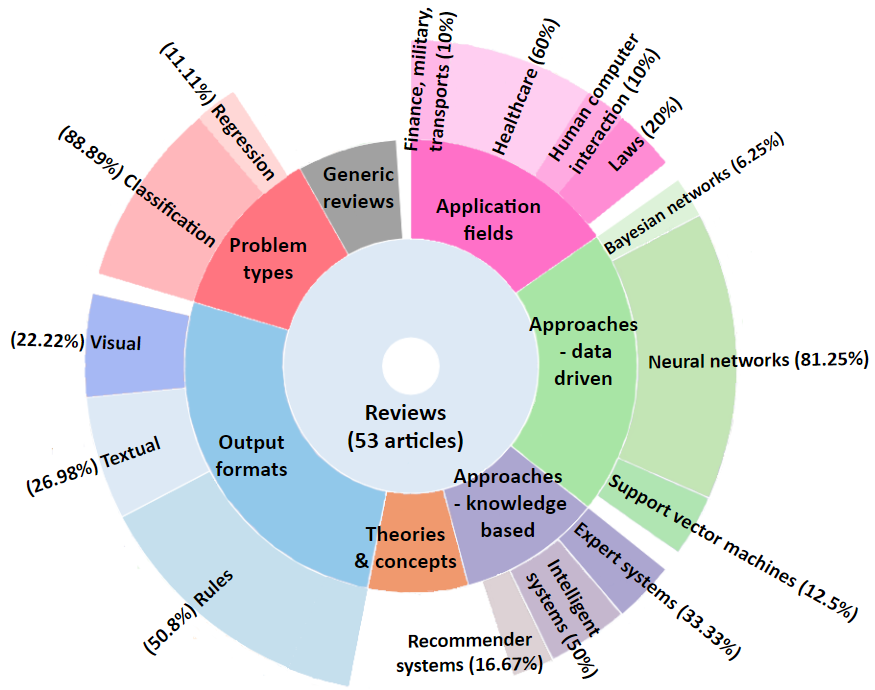}
  \caption{Hierarchical classification of the review articles on explainable artificial intelligence and machine learning interpretability (left) and distribution of the review articles across categories (right).}
  \label{fig:reviews_tree_chart}
\end{figure}

This category contains literature and systematic reviews devoted to specific classes of solutions for explainability, such as systems generating textual explanations \cite{miller2017explainable}, or constrained to specific AI techniques as, for instance, neural networks \cite{jacobsson2005rule} (summary in table \ref{tab:surveys} and figure \ref{fig:reviews_tree_chart}). 
These reviews provide an entry point for researchers to acquire information and get familiar with the key aspects of the rapidly growing body of research related to explainability. They also attempt to summarise the main techniques for explainability and to highlight their strengths and limitations. 
Seven clusters emerged based on distinct aspects of explainability covered by these reviews: 

\begin{itemize}
\item {\verb|application fields|} - reviews on methods for explainability in a specific field of application;
\item {\verb|construction approaches|} - reviews on methods for explainability specifically designed to explain the inferential process of models. This category has been further divided into:
\begin{itemize}
\item {\verb|data-driven|} approaches which focus on extracting new knowledge from trained models from data, but without accounting for the prior knowledge of domain experts. 
\item {\verb|knowledge-driven|} approaches focused on capturing an expert's knowledge and logic, often embedded in the notion of agent;
\end{itemize}
\item {\verb|theories & concepts|} - reviews of the notions related to the concept of explainability; 
\item {\verb|output formats|} - reviews on methods for explainability focused on generating specific formats of explanations, such as visual or rules; 
\item {\verb|problem types|} - review articles  on methods designed to explain the logic of data and knowledge-driven models applied to a specific type of problem, namely regression or classification;
\item {\verb|generic reviews|} - generic reviews that cover a wide range of data/knowledge-driven models as well as their methods for explainability and  cannot be placed within any other category. 
\end{itemize}

In the application fields cluster, the assumption of the methods for explainability is that it is not possible to accept the inference made by a model without understanding its functioning because a decision, supported by a wrong prediction, can have a dramatic impact on people's lives \cite{adadi2018peeking}.
The second cluster, construction data-driven approaches, contains reviews of methods for explainability for specific data-driven learning approaches, mainly neural networks \cite{arras2019evaluating,choo2018visual,dam2018explainable,hoffman2018explaining,mathews2019explainable,montavon2017methods,samek2019towards,zhang2018visual}, bayesian networks \cite{lacave2002review} and SVMs \cite{backhaus2014classification,martens2007comprehensible}, not constrained to a specific type of input data for the approach or a particular output format for an explanation, such as images or texts.
Other scholars instead focused on reviewing methods for knowledge-based approaches such as Expert Systems (ES) \cite{swartout1993explanation} and Intelligent Systems \cite{gregor1999explanations}.
In particular, these surveys analysed what types and formats of explanations were tested on these systems and which ones work better than others. For instance, \cite{papadimitriou2012generalized} showed that rich explanations, based on a combination of information regarding users, items and features, are very effective, while \cite{gregor1999explanations} claimed that explanations should be context-specific to be effective.
The third cluster contains those reviews focused on objectively defining the concept of explainability and its set of related notions, which are discussed in depth in section \ref{xaiattributes}.
One of these studies presented an overview of different theories of explanation borrowed from the cognitive science and philosophy disciplines, contextualised within case-based reasoning \cite{sormo2005explanation}. In details, it is believed that, in order to be effective, an AI system should: (I) explain how it reached the answer and (II) why it is a good answer, (III) why a question is relevant or not, (IV) clarify the meaning of the terms used in the system that might not be understood by the users and, lastly, (V) teach the user about the domain. In short, the goals that an explanation must achieve depend on the domain under consideration, the underlying model and end-users.
Similarly, \cite{miller2017explainable} suggested that explanations should take into account the preferences and preconceptions of end-users. This can be achieved by incorporating more findings from the behavioural and social sciences into the newly emerging field of XAI. For example, people explain their behaviour based on their beliefs, desires and intentions hence these elements must be considered in an explanation. 
Eventually, explanations based on counterfactual examples should help end-users to understand the logic of an underlying model by leveraging on people's capability to infer general rules from a few examples. Counterfactuals add also something new to what is already known from the existing data and provide additional information on how a model behaves in novel, unseen situations \cite{byrne2019counterfactuals}.
The fourth cluster contains reviews of methods for explainability generating a specific output format for an explanation (further discussed in section \ref{newmethods}). 
Methods generating textual explanations are surveyed in \cite{swartout1993explanation} and compared according to some requirements about the structure and content of the explanations to adapt them to the users' needs and knowledge.
\cite{mencar2018paving} focused on written explanations generated from fuzzy rules integrated with natural language generation tools. The underlying reasonable assumption is that the understandability of these rules cannot be given for granted. Researchers studied the capabilities of `data-to-text' approaches that automatically create linguistic descriptions from a complex dataset by means of aggregation functions, implemented as fuzzy rules, that aggregate `computational perceptions'. A computational perception is ``a unit of meaning for the phenomenon under analysis and is identified by a set of linguistic expressions and their corresponding validity values given a situation.'' Some methods combine Logical AI and Statistical AI to generate textual explanations \cite{belle2017logic}. The former is concerned with `formal languages' to represent and reason with qualitative specifications, while the latter is focused on learning quantitative specifications from data. However, the authors claimed that the search for an effective way to learn representations of the inferential process of data-driven models is still open \cite{belle2017logic}.
A body of literature focused on the visual explanation of deep learning models. Explanators generating salient masks were investigated in \cite{choo2018visual, zhang2018visual} whilst \cite{choo2018visual, craven1992visualizing} reviewed methods that graphically represent the inner structure and functioning of neural networks with flow-charts or other explanatory graphs. An interesting alternative was proposed in \cite{lisboa2013interpretability} whereby methods based on nomograms, rule induction, fuzzy logic, graphical models and topographic mapping can be utilised to explain data-driven models and learning techniques. Similarly to textual explanation, the problem of visually inspecting data-driven models has not been resolved and there are still challenges and open questions to be answered.
Some reviews summarised the methods for explainability that generate sets of rules from underlying trained models \cite{hailesilassie2016rule} by extracting frequent relations from a dataset using fuzzy logic and fuzzy rules \cite{fernandez2019evolutionary, guillaume2001designing, lisboa2013interpretability}, the integration of symbolic logic with the neural networks \cite{besold2015towards, garcez2015neural} and, more generally, the usage of automated reasoning to shed a light over the inferential process of automatically constructed data-driven models \cite{bonacina2017automated}.
The fifth cluster contains reviews that analysed the methods for explainability for either regression \cite{otte2013safe} or classification \cite{backhaus2014classification, freitas2010importance,martens2011performance} problems. They have a broader scope than the previous reviews as they range over several fields, AI techniques and explanation types. Their goal was to summarise the important issues, still unresolved, of interpreting prediction models for both problem types and encouraging researchers to improve the existing or discover novel methods for explainability.
Eventually, some reviews have a more generic scope. They are aimed at proposing a comprehensive way of organizing the several methods for explainability \cite{adadi2018peeking, biran2017explanation, guidotti2018survey} or describing them \cite{biran2017explanation, gade2019explainable, xu2019explainable}. A group of these reviews tried to evaluate the performances of various methods. This is done by comparing the explanations automatically produced by these methods \cite{martens2007comprehensible} or by measuring how much they fulfil certain notions of explainability, such as completeness, through the use of either quantitative or qualitative metrics\cite{cui2019integrative} (further discussed in section \ref{xaievaluation}). 

\section{Notions related to the concept of explainability}\label{notions}
Explaining a model induced from data by employing a specific learning technique is not a trivial goal. A body of literature focused on achieving such a goal by investigating and attempting to define the concept of \textit{explainability}, leading to many types of explanation and the formation of several attributes and structures. To organise these, the specific following clusters are proposed:
\begin{itemize}
\item {\verb|attributes of explainability|} - it contains criteria and characteristics used by scholars to try to define the construct of `explainability';
\item {\verb|types of explanation|} - it includes the different ways scholars reported explanations for their ad-hoc applications, what pieces of information are included or left out;
\item {\verb|structure of an explanation|} - it contains the various components an explanation can be constructed on, such as causes, context, and consequences of a model's prediction as well as their ordering.
\end{itemize}

\subsection{Attributes of explainability}\label{xaiattributes}
One of the principal reasons to produce an explanation is to gain the trust of users \cite{dzindolet2003role}. Trust is the main way to increase users' confidence with a system \cite{tintarev2007survey} and to make them feel comfortable while controlling and using it \cite{lipton2018mythos}. Besides trust, researchers determined other positive effects brought by explainability. 
According to \cite{ha2018designing}, it is part of human nature to assign causal attribution of events. A system that provides a causal explanation on its inferential process is perceived more human-like by end-users as a consequence of the innate tendency of human psychology to anthropomorphism. Thus, several scholars spoke at length about causality which is considered a fundamental attribute of explainability \cite{fox2017explainable,chajewska1997defining,holzinger2019causability,lipton2018mythos,miller2017explainable}. Explanations must make the causal relationships between the inputs and the model's predictions explicit, especially when these relationships are not evident to end-users. Data-driven models are designed to discover and exploit associations in the data, but they cannot guarantee that there is a causal relationship in these associations. As pointed out in \cite{lipton2018mythos}, the task of inferring causal relationships strongly depends on prior knowledge, but some associations might be completely unexpected and not explainable yet. Scientists can use these associations to generate hypotheses to be tested in scientific experiments; however, this is outside the scope of the methods for explainability. Other four reasons supporting the necessity to explain the logic of an inferential system or a learning algorithm were suggested in \cite{adadi2018peeking}: 
\begin{itemize}
    \item \textit{explain to justify} - the decisions made by utilising an underlying model should be explained in order to increase their justifiability;
    \item \textit{explain to control} - explanations should enhance the transparency of a model and its functioning, allowing its debugging and the identification of potential flaws;
    \item \textit{explain to improve} - explanations should help scholars improve the accuracy and efficiency of their models;
    \item \textit{explain to discover} - explanations should support the extraction of novel knowledge and the learning of relationships and patterns.
\end{itemize}

Despite the widely recognised importance of explainability, researchers are striving to determine universal, objective criteria on how to build and validate explanations \cite{miller2017explanation}. Numerous notions underlying the effectiveness of explanations were proposed in the literature (as summarised in table \ref{tab:criteria}). \cite{miller2017explanation} surveyed 250 articles from the fields of Philosophy, Psychology and Cognitive Science to analyse in depth how people define, generate, select, evaluate and present explanations. The author also presented an interesting definition of XAI as a human-agent interaction problem where the agent reveals the underlying causes to its or another agent's decision process. In other words, XAI is believed to be a subset of the human-agent interaction field that can be defined as the intersection of AI, social science and HCI.\\

\begin{table}[htbp]
\footnotesize
  \caption{Definition of the notions related to the concept of explainability}
  \label{tab:criteria}
  \begin{tabular}{m{1.7cm} m{10.9cm}}
    \hline
    Notion & Description \& Reference\\
    \hline
    Algorithmic\newline transparency & The degree of confidence of a learning algorithm to behave `sensibly' in general \cite{dam2018explainable,preece2018asking}\\
    Actionability & The capacity of a learning algorithm to transfer new knowledge to end-users \cite{kulesza2015principles,kulesza2013too}\\
    Causality & The capacity of a method for explainability to clarify the relationship between input and output \cite{fox2017explainable,chajewska1997defining,ha2018designing,holzinger2019causability,lipton2018mythos,miller2017explainable}\\
    Completeness & The extent to which an underlying inferential system is described by explanations \cite{cui2019integrative, kulesza2015principles,kulesza2013too}\\
    Comprehensibility &  The quality of the language used by a method for explainability \cite{askira1998knowledge,alonso2018bibliometric, bibal2016interpretability,bratko1997machine, doran2017does, dovsilovic2018explainable,freitas2006we,goebel2018explainable, watson2019clinical}\\
    Cognitive relief & The degree to which an explanation decreases the ``surprise value'' which measures the amount of cognitive dissonance between the explanandum and the user's beliefs. The explanandum is something unexpected by the user that creates dissonance with his/her beliefs \cite{chajewska1997defining}\\
    Correctability & The capacity of a method for explainability to allow end-users  make technical adjustments to an underlying model \cite{kulesza2015principles,kulesza2013too}\\
    Effectiveness & The capacity of a method for explainability to support good user decision-making \cite{tintarev2011designing, tintarev2015explaining, chander2018evaluating} \\
    Efficiency & The capacity of a method for explainability to support faster user decision-making \cite{tintarev2011designing, tintarev2007survey, tintarev2015explaining}\\
    Explicability & The degree of association between the expected behaviour of a robot to achieve assigned tasks or goals and its actual observed actions \cite{zhang2017plan}\\
    Explicitness & The capacity of a method to provide immediate and understandable explanations \cite{melis2018towards}\\
    Faithfulness & The capacity of a method for explainability to select truly relevant features \cite{melis2018towards}\\
    Intelligibility & The capacity to be apprehended by the intellect alone \cite{abdul2018trends, chromik2019dark,dodge2018should, lim2009and, lim2019these}\\
    Interactivity & The capacity of an explanation system to reason about previous utterances both to interpret and answer users' follow-up questions \cite{moore1993planning, madumal2019grounded}\\
    Interestingness & The capacity of a method for explainability to facilitate the discovery of novel knowledge and to engage user's attention \cite{bibal2016interpretability, freitas1999rule, freitas2006we, bratko1997machine, sequeira2019interestingness}\\
    Interpretability & The capacity to provide or bring out the meaning of an abstract concept \cite{bibal2016interpretability,biran2017explanation,carrington2018measures,doran2017does,dovsilovic2018explainable,miller2017explanation,montavon2017methods,sassoon2019explainable,sundararajan2019exploring,van2013research,vellido2012making,zhou2008low}\\
    Informativeness & The capacity of a method for explainability to provide useful information to end-users \cite{lipton2018mythos}\\
    Justifiability & The capacity of an expert to assess if a model is in line with the domain knowledge \cite{adadi2018peeking, bibal2016interpretability, biran2017explanation, gregor1999explanations}\\
    Mental Fit & The ability for a human to grasp and evaluate a model \cite{bibal2016interpretability, weihs2003combining}\\
    Monotonicity & The relationship between a numerical predictor and the predicted class that occurs when increasing the value of the predictor leads to either always increase or decrease the probability of an instance's membership to the class \cite{freitas2014comprehensible}\\
    Persuasiveness & The capacity of a method for explainability to convince users perform certain actions \cite{tintarev2011designing, tintarev2007survey, tintarev2015explaining}\\
    Predictability & The capacity to anticipate the sequence of consecutive actions in a plan \cite{zhang2017plan}\\
    Refinement & The capacity of a method to guide experts in improving the performance/robustness of a model \cite{liu2017towards}\\
    Reversibility & The capacity to allow end-users to bring a ML-based system to an original state after it has been exposed to an harmful action that makes its predictions worse \cite{kulesza2015principles,kulesza2013too}\\
    Robustness & The persistence of a method for explainability to withstand small perturbations of the input that do not change the prediction of the model \cite{alvarez2018robustness, liu2017towards}\\
    Satisfaction & The capacity of a method  to increase the ease of use and usefulness of a ML-based system \cite{tintarev2011designing, tintarev2007survey, tintarev2015explaining}\\
    Scrutability /\newline diagnosis & The capacity of a method for explainability to inspect a training process that fails to converge or does not achieve an acceptable performance \cite{liu2017towards, tintarev2011designing, tintarev2007survey}\\
    Security &  The reliability of a model to perform to a safe standard across all reasonable contexts \cite{mcallister2017concrete}\\
    Selection /\newline simplicity & The ability of a method for explainability to select only the causes that are necessary and sufficient to explain the prediction of an underlying model \cite{miller2017explainable}\\
    Sensitivity & The capacity of a method for explainability to reflect the sensitivity of the underlying model with respect to variations in the input feature space \cite{kindermans2017reliability, sundararajan2017axiomatic}\\
    Simplification & The capacity to reduce the number of variables under consideration to a set of principal ones \cite{offert2017know}\\
    Soundness & The extent to which each component of an explanation's content is truthful in describing an underlying system \cite{kulesza2015principles,kulesza2013too}\\
    Stability & The consistency of a method to provide similar explanations for similar/neighboring inputs \cite{melis2018towards}\\
    Transparency & The capacity of a method to explain how the system works even when it behaves unexpectedly \cite{chromik2019dark, dam2018explainable, dovsilovic2018explainable, lyons2013being, sassoon2019explainable, thelisson2017regulatory, thelisson2017towards, tintarev2011designing, tintarev2007survey, wachter2017transparent, weller2017challenges, zhou2008low}\\
    Transferability & The capacity of a method to transfer prior knowledge to unfamiliar situations \cite{lipton2018mythos}\\
    Understandability & The capacity of a method of explainability to make a model understandable \cite{abdul2018trends,alonso2018bibliometric, bibal2016interpretability, liu2017towards,paez2019pragmatic}\\
  \hline
\end{tabular}
\end{table}

Two studies on \textit{explainability} demonstrated that this concept is utilised in several fields, spanning from Mathematics, Physics, Computer Science to Engineering, Psychology, Medicine and Social sciences \cite{alonso2018bibliometric, abdul2018trends}. Explainability is often replaced with the notion of \textit{interpretability}, considered as synonyms within the general AI community, and in particular by those scholars in automated learning and reasoning, whereas it seems that the software engineering community prefers the term \textit{understandability} \cite{alonso2018bibliometric}. Generally speaking, interpretability is often defined as the capacity to provide or bring out the meaning of an abstract concept and understandability as the capacity to make the model understandable by end-users (see table \ref{tab:criteria}). However, other definitions are proposed in the literature. Explainability or interpretability is defined in \cite{dam2018explainable} as ``the degree to which a human observer can understand the reason behind a decision (or a prediction) made by the model''. An interesting distinction between the concepts of \textit{interpretation} and \textit{explanation} was proposed in \cite{montavon2017methods}. On one hand, an interpretation is the mapping of an abstract concept (as a predicted class) into a domain that the human can make sense of, such as, for instance, images or texts that can be inspected and classified by people. On the other hand, an explanation is the collection of features of an interpretable domain that contributed to produce a prediction for a given item. The authors of \cite{montavon2017methods} did not specify how to determine this collection of features. The selection criteria are to be decided by researchers according to several factors like the type of input data and the degree of refinement in the explanation demanded by end-users. 
An expansion of the definition of interpretability through the determination of its main characteristics was presented in \cite{melis2018towards,miller2017explanation, sundararajan2019exploring}. In detail, \cite{melis2018towards} suggested the following requirements: (I) \textit{fidelity} - the representation of inputs and models in terms of concepts should preserve and present to end-users their relevant features and structures, (II) \textit{diversity} - inputs and models should be representable with few non-overlapping concepts, and (III) \textit{grounding} - concepts should have an immediate human-understandable interpretation.
These requirements were further expanded in \cite{miller2017explanation} by listing a set of characteristics that an explanation should possess:
\begin{itemize}
    \item \textit{contrastive nature of explanations} - people seek for an explanation when they are presented with counterfactual and/or counter-intuitive events;
    \item \textit{selectivity of explanations} - people usually do not expect that an explanation contains the actual and complete list of the causes of an event, but only a selection of the few causes deemed to be necessary and sufficient to explain it. Authors point out the risk that this selection might be influenced by cognitive biases;
    \item \textit{social nature of explanations} - explanations are part of a dialogue aiming at transferring knowledge, therefore, they are based on the beliefs of both the explainer and explainee;
    \item \textit{irrelevance of probabilities to explanations} - referring to the occurrence probabilities of events or to the statistical relationships between causes and events does not produce a satisfactory and intuitive explanation. Explanations are more effective when they refer to the causes and not to their likelihood. 
\end{itemize}
Four further requirements for enhancing the interpretability of visual explanations were added in \cite{sundararajan2019exploring}: i) \textit{graphical integrity} - the representations should highlight the features that contribute the most to the final predictions and distinguish those with positive and negative attribution, ii) \textit{coverage} - a large fraction of the most important features should be visible in the representation, iii) \textit{morphological clarity} - the important features should be clearly displayed, their visualization cannot be `noisy', and iv) \textit{layer separation} - the representation cannot occlude the raw image which should be visible for human inspection.
Other two notions strongly correlated with interpretability are \textit{comprehensibility} \cite{bibal2016interpretability} and \textit{intelligibility} \cite{abdul2018trends}. However, scholars highlighted some differences. \cite{doran2017does} proposed to distinguish between \textit{interpretable systems}, systems in which end-users can mathematically analyse algorithms, and \textit{comprehensible systems} that ``emit symbols enabling user-driven explanations of how a conclusion is reached''. Two studies \cite{abdul2018trends,lim2019these} defined intelligibility as an attribute of user-centric reasoned explanations that are easily interpretable by end-users and that draws from foundational concepts of other disciplines such as Philosophy and Cognitive Psychology. Additionally, both studies recommended exploiting the experience and knowledge of the HCI community in making interfaces that empower people to assure that intelligibility will be one of the core requirements of the next generation of AI systems.
Other authors focused on breaking some of the notions identified in table \ref{tab:criteria} into sub-notions or on assigning further requirements. 
For example, three sub-notions related to \emph{transparency} that should be achieved by any learning model were defined in \cite{dam2018explainable,lipton2018mythos}: 
\begin{itemize}
    \item \textit{simulatability} - the capacity of a model to allow a user to understand its structure and functioning entirely;
    \item \textit{decomposability} - the degree to which a model can be decomposed into its individual components (input, parameters and output) and of their intuitive explainability;
    \item \textit{algorithmic transparency} - the degree of confidence of a learning algorithm to behave 'sensibly' in general (see also table \ref{tab:criteria}). 
\end{itemize}

However, according to \cite{lipton2018mythos}, it is not possible to achieve algorithmic transparency in neural networks because of the current incapacity of experts to understand the inferential process of these models and to prove that they work correctly on new, unseen observations. Scholars attempted to overcome this shortcoming by finding methods to trace the predictions of a model to the most influential features of the input. Examples of these methods are heat-maps \cite{simonyan2014deep} which are created by back-propagating the predictions of a model to the input space and highlighting relevant pixels. Alternatively, \cite{lou2012intelligible} proposed a solution to satisfy the simulatability and decomposability properties by substituting black-box models with Generalized Additive Models (GAMs). GAMs are linear combinations of simple models trained on a single feature of an input dataset, thus allowing end-users to quantify the contribution of each feature to the outcome. 
However, transparency must be handled with caution because it can be dangerous under certain circumstances, as highlighted in \cite{weller2017challenges}. Requiring that data and models are fully visible to end-users prevents the creation of intellectual properties; this can significantly slow down the development of new technologies. Moreover, data can contain sensitive or personal information which cannot be made public without affecting people's privacy. Finally, the displaying of more information might push a researcher to optimise a model on specific instance(s) but deteriorating its overall performance and degree of generalisability.
Scholars extensively investigated \textit{sensitivity} \cite{kindermans2017reliability, sundararajan2017axiomatic}. In this context, sensitivity is considered as the sensibility of explanations to variations in the input features, model implementation and, subsequently, in the model's predictions. \cite{kindermans2017reliability} introduced the requirement of \textit{input invariance} meaning that a method for explainability must mirror the sensitivity of the underlying model with respect to transformations of the inputs in order to ensure a reliable interpretation of their contribution to each prediction. \cite{sundararajan2017axiomatic} focused on the sensitivity of methods for explainability specifically designed for neural networks, in particular those that quantify the contribution of input features to the predictions, such as DeepLift \cite{shrikumar2017learning} and Layer-wise Relevance Propagation (LRP) \cite{bach2015pixel}. In this case, a method for explainability satisfies the sensitivity requirement if it assigns a non-zero contribution to an input feature when two instances, in the input space, differ in that feature only but lead to different predictions. According to \cite{sundararajan2017axiomatic}, methods for explainability must also fulfill the requirement of \textit{implementation invariance}. This suggests that a method applied to functionally equivalent neural networks should assign identical contributions to the features of the input. Two neural networks are \textit{functionally equivalent} if their predictions are equal for all inputs despite having different implementations and architectures.
Finally, scholars identified various factors that might affect the \textit{interestingness} of a model, in particular of the rule-based ones \cite{freitas1999rule, freitas2006we}. First, \textit{rule size} is the number of instances satisfied by a rule. Usually, small size rules are undesirable as they explain only a few instances. The main aim is to discover rules that cover a large portion of the input data. However, there are situations where small rules might capture exception occurring in the data that can be of interest for scientists. Second, \textit{imbalance of class distributions} occurs when the instances belonging to a class are more frequent than those of another class. It might be more difficult, hence more interesting, to discover those rules aimed at predicting the minority classes. \textit{Attribute costs} represent the cost to get access to the actual value of an attribute of the data. For example, it is easy to assess the gender of a patient but the determination of some health-related attributes can require an expensive investigation. Rules that utilise only `cheap' attributes are more interesting. Eventually, the interestingness of a rule must take into account the \textit{misclassification costs}. In some domain of application, the erroneous classification of an instance might have a significant impact, not only in terms of money. In case of medical diagnosis, classifying as healthy a patient affected by a lethal disease might lead to premature death. Interestingness was also examined for Reinforcement Learning (RL) agents which are designed to take actions in a specific environment with the aim to maximize a cumulative reward \cite{sequeira2019interestingness}.
The authors proposed a framework to make the behaviour of these agents explainable by analysing their historical interactions with the environment and extracting a few \textit{interestingness elements}. Examples of interesting elements of these interactions are the portion of environment observed by the agent, the frequency of certain types of interactions and the cost (in terms of a reward) of the interactions carried out.

\subsection{Types of explanations}\label{explanation_type}
Researchers tried to create a classification system for the types of explanation suitable for interpreting the logic of learning algorithms. A method for explainability should answer several questions to form an exhaustive explanation. The two most common questions are \textit{why} and \textit{how} the model under scrutiny produces its predictions/inferences \cite{herlocker2000explaining,krause2016interacting,preece2018asking,wang2007recommendation}. However, scholars identified other questions that might arise and that require different answers, thus different types of explanations \cite{ribera2019can}. Additionally, as pointed out in \cite{de2017people, harbers2009study}, distinct behaviours, distinct problems and distinct types of users require distinct explanations. This has led to many ad-hoc classifications that are domain-dependent and are hard to be merged into one. 
For example, \cite{wick1989reconstructive} focused on the types of users of methods for explainability. They proposed a two-class system consisting of \textit{traced-based explanations}, useful for system designers, that accurately reflects the reasoning implemented within a model, and \textit{reconstructive explanations}, designed for end-users, based on an active, problem-solving approach. A reconstructive explanation tends to build a `story' exposing the input features contributing to a prediction. For instance, an image of a bird was assigned to a certain class because of the colour of the bird. However, the model might have analysed other features that did not influence the final assessment, like the image's background. These characteristics can be included in the traced-based explanations but excluded from the reconstructive explanations. The same scholars also developed Reconstructive EXplanation (REX) \cite{wick1989reconstructive,wick1993second}, an explanatory tool capable of producing reconstructive textual explanations for expert systems. REX is built on a model that maps the execution of the expert system onto a textbook representation of the domain. A textbook representation presents the domain knowledge in human-understandable explanations, much of which comes from domain textbooks. The explanation consists of mapping over key elements from the execution trace and expanding on them using the more structured textbook knowledge, which is a collection of relationships between cues, hypotheses and goals as illustrated by this example: ``The presence of damages to the drainage pipes is a sign that the cause of an excessive high uplift pressures on a concrete dam is internal erosion of soil under the dam. Erosion would lead to broken pipes, therefore slowing drainage and causing high uplift pressures''. The goal is to determine the cause of high uplift pressure on a concrete dam, the cues consist of the presence of broken pipes and the hypothesis is the erosion of soil. 
Another classification of the types of explanations was proposed in \cite{haynes2009designs} for intelligent systems which include intelligent agents, such as those AI assistants utilised in customer support chats, or other support decision systems like those for medical diagnoses.
Here, traced-based explanations were defined as \textit{mechanistic explanations} and correspond to the answer of the question ``How does it work?''. Hence, they must offer insights into the causes and consequences of events and how these events and the different components of the intelligent systems interact to give rise to complex actions. Reconstructive explanations were instead called \textit{ontological explanations} and describe the structural properties of the intelligent systems: its components, their attributes, and how they are related to each other. \cite{haynes2009designs} also added a third category, referred to as \textit{operational explanations} which respond to the question ``How do I use it?'' by relating goals to the mechanics designed to realise them. 
A more articulated classification of the types of explanations was introduced in \cite{sheh2017introspectively} and it is based on five types of explanations that intelligent systems should produce. The first one, \textit{teaching explanations}, aims at informing humans about the concepts learned by the system such as, for example, the presence of some physical constraints (walls or other obstacles) that can limit its actions. \textit{Introspective tracing explanations} have the goal of finding the cause of and the solution to a fault whilst \textit{introspective informative explanations} aim at explaining predictions based on the reasoning process to improve human-system interaction. The last two types of explanations, \textit{post-hoc explanations} and \textit{execution explanations}, are respectively focused on explaining the decisions and their execution without necessarily following the same reasoning process and directly linking them with the inputs. An example of post-hoc and execution explanation is a robot describing the path it wants to follow to go from point A to point B and all the movements it must do to cover that path. This explanation can mention the characteristics of the surrounding environment that have been considered while planning the path, but it does not mention that alternative paths were considered and discarded and the reasons beyond these decisions.
Finally, \cite{barzilay1998new} presented a classification of the types of knowledge intrinsically embedded in an explanation. Explanations based on \textit{reasoning domain knowledge} focus on the domain knowledge needed to perform reasoning, including rules and terminology. \textit{Communication domain knowledge} is instead about the domain knowledge needed to inform, clearly and comprehensively, end-users about the underlying domain, and it might include additional information not strictly necessary for reasoning. Eventually, \textit{domain communication knowledge} focuses on how to communicate within a certain domain of application and it deals with practical aspects of the communication process, such as the language to be used, the most effective strategies for effective explanations and the communication medium. This knowledge must be tuned to the prior knowledge and cognitive state of the hearer.

\subsection{Structures of explanations}\label{explnation_structure}
The most effective way to structure explanations is still an open problem despite being tackled by several scholars. As highlighted in \cite{lombrozo2006structure}, two properties of the structure of an explanation can have a significant effect on learning, namely the capacity to ``accommodate novel information in the context of prior beliefs and do so in a way that fosters generalization''. 
As prior beliefs greatly vary according to the application field and the domain knowledge of end-users, researchers examined and proposed different structures for explanations which are domain-dependent.
The first studies on the most suitable and effective structures of textual explanations were carried out in the 80-90s and focused on interpreting the inferential process of expert models. Most of these explanations were planned as dialogues where end-users were allowed to ask a (limited) number of questions via an explanatory tool. Blah \cite{weiner1980blah}, an example of these tools, was primarily concerned with structuring explanations so that they do not appear too complex. It was based on a series of psycho-linguistic studies that analyzed how human beings explain decisions, choices, and plans to one another. 
Different ways to structure a conversational explanation, or dialogue, to successfully transfer knowledge from an explainer to an explainee were listed in \cite{walton2011dialogue, cawsey1991generating, cawsey1993planning,cawsey1993user, madumal2019grounded}. All these studies proposed to split a dialogue into three stages: opening, explanation and closing stage. Each stage has to obey a set of rules to ensure that the knowledge about the model's inferential process can be successfully transferred to end-users. On one hand, \cite{madumal2019grounded} grounded this three-stage formal protocol on the data collected from almost four hundred real dialogues which were examined to detect the key components of an explanation, the relationships between them and their order of occurrence. These main components can be synthesised by a set of questions (mainly how, why and what) and the relative arguments presented by an explainer to an explainee who, respectively, answer the questions and acknowledge the explanation or challenge it with counterfactual examples. 
On the other hand, \cite{cawsey1991generating, cawsey1993planning,cawsey1993user} focused on the most effective set of rules to manage interactive dialogues with interruptions from the user while maintaining coherence between the different sections of an explanation. They also developed a tool, called EDGE, that generates dialogues based on these rules. EDGE updates assumptions about the user's knowledge based on his/her questions and uses this information to influence the further planning of the explanation. Other studies on interactive dialogues \cite{pollack1982user,johnson1993explanation, moore1989planning, moore1989reactive, moore1991reactive} focused on the structure, the language and main components (what pieces of information must be included) of these dialogues.
Based in these early studies, \cite{core2006building,gomboc2005design,lane2005explainable,van2004explainable} proposed a modular architecture for explaining the behavior of simulated entities in military simulations. It consists of three modules: a reasoner, a natural language generator and a dialogue manager. The user can stop simulation and query about what happened at the current time point by selecting questions from a list. The dialogue manager orchestrates the system’s response: firstly, by using the reasoner to retrieve the relevant information from a relational database, then producing English responses using the natural language generator. 
More recently, interactive dialogues were used as the explanation format of choice in knowledge-based systems other than expert systems. AutoTutor \cite{graesser2005autotutor}, designed to be integrated into tutoring systems, is grounded on learning theories and tutoring research. It simulates a human tutor by holding a conversation with the learner in natural language.\\ 

The explanations of task planning systems, according to \cite{fox2017explainable,langley2017explainable}, must contain information on (I) why a planner choose an action, (II) why a planner did not choose another action, (III) why the decisions of a planner are the best among a set of possible alternatives, (IV) why certain actions cannot be executed and (V) why one needs or does not need to change the original plan.
The criterion of \textit{episodic memory} was added to the above list by \cite{langley2017explainable}, whereby an agent should remember all the factors that influenced the generation and execution of a plan such as ``states, actions, and values considered during plan generation, traces of plan execution in the environment, and anomalous events that led to plan revision''. 
A formal framework to generate \textit{preferred explanations} of a plan was introduced in \cite{sohrabi2011preferred}. Preferences over explanations must be contextualized with respect to complex observational patterns. Actions might be affected by several causes and requires reflecting on the past, meaning that explanations must take into consideration previous events and information.

\section{Development of new methods for explainability}\label{newmethods}
More than 200 scientific articles were found that aim at developing new methods for explainability. Over time, researchers have tried to comprehend and unfurl the inner mechanics of data- driven, knowledge-driven models in various ways. 
From an examination of these articles, two main criteria exist for discriminating methods for explainability: 
\begin{itemize}
\item {\verb|scope|} - it refers to the scope of an explanation that can be either \textit{global} or \textit{local}. In the former case, the goal is to make the entire inferential process of a model transparent and comprehensible as a whole. In the latter case, the objective is to explicitly explain each inference of a model   \cite{bengio2013representation, dam2018explainable, lipton2018mythos, samek2019towards};
\item {\verb|stage|} - it refers to the stage at which a method generates explanations. \textit{Ante-hoc} methods are generally aimed at considering explainability of a model from the beginning and during training to make it naturally explainable whilst still trying to achieve optimal accuracy or minimal error \cite{dovsilovic2018explainable, lou2012intelligible, lou2013accurate}; \textit{post-hoc} methods are aimed at keeping a trained model unchanged and mimic or explain its behaviour by using an external explainer at testing time \cite{dovsilovic2018explainable, lipton2018mythos, montavon2017methods, paez2019pragmatic}.
\end{itemize}

Taking into account the articles examined in this systematic review, and inspired by the classification  system in \cite{guidotti2018survey}, we propose additional criteria:
\begin{itemize}
\item {\verb|problem type|} - methods for explainability can vary according to the underlying problem: \textit{classification} or \textit{regression};
\item {\verb|input data|} - the mechanisms followed by a model to classify images can substantially differ from those used to classify textual documents, thus the input format of a model (\textit{numerical/categorical}, \textit{pictorial}, \textit{textual} or \textit{times series}) can play an important role in constructing a method for explainability;
\item {\verb|output format|} - similarly, different formats of explanations useful for different circumstances can be considered by a method for explainability: \textit{numerical}, \textit{rules}, \textit{textual}, \textit{visual} or \textit{mixed}.
\end{itemize}

\begin{figure}[!ht]
\begin{minipage}{\textwidth}
\centering
  \includegraphics[scale=0.35, align=c]{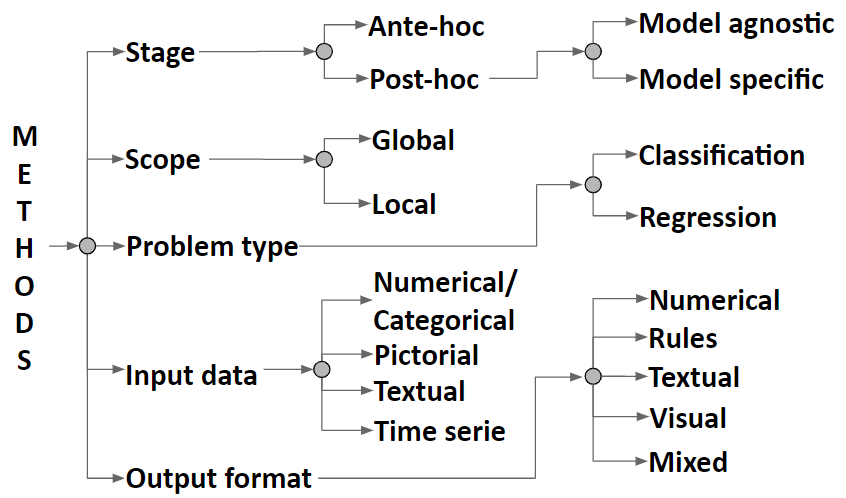}
  \includegraphics[scale=0.37, align=c]{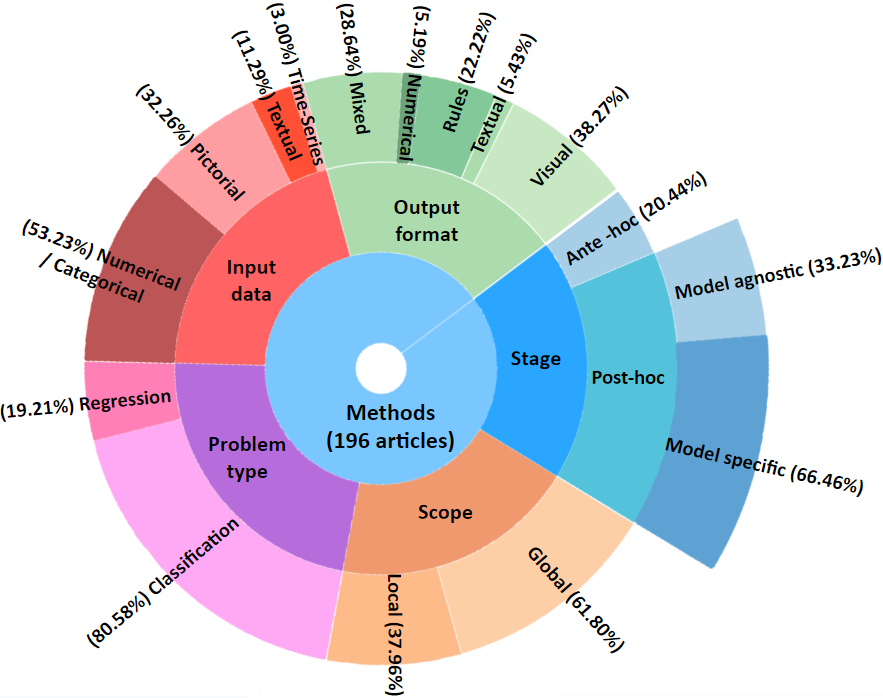}
  \caption{Classification of methods for explainability (left) and  distribution of articles across categories (right).}
  \label{fig:methods_tree_sunburst}
\end{minipage}
\end{figure}

Figure \ref{fig:methods_tree_sunburst} depicts the main branches of methods for explainability and shows the distribution of the articles across these branches.
Each of the many methods for explainability retrieved from the scientific literature can be robustly described by using the five categories of figure \ref{fig:methods_tree_sunburst} (stage, scope, problem type, input data and output format). 
Additionally, as it is possible to notice from Figure \ref{fig:methods_tree_sunburst}, the post-hoc methods are further divided into \textit{model-agnostic} and \textit{model-specific} methods \cite{guidotti2018survey}.
The former methods do not consider the internal components of a model such as weights or structural information, therefore they can be applied to any black-box model.
The latter methods are instead limited to specific classes of models. For example, the interpretation of the weights of a linear regression model is specific to the learning approach (linear regression). Similarly, methods that only work with the interpretation of neural networks are model-specific \cite{arras2019evaluating, dam2018explainable, zhang2018visual}.
Model agnosticity and specificity do not usually apply to the class of `ante-hoc' methods because their goal is to make the functioning of a model transparent, so almost all them are intrinsically model-specific \cite{dovsilovic2018explainable}. 
Some post-hoc methods for explainability can be applied both at a global or local scope \cite{bologna2018comparison} and can work for either regression or classification problems \cite{spinner2019explainer}. \\

The following sections try to succinctly describe the main classes of methods for explainability found during this systematic review, accompanied by tables for reporting their stage, scope, problem type, input data and output format and sorting them in alphabetic order.
Given the large number of methods found, it was decided to group them into five thematic classes.

\subsection{Output formats}
\emph{Visual explanations} are probably the most natural way of communicating things and a very appealing way to explain them.
Visual explanations can also be used to illustrate the inner functioning of a model via graphical tools. For instance, heat-maps can highlight specific areas of an image or specific words of a text that mostly influence the inferential process of a model by using different colours \cite{ribeiro2016should, strobelt2018lstmvis}. 
Similarly, a graphical representation can be employed to represent the inner structure of a model, such as the graphs proposed in \cite{wongsuphasawat2018visualizing} where each node is a layer of the network and the edges the connections between layers.
Another intuitive form of explanation for humans are \emph{textual explanations}, natural language statements that can be either written or orally uttered. An example is the phrase ``This is a Brewer Blackbird because this is a blackbird with a white eye and long pointy black beak'' shown by an explainer of an image classification model \cite{hendricks2018grounding}. 
A schematic, logical format, more structured than visual and textual explanations but still intuitive for humans, are \emph{rules} that can be used to explain the inferences produced by models induced from data. Rules can be in the form of `IF ... THEN' statements with \textit{AND/OR} operators and they are very useful for expressing combinations of input features and their activation values \cite{fung2005rule, bologna2017characterization}. 
Technically, rules of these type employ symbolic logic, a formalized system of primitive symbols and their combinations (example: `$(Country = USA) \wedge (28<Age<=37) \rightarrow  (Salary>50K$)' \cite{ribeiro2018anchors}). The parts before and after the $\rightarrow$ logical operator are respectively referred to as antecedent and consequent. Given this logic, rules can be implemented as fuzzy rules, linking one or more premises to a consequent that can be true to a degree, instead of being entirely true or false. This can be obtained by representing each antecedent and consequent as fuzzy sets \cite{guillaume2001designing}.
Combining fuzzy rules with learning algorithms can become a powerful tool to perform reasoning and, for instance, explain the inner logic of neural networks \cite{palade2001interpretation}.
Similarly, the combination of antecedents and consequent can be seen as an argument in the discipline of argumentation, and a set of arguments can be put together in a dialogical structure by employing attacks, the link between arguments that model conflictuality \cite{rizzo2019inferential, rizzo2018qualitative}.
Arguments and attacks form a complex structure but with high explanatory power, suitable for explaining the inner functioning of data-driven models.
Explanations can also be constructed by only employing numerical formats as crisp values, vectors of numbers, matrices or tensors as in Probe \cite{alain2017understanding} and Concept Activation Vectors (CAVs) \cite{kim2018interpretability}, two methods for explainability. A Probe consists of a linear classifier fitted to the features, treated independently, learned by each layer of a neural network. Probes are engineered to better understand the roles and dynamics of the internal layers. The numerical explanations are the probability scores assigned by the probes to each class \cite{alain2017understanding}. 
CAVs separates the activation values of a neural network's hidden layer relative to instances belonging to a class, forming a set, from those generated by the remaining part of the input dataset, forming a second set. Subsequently, a binary linear classifier is trained to distinguish the activation values of the two sets. Then, CAVs computes directional derivatives on this classifier to measure the sensitivity of the model to changes in inputs towards the class of interest. This is a scalar quantity, calculated for each class over the whole dataset, which quantifies how important a user-defined concept is to classify the input instances in the class under analysis. For example, CAVs measures how sensitive the class `zebra' is to the presence of stripes in an input image.
Eventually, the most powerful format of explanations are those that employ one or more of the formats described so far (visual, textual, rules, numeric). An example of a combination of visual and numerical explanation is utilized by Important Support Vectors and Border Classification \cite{barbella2009understanding} that provide insight into local classifications produced by a Support Vector Machine (SVM). The former method returns the support vectors which influence the most the final classification for a particular instance. The latter determines which features of a data point would need to be altered (and by how much) to be placed on the separating surface between two classes. The explanations are in the form of an interactive interface where the user can select a point and the tool shows the attributes that had the largest effect on classifying it and the closest border value. The user can modify the selected point's attributes to see how the SVM reclassifies it.
Image Caption Generation with Attention Mechanism \cite{xuk2015show} is an example of visual and textual explanations jointly employed. It returns attention maps for a combination of a Convolutional Neural Network (CNN) and a Long-Short Term Memory (LSTM) network where the CNN performs object recognition in images and the LSTM generates their captions.

\subsection{Model agnostic methods for explainability}
Several methods for explainability were designed to work with any learning technique. However, this does not mean that they can be universally applied as they might be constrained to the types of inputs of the technical problem they try to solve and the explanation they try to provide.

\subsubsection{Numeric explanations}
\begin{figure}[!ht]
\begin{minipage}{\textwidth}
\centering
  \subcaptionbox{Distill-and-Compare \cite{tan2018distill}}
    {\includegraphics[scale=0.28, align=c]{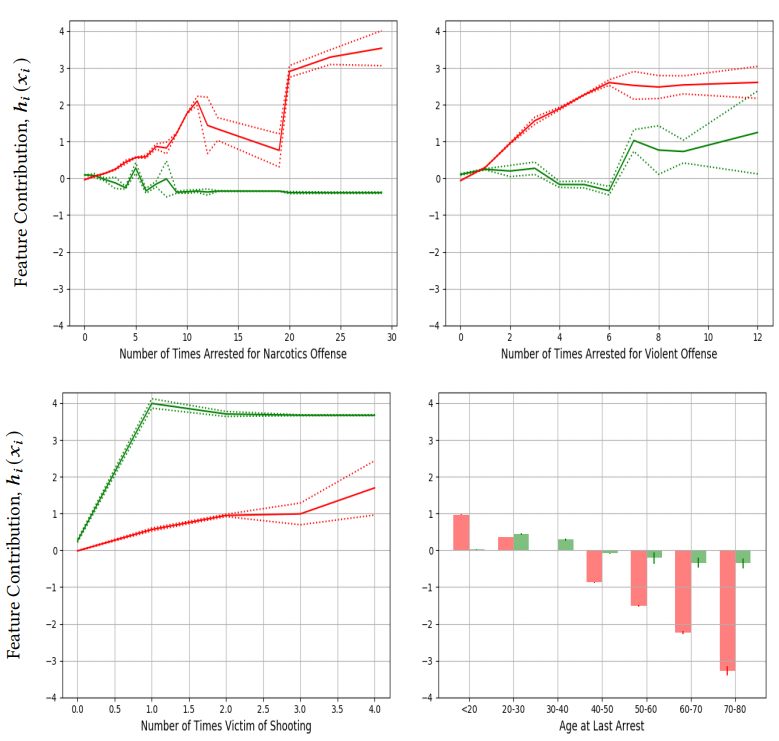}}
  \subcaptionbox{GSA \cite{cortez2011opening}}
    {\includegraphics[scale=0.28, align=c]{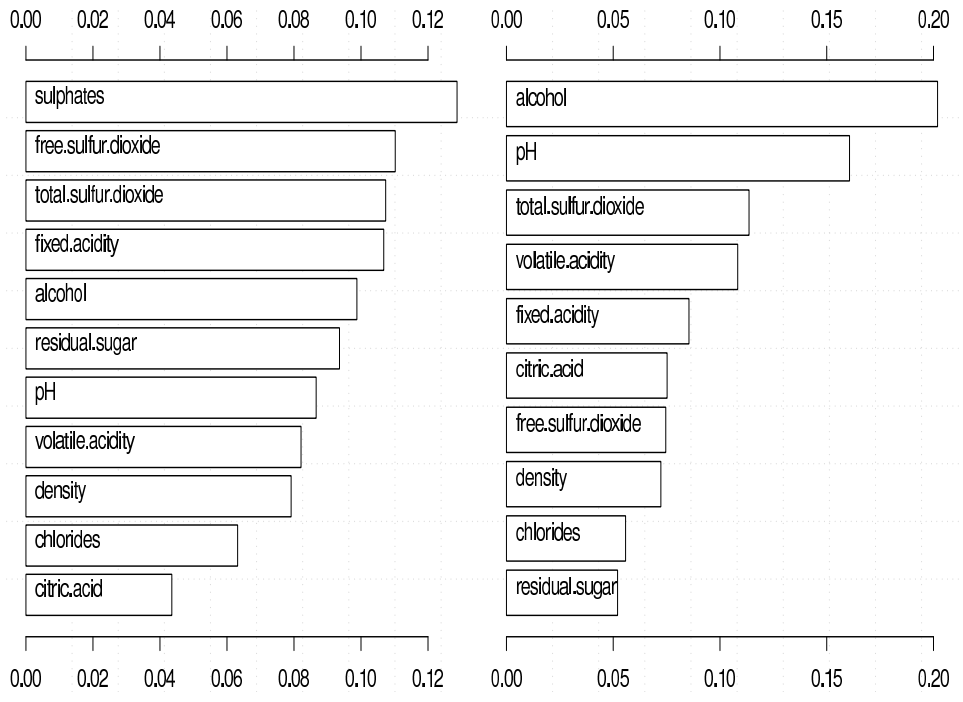}}
  \subcaptionbox{GFA \cite{adler2018auditing}}
    {\includegraphics[scale=0.28, align=c]{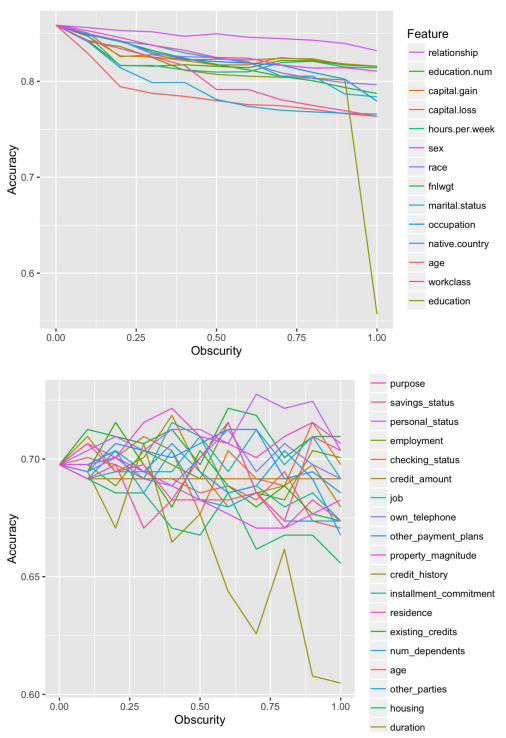}}
  \caption{Examples of numerical explanations generated by model-agnostic methods for explainability.}
  \label{fig:model_agnostic_numeric}
\end{minipage}
\end{figure}

A few model agnostic methods for explainability produce numerical explanations (see table \ref{tab:model-agnostic-numerical} and figure \ref{fig:model_agnostic_numeric}). Most of them focus on measuring the contribution of an input variable (or a group of them) with quantitative metrics.
Distill-and-Compare \cite{tan2018distill} trains a transparent, simpler model, called student, on the output obtained from a large, complex model, considered as a teacher, to mimic its inferential process. In this study, the student model was constrained to be GAMs which allow to easily assess the contribution of each feature in a numerical format. Similarly, SHapley Additive exPlanations (SHAP) \cite{lundberg2017unified} utilizes additive feature attribution methods, basically linear combinations of the input features, to build a model which is an interpretable approximation of the original model.
Some methods for explainability are based on an `input perturbation' approach and, generally speaking, they work by modifying the reported values of the variables of an input instance to cause a change in the model's prediction.
Explain and Ime \cite{robnik2008explaining,robnik2018explanation} assess respectively the contribution of a particular input variable or a set of variables. This is done by replacing the actual values of the variables describing each input instance with other values sampled from the same variable(s) and measuring the differences in the output probability scores. The assumption is that the larger the difference in the outcome, the more relevant the variable is for the prediction process. 
Similarly, the Global Sensitivity Analysis (GSA) method \cite{cortez2011opening, cortez2013using} ranks input features by quantifying the effects on the predictions of a given model when they are varied through their range of values.
\cite{Kononenko10anefficient,kononenko2013explanation,vstrumbelj2009explaining,vstrumbelj2008towards,vstrumbelj2010explanation} proposed a method to explain the prediction of a model at instance level also based on the contribution of each feature estimated by comparing the model output when all the features are known and when one or more of them are omitted. The contribution is positive for the features that lead to the prediction towards a class, negative for those that push the prediction against a class and zero when they don't have influence.
Four methods, Quantitative Input Influence (QII) functions \cite{datta2016algorithmic}, Gradient Feature Auditing (GFA) \cite{adler2018auditing}, Influence functions \cite{koh2017understanding} and Monotone Influence Measures \cite{sliwinski2017characterization}, utilize influence functions to assess the contribution of each feature to certain predictions. An influence function is a classic technique from statistics \cite{koh2017understanding} measuring the sensitivity of a model to changes in the distributions of the independent variables. The perturbation of the input can be done in different ways such as applying a constant shift (Influence functions \cite{koh2017understanding}), obscuring parts of the input (GFA \cite{adler2018auditing}), rotating, reflecting or randomly assign labels to the input (Monotone Influence Measures \cite{sliwinski2017characterization}).
Feature Importance \cite{henelius2014peek} and Feature Perturbation \cite{vstrumbelj2014explaining} are also based on algorithms that modify subsets of the input features to find groups of interacting attributes used by different classifiers and to determine the extent to which a model exploits such interactions.\\

\newpage

\subsubsection{Rule-based explanations}
\begin{figure}[!ht]
\begin{minipage}{\textwidth}
\centering
  \subcaptionbox{G-REX \cite{johansson2004accuracy}}
    {\includegraphics[scale=0.42, align=c]{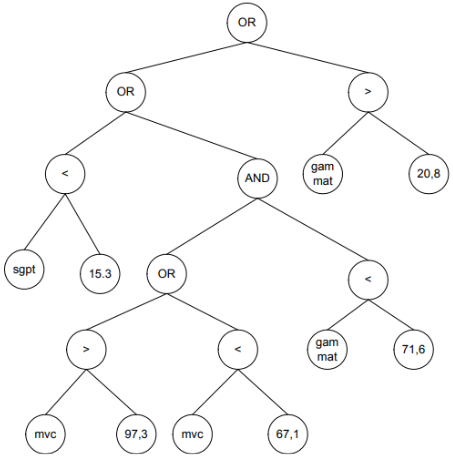}}
  \subcaptionbox{Anchor \cite{ribeiro2018anchors}}
    {\includegraphics[scale=0.4, align=c]{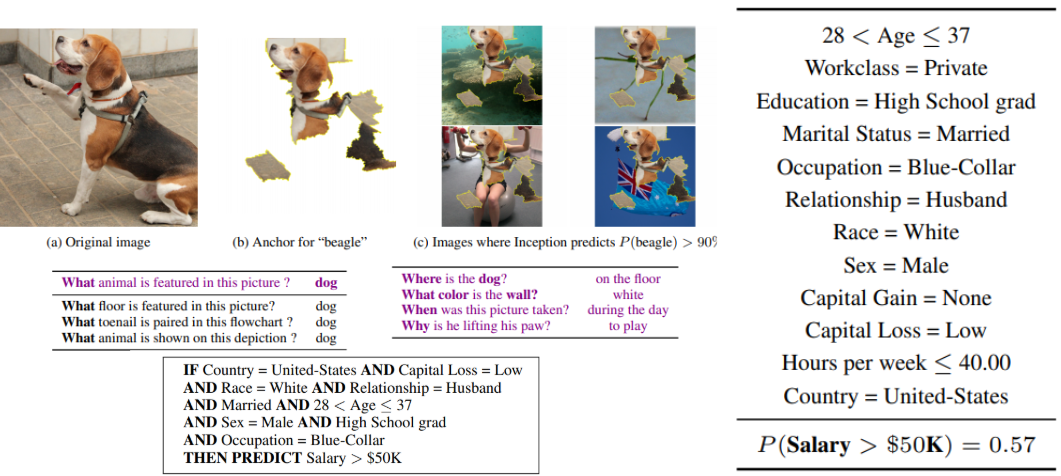}}
  \caption{Examples of rule-based explanations generated by model-agnostic methods which can be visualized as (a) a decision tree (b) a list of rules accompanied by textual and visual examples.}
  \label{fig:model_agnostic_rules}
\end{minipage}
\end{figure}

A few model-agnostic methods for explainability produce rule-based explanations by exploiting several rule-extraction techniques (see table \ref{tab:model-agnostic-rule} and figure \ref{fig:model_agnostic_rules}), such as automated reasoning-based approaches. The method presented in \cite{bride2018towards} extracts logical formulas as decision trees by combining split predicates along paths from inputs to predictions into logical conjunctions and all the paths related to an output class into logical disjunctions. These rules can be analyzed with logical reasoning techniques to extract information about the decision-making process.
Similarly, Genetic Rule EXtraction (G-REX) \cite{johansson2004accuracy,johansson2004truth} employed genetic algorithms to generate IF-THEN rules with AND/OR operators. 
Anchor \cite{ribeiro2018anchors} uses two algorithms to extract IF-THEN rules which highlight the features of an input instance, called `anchors', that are sufficient for a classifier to make a prediction. In an analogical manner, the words ``not bad'' are often used in sentences expressing a positive sentiment, and thus can be considered anchors in sentiment analyses. These two algorithms, a bottom-up formation of and a beam-search for anchors, identify the candidate rules with the highest estimated precision over a dataset where precision is equal to the fraction of correct predictions. The first algorithm starts from an empty set of rules and adds, at each iteration, a rule for each feature predicate. The second one instead starts from a set containing all the possible candidate rules and then selects the best ones in terms of precision.
Model Extraction \cite{bastani2017interpretability} and Partition Aware Local Model (PALM) \cite{krishnan2017palm} utilize decision trees (DTs) to approximate complex models with the assumption that, as long as the approximation quality is good, the statistical properties of the complex model are reflected in the interpretable ones. End-users have also the faculty to examine the DT's structure and determine whether the rules match intuition. Model Extraction generates DTs by using the Classification And Regression Trees algorithm (CART) and trains them over a mixture of Gaussian distributions fitted to the input data using expectation maximization. PALM uses a two-part surrogate model: a meta-model, constrained to be a DT, that partitions the training data, and a set of sub-models fitting the patterns within each partition.

\subsubsection{Visual explanations}
\begin{figure}[!ht]
\begin{minipage}{\textwidth}
\centering
  \subcaptionbox{Explanation Graph \cite{alvarez2017causal}}
    {\includegraphics[scale=0.31, align=c]{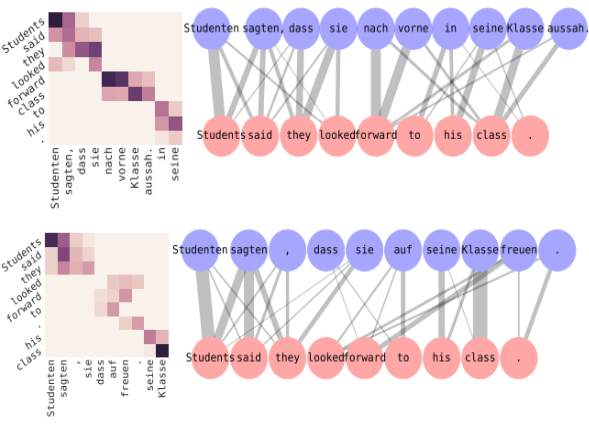}}
  \subcaptionbox{RSRS \cite{liu2012has}}
    {\includegraphics[scale=0.33, align=c]{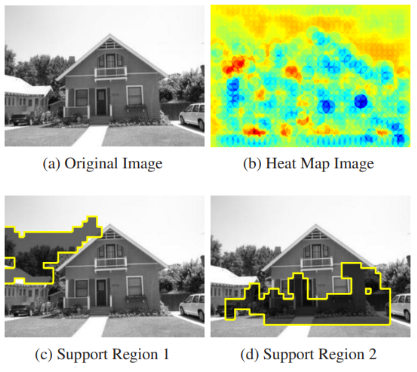}}
  \subcaptionbox{SpRAy \cite{lapuschkin2019unmasking}}
    {\includegraphics[scale=0.33, align=c]{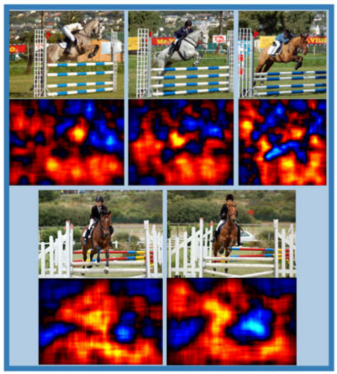}}
  \subcaptionbox{PI and ICI plots\cite{casalicchio2018visualizing}}
    {\includegraphics[scale=0.36, align=c]{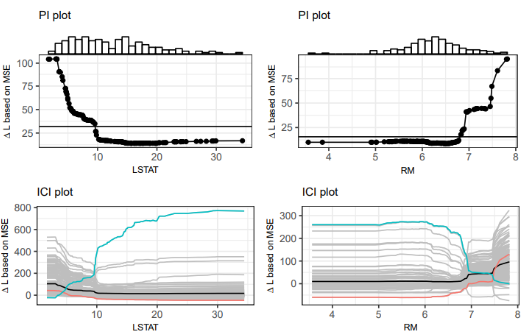}}
  \caption{Examples of visual explanations generated by model-agnostic methods as (a) graphs, (b) restricted support regions, (c) heat-maps, or (e) plots.}
  \label{fig:model_agnostic_visual}
\end{minipage}
\end{figure}

Visual explanations try to explain the inner functioning of a model via graphical aids and many model-agnostic methods exploit them (table \ref{tab:model-agnostic-visual} and figure \ref{fig:model_agnostic_visual}). 
One of the most widely used among these aids is represented by `salient masks' that are efficient ways to point out what parts of input, especially when images or texts are treated, most affect a model's prediction by superimposing a mask highlighting them. 
Layer-Wise Relevance Propagation (LRP) \cite{bach2015pixel} was developed as a model-agnostic solution to the problem of understanding image classification predictions by pixel-wise decomposition of nonlinear classifiers. In its general form, LRP assumes that the classifier can be decomposed into several layers of computation and it traces back contributions of each pixel to the final output, layer by layer, to attribute relevance to individual inputs. The pixel contributions can be visualized as heat-maps.
Spectral Relevance Analysis (SpRAy) \cite{lapuschkin2019unmasking} consists of spectral clustering on a set of LRP explanations in order to identify typical and atypical decision behaviours of an underlying data-driven model. For example, to analyse the inferential process of a classifier trained on a dataset of images of animals, SpRAy produces an LRP heat-map for each image. Then, it checks if the heat-maps highlight the area representing the animal or if, for a specific animal, the classifier is focusing on other parts, such as the presence of a rider in case the animal is a horse.
Image Perturbation \cite{fong2017interpretable} produces explanations in the forms of saliency maps by blurring different areas of the image and checking which ones most affect the prediction accuracy when perturbed. Similarly, the Restricted Support Region Set (RSRS) Detection method \cite{liu2012has} visualizes a set of size-restricted and non-overlapping regions of an image that are critical to classification. This means that if any of them is removed, then the image is wrongly classified. The explanation consists of the original image with its critical regions determined by RSRS greyed out.
The IVisClassifier \cite{choo2010ivisclassifier} is based on linear discriminant analysis (LDA). It attempts at reducing the dimension of the input data and produces heat-maps that gives an overview of the relationship among clusters in terms of pairwise distances between cluster centroids both in the original and reduced dimensional spaces.
The Saliency Detection method \cite{dabkowski2017real} utilizes a U-Net neural network trained to generate a saliency map, in a single forward pass, for any image and classifier received as inputs. The output map then highlights the parts of the image that are considered important by the classifier.\\

Some methods use other visual aids, like graphs and scatter-plots, to generate visual explanations.
The Sensitivity Analysis method \cite{baehrens2010explain} generates explanations that correspond to local gradients. These gradients indicate how a data point must be moved to change its predicted label. The explanations can be either scatter-plots of the gradient vectors or heat-maps showing which parts of the inputs must be modified to change the predicted class.
Individual Conditional Expectation (ICE) plots \cite{goldstein2015peeking} are line charts graphing the functional relationship between a predicted response and a feature for each individual observation when keeping all the other features fixed and varying the value of the feature under analysis. \cite{casalicchio2018visualizing} proposed two alternatives to ICE plots, called Partial Importance (PI) and Individual Conditional Importance (ICI) plots, which visualize the feature importance rather than its prediction. Both plots are aimed at showing how changes in a feature affect model performance. PI works at the global level by visualizing the point-wise average of all ICI curves across all observations, whereas ICI works at the local level by presenting changes for each observation. The importance of each feature is assessed using the Shapley Feature Importance measure which fairly distributes the model's performance among them according to their marginal contribution.
Explanation Graph \cite{alvarez2017causal} is based on the perturbations of the input features. It works by training a model on both the original and the perturbed data. Subsequently, a comparison of the original and perturbed input-output pairs is performed to infer causal dependencies between input and output. This method was tested across several word sequence generation tasks in Natural Language Processing (NLP) applications. The perturbed input contains statements that are semantically similar to the originals but differ in some elements (words and punctuation) and their order. The inferred dependencies are shown in graphs where the nodes contain the words of the original and perturbed inputs and their relative outputs and the edges represent the connections between them.
A Worst-Case Perturbation \cite{goodfellow2015explaining} corresponds instead to the smallest perturbation such that the perturbed input leads to an incorrect answer with high confidence. This method was applied only to images and the explanation consists of the perturbed images.
Class Signatures \cite{krause2016using} is a visual analytic interface that allows end-users to detect and interpret input-output relationships by presenting a mix of charts (line, bar charts and scatter plots) and tables organised in such a way that relationships become evident. Similarly, ExplainD \cite{poulin2006visual} was designed to explain predictions made by classifiers that use additive evidence, such as linear SVMs and regressors. The graphs produced by this method represent the contribution of each feature to the prediction and how the prediction changes when the value of a feature varies across their value ranges. 
Manifold \cite{zhang2019manifold} and MLCube Explorer \cite{kahng2016visual} are two visual analytical tools that provide comparative analysis for multiple models. They also enable end-users to define instance subsets using feature conditions, to identify instances that generate erroneous results so to explain potential reasons of these errors, and to iteratively refine the performance of a model by using different graphical aids such as scatter-plots, bar and line charts.\\

\subsubsection{Mixed explanations}
\begin{figure}[!ht]
\begin{minipage}{\textwidth}
\centering
  \subcaptionbox{ExplAIner \cite{spinner2019explainer}}
    {\includegraphics[scale=0.408, align=c]{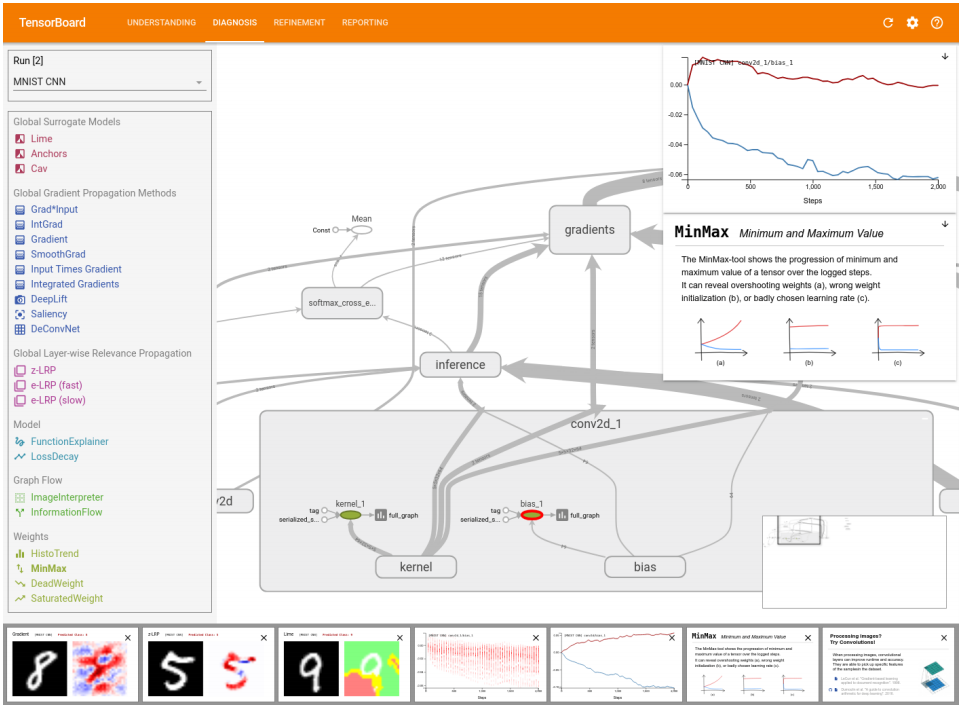}}
  \subcaptionbox{MMD-critic \cite{kim2016examples}}
    {\includegraphics[scale=0.323, align=c]{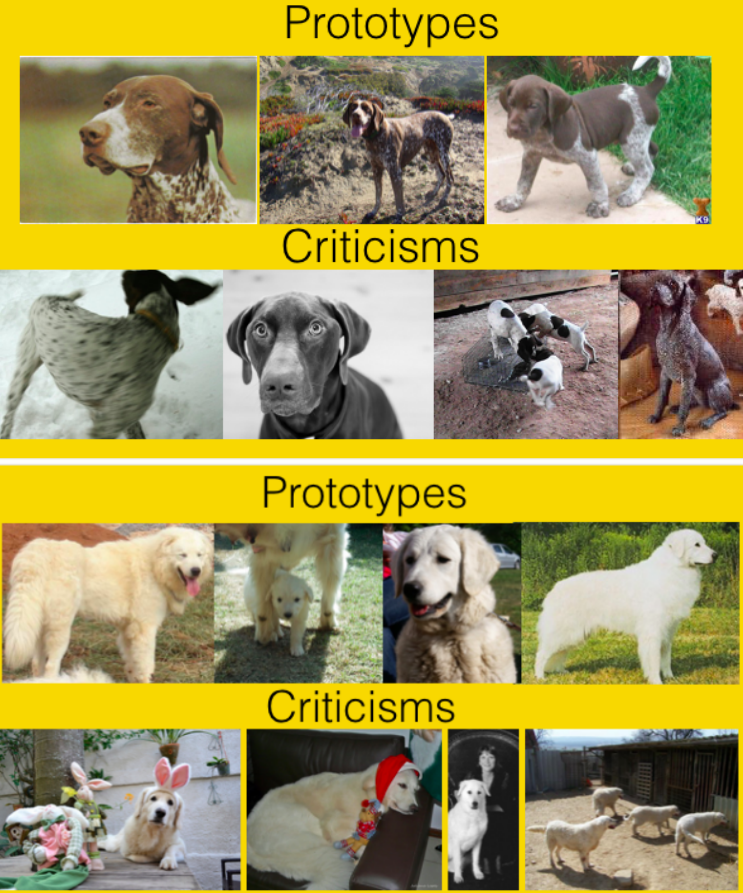}}
  \caption{Examples of mixed explanations generated by model-agnostic methods for explainability which consists of a combination of visual and textual explanations in (a) interactive interfaces or (c) a selection of prototypes from  inputs.}
  \label{fig:model_agnostic_mixed}
\end{minipage}
\end{figure}

There are many methods for explainability that produce numerical explanations along with graphical representations to make them more interpretable for lay people (see table \ref{tab:model-agnostic-mixed} and figure \ref{fig:model_agnostic_mixed}). The Functional ANOVA decomposition \cite{hooker2004discovering} quantifies the influence of non-additive interactions within any set of input variables and depict them with Variable Interaction Network (VIN) graphs where the nodes represent the variables and the edges the interactions.
The Justification Narratives method for explainability \cite{biran2014justification} consists of a simple model-agnostic mapping of the essential values underlying a classification (identified with any feature selection method) to a semantic space that automatically produces these narratives and realizes them visually (as bar-charts reporting the assessed relevance value of each variable) or textually.
ExplAIner \cite{spinner2019explainer} and Rivelo \cite{tamagnini2017interpreting} are two user interfaces showing mixes of numerical, visual and textual explanations. ExplAIner was designed to display visual and textual explanations of ML models which are the outcome of an iterative workflow of three stages: model understanding, diagnosis, and refinement. Using TensorBoard (a visualization tool developed by Google for machine learning) as a starting point, ExplAIner produces an interactive graph view of the model to be explained. The nodes of the graph represent the model's components, such as inputs, parameters and outputs, accompanied by textual definitions, and the edges represent the relationships between the components. There are also other visual explanatory tools in support of the model's graph, such as line-charts of metrics, like loss and accuracy, and examples of input data together with their relative heat-maps generated with other visual methods for explainability.
Rivelo works exclusively with binary classification problems and binary input features. It enables end-users to understand the causes behind predictions by interactively exploring a set of visual and textual instance-level explanations which lists the most relevant input features (words or image areas in a document/image), their frequency, number of instances with the feature with positive labels and are correctly/wrongly classified. \\

Other mixed explanations-based methods utilize a selection of prototypes, which are samples from the input that are correctly predicted by the model and can be considered as positive and iconic examples, or adversarial examples, which are samples misrepresented by the model and are used to generate contrastive explanations (see Section \ref{xaiattributes}). This subset helps end-users understand the model by leveraging on the human ability to induce principles from a few examples. Being a subset of a training dataset, these explanations were classified as mixed as their format depends on the nature of the input data.
The Bayesian Teaching methods for explainability \cite{yang2017explainable} selects a small subset of prototypes that would lead the model to the correct inference as if trained on the overall dataset.
\cite{khanna2019interpreting} proposed to use Sequential Bayesian Quadrature (SBQ) in conjunction with Fisher kernels to select salient training data points. All the instances in a training dataset are firstly embedded in the space induced by the Fisher kernels. This provides a way to quantify the closeness of pairs of instances which, if close enough, should be treated similarly by a model. The embedded instances are inputted into SBQ, an importance-sampling-based algorithm that estimates the expected value of a function under a distribution using discrete samples drawn from it.
Set Cover Optimization (SCO) \cite{bien2011prototype} aims at selecting prototypes in such a way that they capture the full structure of the training examples in each class of the dataset, no points have a prototype of a different class in its neighbourhood and the prototypes are as few as possible. This leads to a set cover optimization problem that can be solved approximately with standard approaches such as, for instance, `linear program relaxation with randomized rounding'.
Neighbourhood-Based Explanations \cite{caruana1999case} is based on a Case-Based Reasoning (CBR) approach. It presents to end-users the entries of a training dataset that are the most similar to the new input instance that needs to be explained. Similarity is measured through the Euclidean metrics applied to all the input features.
Adversarial examples are instead used in Evasion-Prone Samples Selection \cite{liu2018adversarial}, Maximum Mean Discrepancy (MMD)-critic \cite{kim2016examples} and Pertinent Negatives \cite{dhurandhar2018explanations}. Evasion-Prone Samples Selection aims at detecting the instances closed to the classification boundaries that can be easily misclassified if slightly perturbed whereas MMD-critic utilizes the maximum mean discrepancy and an associated witness function to identify the portions of the input space most misrepresented by the underlying model. Pertinent Negatives highlights what should be minimally and necessarily absent to justify the classification of an instance. For example, the absence of glasses is a necessary condition to say if a person has a good sight. The input data are modified by removing some parts and the pertinent negatives are identified as those perturbations that maximise the prediction accuracy.
eventually, some methods for explainability produce mixed explanations by approximating a black-box model with simpler, more comprehensible models that the end-users can inspect to assess the contribution of each feature. Local Interpretable Model-Agnostic Explanations (LIME) \cite{ribeiro2016model,ribeiro2016should} explains the prediction of any classifiers by learning a local self-interpretable model (such as linear models or decision trees), sometimes referred to as `white-box' modes, trained on a new dataset which contains interpretable representations of the original data. These representations can be the binary vectors representing the presence or absence of certain characteristics, such as words in texts or super-pixels (contiguous patch of similar pixels) in images. The black-box model can be explained through the weights of the white-box estimator which does not need to fully work globally, but it should approximate the black-box well in the vicinity of a single instance. However, the authors proposed the Sub-modular Pick (SP-LIME) to select, from an original dataset, a representative non-redundant explanation set of instances that is a global representation of the model.

\subsection{Model-specific methods for explainability based on neural networks}
\begin{figure}[!ht]
\begin{minipage}{\textwidth}
  \begin{subfigure}[b]{.18\linewidth}
  \centering
  \subcaptionbox{GradCam \cite{selvaraju2017grad}}{\includegraphics[scale=0.45]{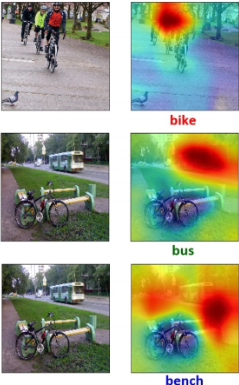}}
  \end{subfigure}
  \begin{subfigure}[b]{.22\linewidth}
  \centering
  \subcaptionbox{Guided BP \cite{goyal2016towards}}{\includegraphics[scale=0.45]{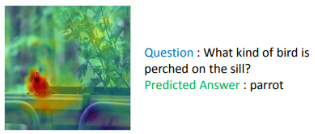}}
  \newline
  \subcaptionbox{Neural network-CBR Twin system \cite{kenny2019twin}}{\includegraphics[scale=0.33]{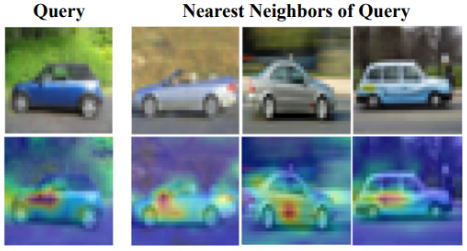}}
  \end{subfigure}
  \begin{subfigure}[b]{.23\linewidth}
  \centering
  \subcaptionbox{Compositionality \cite{li2016visualizing}}{\includegraphics[scale=0.24]{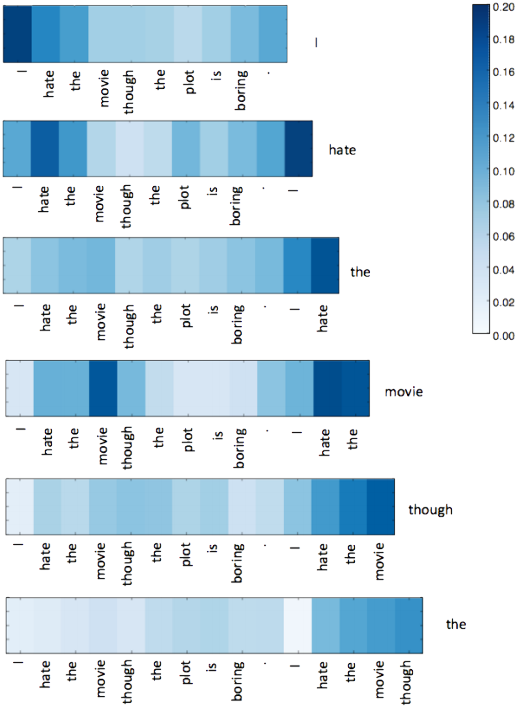}}
  \end{subfigure}
  \begin{subfigure}[b]{.14\linewidth}
  \centering
  \subcaptionbox{PCA \cite{aubry2015understanding}}{\includegraphics[scale=0.3]{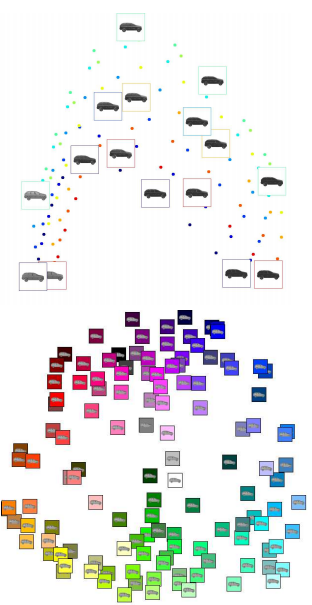}}
  \end{subfigure}
  \begin{subfigure}[b]{.18\linewidth}
  \centering
  \subcaptionbox{t-SNE maps \cite{zahavy2016graying}}{\includegraphics[scale=0.39]{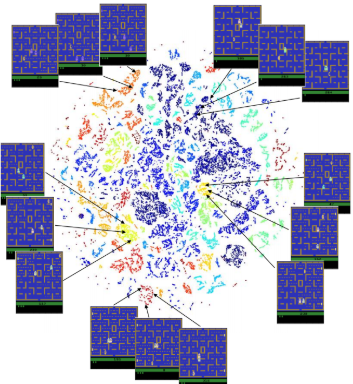}}
  \end{subfigure}
  \caption{Examples of visual explanations, as salient masks (a-d) and scatter-plots (e-f).}
  \label{fig:neural_networks_masks_plots}
\end{minipage}
\end{figure}

\begin{figure}[!ht]
\begin{minipage}{\textwidth}
\centering
  \begin{subfigure}[b]{.17\linewidth}
  \centering
  \subcaptionbox{GAN Diss \cite{bau2019gandissect}}{\includegraphics[scale=0.31]{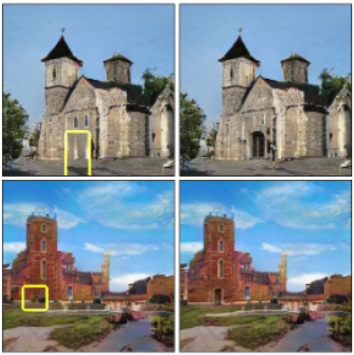}}
  \newline
  \subcaptionbox{Activation Max \cite{nguyen2016multifaceted}}{\includegraphics[scale=0.34]{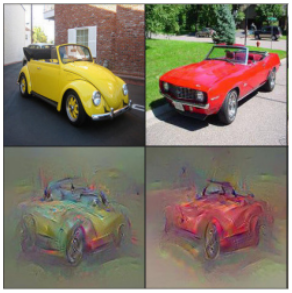}}
  \end{subfigure}
  \begin{subfigure}[b]{.23\linewidth}
  \centering
  \subcaptionbox{AOG \cite{zhang2017growing}}{\includegraphics[scale=0.32]{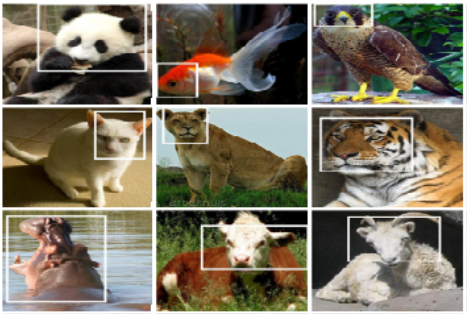}}
  \newline
  \subcaptionbox{SGR \cite{liang2018symbolic}}{\includegraphics[scale=0.37]{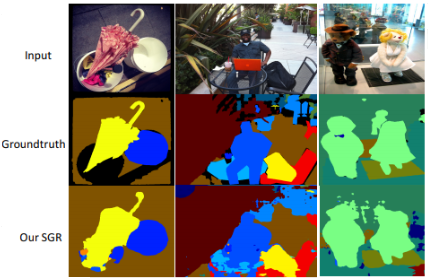}}
  \end{subfigure}
  \begin{subfigure}[b]{.27\linewidth}
  \centering
  \subcaptionbox{Cell Activation \cite{karpathy2015visualizing}}{\includegraphics[scale=0.33]{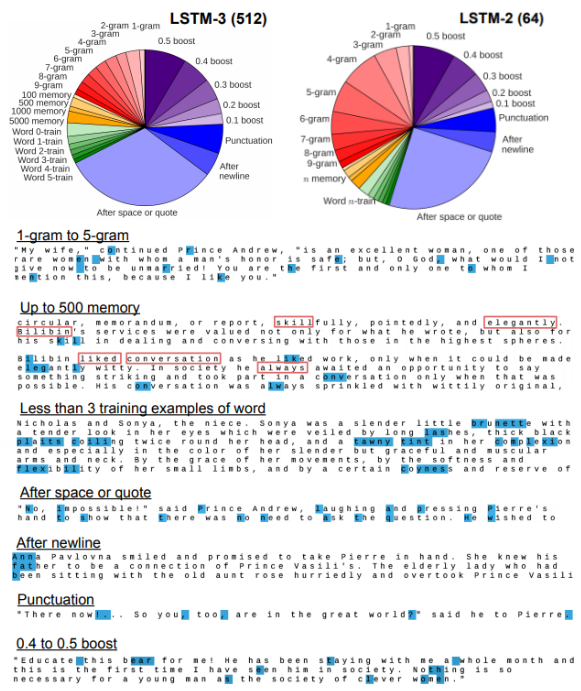}}
  \end{subfigure}
  \begin{subfigure}[b]{.23\linewidth}
  \centering
  \subcaptionbox{Data-flow graphs \cite{wongsuphasawat2018visualizing}}{\includegraphics[scale=0.33]{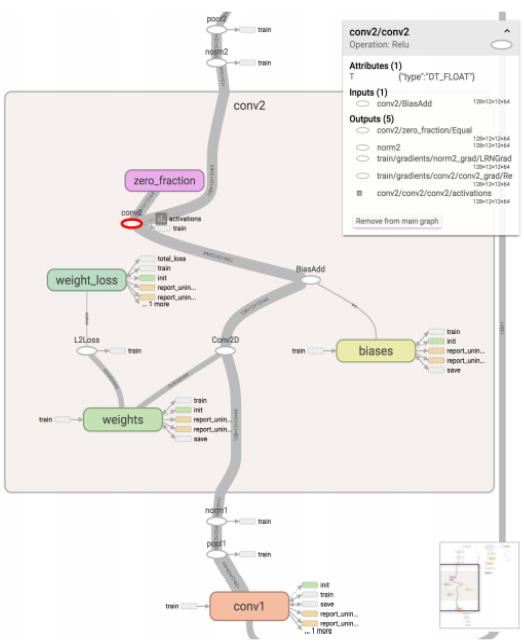}}
  \end{subfigure}
  \caption{Examples of miscellaneous visual explanations generated by methods for explainability for neural networks.}
  \label{fig:neural_networks_miscellaneous}
\end{minipage}
\end{figure}

\begin{figure}[!ht]
\begin{minipage}{\textwidth}
\centering
  \subcaptionbox{DT extraction \cite{frosst2017distilling}}
    {\includegraphics[scale=0.41, align=c]{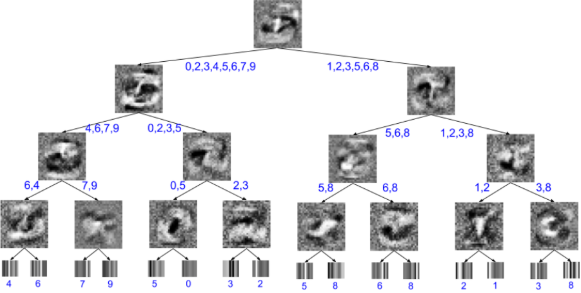}}
  \subcaptionbox{Interval Propagation \cite{palade2001interpretation}}
    {\includegraphics[scale=0.41, align=c]{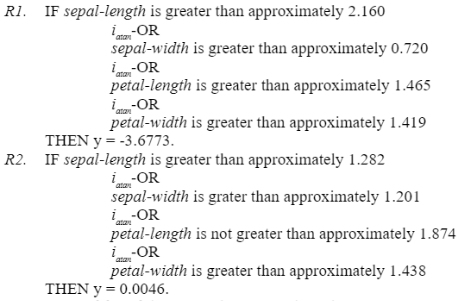}}
  \subcaptionbox{Word Importance Scores \cite{murdoch2017automatic}}
    {\includegraphics[scale=0.42, align=c]{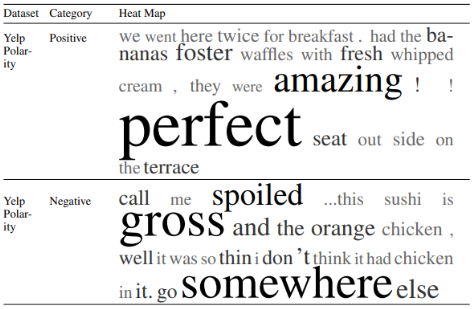}}
  \caption{Examples of rule-based explanations generated by methods for explainability for neural networks and visualized as (a) decision trees, (b) list of rules or (c) by showing the most relevant input.}
  \label{fig:neural_networks_rules}
\end{minipage}
\end{figure}

\begin{figure}[!ht]
\begin{minipage}{\textwidth}
\centering
  \begin{subfigure}[b]{.52\linewidth}
  \centering
  \subcaptionbox{InterpNET \cite{barratt2017interpnet}}{\includegraphics[scale=0.39]{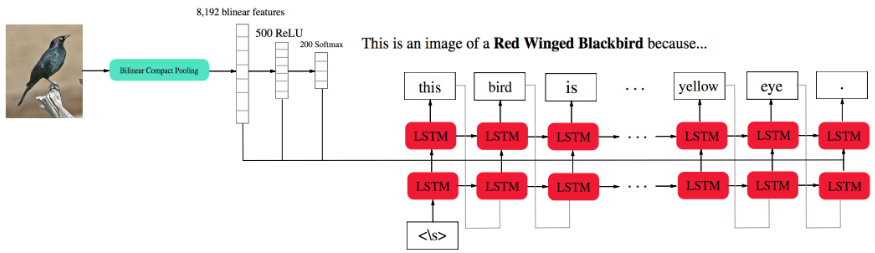}}
  \newline
  \subcaptionbox{Contextual Importance and Utility \cite{framling1996explaining}}{\includegraphics[scale=0.41]{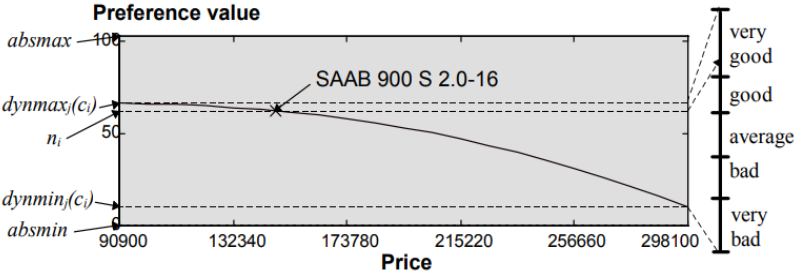}}
  \end{subfigure}
  \begin{subfigure}[b]{.47\linewidth}
  \centering
  \subcaptionbox{SVCCA \cite{raghu2017svcca}}{\includegraphics[scale=0.39]{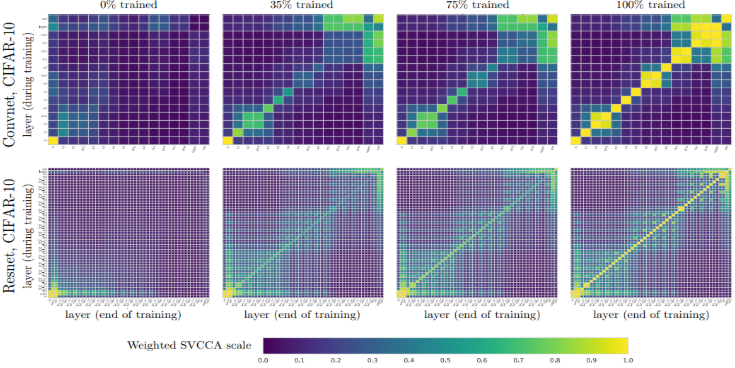}}
  \newline
  \subcaptionbox{CAVs \cite{kim2018interpretability}}{\includegraphics[scale=0.39]{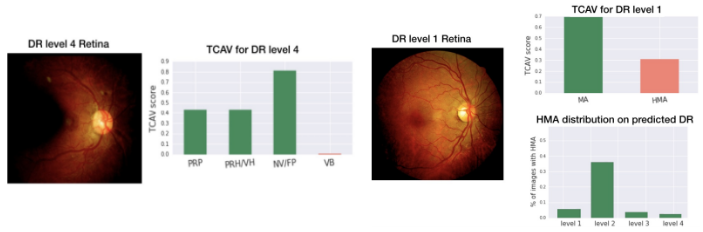}}
  \end{subfigure}
  \caption{Examples of textual (a) and numerical (b-d) explanation generated by a method for explainability for neural networks.}
  \label{fig:neural_networks_textual_numeric}
\end{minipage}
\end{figure}

\begin{figure}[!ht]

\begin{minipage}{0.99\textwidth}
    \centering
    \includegraphics[scale=0.60, align=c]{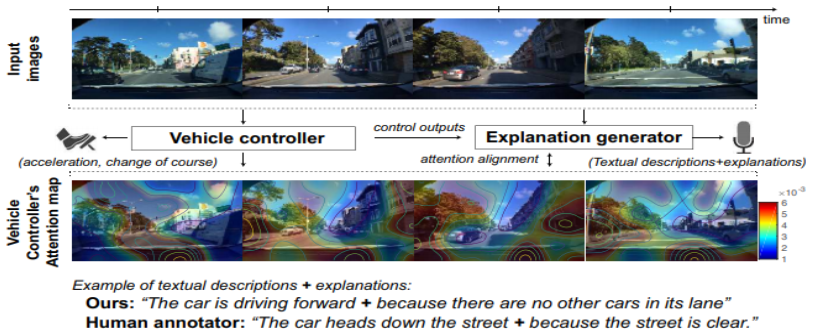}
    \subcaption[]{Attention Alignment \cite{kim2018textual}}
\end{minipage}    
\begin{minipage}{0.49\textwidth}
   \centering
   \includegraphics[scale=0.45, align=c]{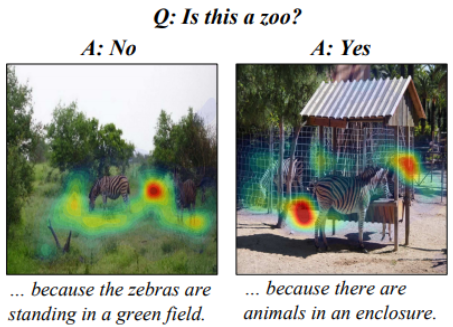}
   \subcaption[]{PJ-X \cite{park2018multimodal}}
\end{minipage}    
\begin{minipage}{0.49\textwidth}
    \centering
    \includegraphics[scale=0.45, align=c]{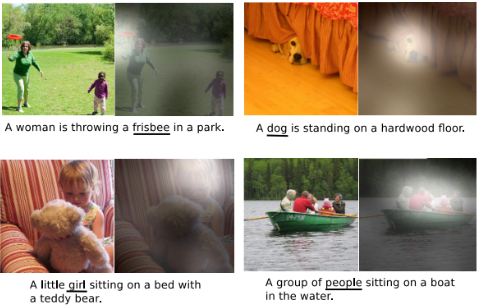}
    \subcaption[]{Attention Mechanism \cite{xuk2015show}}
\end{minipage}   
  \caption{Examples of mixed explanations, consisting of combinations of images and texts, generated by the following methods for explainability for neural networks.}
  \label{fig:neural_networks_mixed}

\end{figure}

A considerable portion of the reviewed scientific articles about new methods for explainability is focused on interpreting deep neural networks (DNNs). This is not surprising giving the momentum of Deep Learning. Most of these methods produce visual explanations (table \ref{tab:neural-networks-visual}), mostly in the form of salient masks and scatter-plots (figure \ref{fig:neural_networks_masks_plots}), some as other visual aids (figure \ref{fig:neural_networks_miscellaneous}), rules (table \ref{tab:neural-networks-rule} and figure \ref{fig:neural_networks_rules}), textual and numerical explanations (table \ref{tab:neural-networks-textual-numeric} and figure \ref{fig:neural_networks_textual_numeric}) or a combination of them (table \ref{tab:neural-networks-mixed} and figure \ref{fig:neural_networks_mixed}).

\subsubsection{Visual explanations as salient masks}
CLass-Enhanced Attentive Response (CLEAR) \cite{kumar2017explaining} produces attention maps for image classification applications by back-propagating the activation values of the output layer. CLEAR was designed to return the attentive regions responsible for the prediction, along with their attentive levels to understand their influence and the dominant output class associated with these regions.
DeepResolve \cite{liu2017visualizing} and GradCam \cite{selvaraju2017grad} are two gradient ascent-based methods. DeepResolve computes and visualizes intermediate layer feature maps that summarize how a network combines elemental layer-specific features to predict a specific class. GradCam instead uses the gradients of any target concept (say `dog' for instance) flowing into the final convolutional layer to generate a heat-map highlighting the influential regions in the image for predicting that concept. Heat-maps are generated by the last convolutional layer because the fully-connected layers do not retain spatial information and it is expected that it has the best compromise between high-level semantics and detailed spatial information. 
Stacking with Auxiliary Features (SWAF) \cite{rajani2017using} utilizes heat-maps generated by GradCam to interpret and improve stacked ensembles for visual question answering (VQA) tasks. VQA includes answering a natural language question about the content of an image by returning, usually, a word or phrase or, in this case, a heat-map highlighting the relevant regions for a prediction. 
Guided BackProp and Occlusion \cite{goyal2016towards} find what part of an input (pixels in images or words in questions) the VQA model focuses on while answering the question.
Guided BackProp is another gradient-based technique to visualize the activation values of neurons in different layers of CNNs. It computes the gradients of the probability scores of predicted classes but restricts negative gradients from flowing back towards the input layer, resulting in sharper images showcasing the activation. Occlusion consists of masking, or occluding, subsets of an input (either a region of the image or a word of the question), then forward propagating it through the VQA model and computing the change in the probability of the answer predicted with the original input. A similar method, Occlusion Sensitivity \cite{zeiler2014visualizing}, maps those features considered relevant in the intermediate layers of a DNN, by projecting the top nine activation values of each layer down to the input pixel space and masking the rest of the image.
Net2Vec \cite{fong2018net2vec} maps instead semantic concepts to corresponding individual DNN filter responses. It returns images that are entirely greyed out except in the region related to a semantic concept, such as for instance the area representing a door of a building. The pixels of this region generate activation values that are above a threshold, corresponding to the 99.5th percentile of the distribution of all the activation values.
Inverting Representations \cite{mahendran2015understanding} inverts the representations of images produced by the inner layers and projects them on the input image as heat-maps. A representation can be thought of as a function of the image that characterise the image information. By reconstructing an approximate inverse function, it should be possible to reproduce the representations built by the layers. This method is based on the hypothesis that the layers consider only the relevant features and discard the irrelevant differences between images (such as, for instance, illumination or viewpoint) and consists of a reconstruction problem solved by optimizing an objective function with gradient descent. \\

Similarly, Guided Feature Inversion \cite{du2018towards} generates an inversion image representation consisting of the weighted sum between the original image and another noisy background image, such as a grey-scale image with each pixel set to an average colour, a Gaussian white noise or a blurred image. The weights are calculated in such a way to highlight the smallest area that contains the most relevant features and to blur out everything else, especially things that might lead to an erroneous prediction, like objects belonging to other classes.
SmoothGrad \cite{smilkov2017smoothgrad} was designed to sharpen in two ways gradient-based sensitivity maps, which are often visually noisy as they highlight pixels that, to a human, seem randomly selected. The first approach considers an image of interest along with sample similar images. The second approach generates a perturbed version of the image of interest by adding Gaussian white noise. Both approaches generate individual saliency maps with other methods for explainability such as GradCam, for instance, and take the average of the resulting maps.
Deep Learning Important FeaTures (DeepLIFT) \cite{shrikumar2017learning} computes the importance scores of features based on the difference between the activation of each neuron to a `reference activation' value, computed by propagating a `reference input' through the network. This represents a default or neutral input, such as a white image, chosen according to the problem at hand. According to the authors, this difference-from-reference approach has two advantages over the other methods producing saliency maps: (I) it can propagate importance signals even when the gradient is zero, avoiding artifacts caused by discontinuities in the gradient and (II) it can reveal dependencies missed by other approaches because it can separately consider the effects of positive and negative contributions. Thus, the saliency maps produced by DeepLIFT contains all and only the important features that support or go against a certain prediction. Similarly, Integrated Gradients \cite{sundararajan2017axiomatic} attributes the prediction of a DNN to specific parts of the input. The attribution is measured as the cumulative sum of the gradients of the classification function representing the network calculated at all points along the straight-line path from a baseline input (a black image or an empty text, for example) to a specific input instance.\\

Feature Maps \cite{zhang2018interpretable} and Prediction Difference Analysis \cite{zintgraf2017visualizing} produce respectively feature- and heat-maps highlighting areas in an input image that gives evidence for or against a predicted class. Feature Maps utilizes a loss function that pushes each filter in a convolutional layer to encode a distinct and unique object part, exclusive of the object class under analysis. Prediction Difference Analysis instead is based on Explain \cite{robnik2008explaining}, which was designed to evaluate the contribution of a feature at a time. In this case, a feature should correspond to a pixel of the image, but the authors proposed to consider patches of pixels. The assumption is that the value of each pixel is highly dependent on the surrounding pixels. The patches are overlapping so that, ultimately, an individual pixel’s relevance is calculated as the average relevance of the different patches it was in. 
Two studies proposed variations of LRP, namely LRP with Relevance Conservation \cite{arras2017explaining} and LRP with Local Renormalization Layers \cite{binder2016layer}. LRP was used in conjunction with the Pixel-wise Decomposition methods for explaining the automated image classification process of neural networks \cite{bach2015pixel}. In both studies, the authors wanted to extend LRP to DNNs with non-linearities, such as LSTM models that have multiplicative interactions within their architecture \cite{arras2017explaining} or networks with local renormalization layers \cite{binder2016layer}. \cite{arras2017explaining} proposed a strategy to back-propagate the relevance of the neurons in the output layer back to the input layer through the two-way multiplicative interactions between lower-layer neurons of the LSTM. The algorithm sets to zero the relevance related to the gate neuron and propagate the relevance of the source neuron only. The extension of LRP proposed in \cite{binder2016layer} is based on first-Taylor expansion for non-linearities in the renormalization layers. \cite{simonyan2014deep} proposed to generate saliency maps by computing the first-order Taylor expansion of the function that links each pixel of an input image to the function, representing the neural network, that assigns a probability score to each output class. \\

Similarly, \cite{montavon2017explaining} analysed the use of Taylor decomposition for interpreting generic multi-layer DNNs by decomposing the network's output classification into the contributions of its input elements and back-propagating them from the output to the input layer, which are then visualized as heat-maps. Receptive Fields \cite{he2017deep} focused on visualizing the input patterns, called precisely receptive fields, that are most strongly related to individual neurons by reconstructing these from the highest activation values of each layer. 
PatternNet and PatternAttribution \cite{kindermans2018learning} aim at measuring the contribution of the input `signal' dimension, which is the part of the input that contains information about the output class, to the prediction as well as how good the network is at filtering out the `distractor', which is the rest of the input (like the image background). PatterNet yields a layer-wise back-projection of the estimated signal to the input space whereas PatternAttribution produces explanations consisting of neuron-wise contributions of the estimated signal to the classification scores.
Relevant Features Selection \cite{mogrovejo2019visual} automatically identifies the relevant internal features of a neural network via a two-step algorithm. First, a set of relevant layer/filter pairs are identified for every class of interest by finding those pairs that reduce at the minimum the differences between the predicted and the actual labels. This results in a relevance weight for every filter-wise response computed internally by the network. Then, an image is pushed through the network producing the class prediction and it generates a heat-map by taking into account the internal responses and relevance weights for the predicted class.
A combination of a Neural Network and Case Base Reasoning (CBR) Twin-systems was proposed in \cite{kenny2019twin}. This method maps the features' weights from the DNN to the CBR system to find similar cases from a training dataset that explain the prediction of the network of a new instance. To extract the weights of features, the authors proposed the Contributions Oriented Local Explanations (COLE) technique which is based on the premise that the feature contributions to the model's predictions are the most sensible basis to inform CBR explanations. COLE uses saliency maps methods, such as LRP and DeepLift, to estimate these contributions. This was tested on image classification problems with explanations generated in the form of similar images whose discriminating features were highlighted by saliency maps.
Compositionality \cite{li2016visualizing} consists of building the meaning of a sentence from the meanings of single words and phrases. This method is designed for visualizing compositionality in neural models trained for NLP tasks by plotting the salience value of each word as saliency maps. The salience values indicate the contribution of the words to the sentence meaning. For instance, the word `hate' and `boring' in the phrase `I hate the movie because the plot is boring' can be considered the two most relevant ones in a sentiment analysis problem.
The OpenBox method \cite{chu2018exact} computes exact and consistent interpretations for the family of Piecewise Linear Neural Networks (PLNN) by transforming them into a mathematically equivalent set of linear classifiers. Subsequently,  each linear classifier is interpreted by the features that dominate its prediction and the decision boundaries of each feature can be determined and visualized as scatter-plots (for numeric inputs) or heat-maps (for images).

\subsubsection{Visual explanations as scatter-plots}
The Convolutional Neural Network Interpretation method (Cnn-Inte) \cite{liu2018interpretable} uses a two-level k-means clustering algorithm to split into clusters the activation values of the neurons of hidden layers relative to each input feature. Clusters might contain the activation values of instances belonging to different classes. A random forest algorithm is then trained on each cluster. The results are visually displayed using scatter plots to show how a specific test instance is classified. \cite{aubry2015understanding} instead presented a method based on Principal Component Analysis (PCA) for analyzing the variation of features generated by CNNs to scene factors that occur in images such as object style, colour and lighting configuration. It analyzes CNN feature responses (or activation values) in the different layers by decomposing them as a linear combination of uncorrelated components associated to the different factors of variation and visualizing them into scatter-plots by using PCA.
t-Distributed Stochastic Neighbor Embedding (t-SNE) maps \cite{zahavy2016graying} analyzes Deep Q-networks (DQNs) in reinforcement learning applications, in particular for agents that autonomously learn, for instance how to play video-games. This method extracts the neural activation values of the last DQN layer and apply t-SNE for dimensionality reduction and for generating cluster plots where each dot correspond to a particular learning phase. Similarly, Hidden Activity Visualization \cite{rauber2017visualizing} uses t-SNE to visualize  the projections of the activation values of the hidden neurons as a 2D scatter-plot with points coloured according to the class of the instances originating them. The distribution of the points in the scatter-plot gives a graphical representation of the data distribution, relationships between neurons and the presence of clusters in the activation values.
Finally, TreeView \cite{thiagarajan2016treeview} consists of a scatter plot representation of a DNN via hierarchical partitioning of the feature space. Features are clustered according to the activation values of the hidden neurons in such a way that each cluster comprised of a set of neurons with similar distribution of activation values across the whole training set.

\subsubsection{Visual explanations - miscellaneous}
A few other methods use alternative visualization tools. Generative Adversarial Network (GAN) Dissection \cite{bau2019gandissect} was designed to understand the inferential process of GANs at different levels of abstraction, from each neuron to each object, and the relationship between objects, by identifying units (or groups of units) that are related to semantic classes (doors, for example). This method intervenes on them by adding or removing these objects from the image and observing how the GAN network reacts to these changes. These reactions are represented as a new version of the input image where other objects or areas of the background are modified. For instance, if a door is intentionally removed from a building, the GAN might substitute it with a window or bricks.
The Important Neurons and Patches method \cite{lengerich2017towards} analyzes the predictions of a DNN in terms of its internal features by inspecting information flow through the network. For instance, given a trained network and a test image, important neurons are selected according to two metrics, both measured over a set of perturbed images (each pixel is multiplied by a Gaussian noise): (I) the magnitude of the correlation between the neuron activation and the network output which approximates the influence of each neuron on the output, and (II) the precision of the activation of a neuron, which estimates the generalizability of the feature(s) encoded by it, by selecting those neurons whose activation values were not significantly affected by the perturbations. Given a rank of neurons, the top N are selected and their related image patches are determined by using a multi-layered deconvolutional network and enclosed in bounding boxes applied to the input image.
\cite{erhan2010understanding} and \cite{nguyen2016multifaceted,nguyen2016synthesizing} proposed two similar methods, based on Activation Maximization, which modify the input images in such a way to maximise the activation of a given hidden neuron with respect to each pixel. These modified images should provide a good representation of what a neuron is doing. \cite{hamidi2019interactive} instead presented a method to generate Activation maps which show what features activate the neurons in the penultimate layers. It is based on the idea that the final prediction of a DNN is dominated by the most highly-weighted neuron activations of this layer. 
Shifting from pictorial to textual inputs, Cell Activation Values \cite{karpathy2015visualizing} is a method of explainability for LSTMs and uses character-level language models as an interpretable test-bed for understanding the long-range dependencies learned by LSTMs by highlighting sequences of relevant characters.\\ \newpage

A group of methods that produce visual explanations in the form of graphs. The method proposed in \cite{wongsuphasawat2018visualizing} generates data-flow graphs to visualize the structure of DNNs created and trained in Tensorflow. Similarly, Explanatory Graph \cite{zhang2018interpreting} produces graphs from CNNs where each node represents a `part pattern', which correspond to the peak activation in a layer related to a part of the input, and each edge connects two nodes in adjacent layers to encode co-activation relationships and spatial relationships between patterns. \cite{liang2018symbolic} instead added to CNNs a new Symbolic Graph Reasoning (SGR) layer which performs reasoning over a group of symbolic nodes whose outputs explicitly represent different properties of each semantic in a prior knowledge graph. To cooperate with local convolutions, each SGR is constituted by three modules: a) a primal local-to-semantic voting module where the features of all symbolic nodes are generated by voting from local representations; b) a graph reasoning module that propagates information over the knowledge graph to achieve global semantic coherency; c) a dual semantic-to-local mapping module that learns new associations of the evolved symbolic nodes with local representations, and accordingly enhances local features.
Lastly, And-Or Graph (AOG) \cite{zhang2017growing} is a method to grow a semantic AOG on a pretrained CNN. An AOG is a graphical representation of the reduction of problems (or goals) to conjunctions (AND) and disjunctions (OR) of sub-problems (or sub-goals). The AOG is used for parsing the part of the input images which corresponds to a semantic concept and the output explanation consists of the input image where the semantic part is included in a bounding box. 
Many scholars studied ways to exploit the visual explanatory tools, described so far, to create interactive interfaces for the lay audience. For example, \cite{olah2018building} studied the usage of saliency maps as the building blocks of interactive interfaces to explain the inferential logic of CNNs. ActiVis \cite{kahng2018cti} is an interactive visualization system for DNNs that unifies instance- and subset-level inspection by using flowcharts that show how neurons are activated by user-specified instances or instance subsets. Deep Visualization Toolbox \cite{yosinski2015understanding} is based on two visualization tools. The first one depicts the activation values produced, while processing an image or video, on every layer of a trained CNN as heat-maps. The second tool modifies the input images via regularised optimization methods to enable a better visualization of the learned features by individual neurons at every layer. Deep View (DV) \cite{zhong2017evolutionary} measures the evolution of a DNN by using two metrics that evaluate the class-wise discriminability of the neurons in the final layer and the output feature maps. iNNvestigate \cite{alber2019innvestigate} compares different methods for explainability, namely PatternNet, PatternAttribution and LRP. LSTMVis \cite{strobelt2018lstmvis} is a visual analysis tool for recurrent neural networks, LSTM in particular, that facilitates the understanding of their hidden state dynamics. It is based on a set of interactive graphs and heat-maps of relevant words. A user can select a range of text in the heat-maps, which results in the selection of a subset of hidden states visualized in a parallel coordinate plot where each state is a data item, time-steps are the coordinates, and the tool then matches this selection to similar patterns in the dataset for further statistical analysis. Seq2seq-Vis \cite{strobelt2018s} is similar to LSTMVis but it focuses on sequence-to-sequence models, also known as encoder-decoder models, for automatic translation of texts. Seq2seq-Vis allows interactions with trained models trough each stage of the translation process intending to identify the learned pattern, detect errors and probe the model with counterfactual scenarios.
Finally, N$^2$VIS \cite{streeter2001nvis} is an interactive visualization tool for feed-forward neural networks trained with evolutionary computation which allows end-users to adjust training parameters during adaptation and to immediately see the results of this interaction. It considers graphs representing the network topology, connection weights and activation levels for specific inputs and weight volatility to facilitate the process of understanding the inferential process of a neural network and to improve its performances in terms of efficiency and prediction accuracy.

\subsubsection{Rule-based explanations}
Several methods for explainability are focused on rule-based explanations of the inferential process of neural networks, usually in the form of IF-THEN rules. Scholars divided these methods into three classes \cite{hailesilassie2016rule,bologna2018comparison}: (I) \textit{decompositional} methods work by extracting rules at the level of hidden and output neurons by analysing the values of their weights, (II) \textit{pedagogical} methods treat an underlying neural network as a black-box and the rule extraction consists of mimicking the function computed by the network; weights are not subjected to analysis, and (III) \textit{eclectic} methods that are a combination of the decompositional and pedagogical ones.\\

Regarding the decompositional methods, Discretizing Hidden Unit Activation Values by Clustering \cite{setiono1995understanding} generates IF-THEN rules by clustering the activation values of hidden neurons and replacing them with the cluster's average value. The rules are extracted by examining the possible combinations in the outputs of the discretised network. Similarly, Neural Network Knowledge eXtraction (NNKX) \cite{bondarenko2017classification} produces binary decision trees from multi-layered feed-forward sigmoidal artificial neural networks by clustering the activation values of the last layer and propagating them back to the input to generate clusters.
Interval Propagation \cite{palade2001interpretation} is an improved version of Validity Interval Analysis (VIA) \cite{thrun1995extracting} to extract IF-THEN crisp and fuzzy rules. VIA consists of finding a set of validity intervals for the activation range of each unit (or a subset of units) such that the activation values of a DNN lie within these intervals. The precondition of each extracted rule is given by a set of validity intervals and the output is a single target class. According to \cite{palade2001interpretation}, VIA has two shortcomings: it fails sometimes to decide whether a rule is compatible or not with the network and the intervals are not always optimal. Interval Propagation overcomes these limitations by setting intervals to either the input or output and feed- or back-propagating them through the network. However, this method has still a drawback. Some neural networks require a big number of crisp rules to be approximated and to reach similar performances in terms of prediction accuracy. Then, \cite{palade2001interpretation} proposed to compact these crisp rules into fuzzy rules by using a fuzzy interactive operator which introduces the OR operators between rules.
Discretized Interpretable Multi-Layer Perceptrons (DIMLP) \cite{bologna2017characterization,bologna1998symbolic,bologna2018rule, bologna2018comparison} returns symbolic rules from Interpretable Multi-Layer Perceptrons (IMLP) which are CNNs where each neuron of the first hidden layer is connected to only an input neuron and its activation function is a step function while the remaining hidden layers are fully connected with a sigmoid activation function. In DIMPL, the step activation function becomes a staircase function that approximates the sigmoid one. The rule extraction is performed after a max-pool layer by determining the location of relevant discriminative hyperplanes, which are the boundaries between the output classes. Their relevance corresponds to the number of points passing through each hyperplane as they move to a different class. An example of a ruleset generated with DIMLP from a neural network with thirty neurons, represented as $x_i$ with $i=1, \ldots, 30$, in a unique hidden layer and three output neurons is: Rule 1 - $(\neg x_3)\ (\neg x_8)\ (x_{17} > 0.0061)\ (x_{19} < 0.151)\ (x_{21} > 0.065) \quad Class\_1$, \quad Rule 2: $(x_{17} > 0.0061)\ (x_{21} < 0.065) \quad Class\_2$, \quad Default: $Class\_3$.
Rule Extraction by Reverse Engineering (RxREN) \cite{augasta2012reverse} relies on a reverse engineering technique to trace back input neurons that cause the final result, whilst pruning the insignificant ones, and to determine the data ranges of each significant neuron in respective classes. The algorithm is recursive and generates hierarchical rules where conditions for discrete attributes are disjoint from the continuous ones. Rule Extraction from Neural Network using Classified and Misclassified data (RxNCM) \cite{biswas2017rule} is a modification of RxREN.  
It incorporates also the input instances correctly classified in the range determination process, not only the misclassified ones as done by RxREN.
Most of the rule-based methods for explainability are monotonic, that means they produce an increasing set of rules, thus the prepositions that can be derived. However, sometimes adding new rules might lead to the invalidation of some conclusion inferred by other rules, as in \cite{garcez2001symbolic} where a method that captures non-monotonic symbolic rules coded in the network was presented.
The rule extraction algorithm starts by partially ordering the vectors of a training dataset according to the activation values of the output layer. Then, it determines the minimum input point that activates an output neuron and creates a rule whose antecedents are based on the feature values of the selected instance. Thus, the expected set of rules has the following form: $L_1,\ldots,L_n, \sim L_{n+1}, \ldots, \sim L_m \to L_{m+1}$ where $L_i (1 \leq i \leq m)$ represents a neuron in the input layer, $L_{m+1}$ represents a neuron in the output layer, $\sim$ stands for default negation and $\to$ means causal implication.
Finally, \cite{frosst2017distilling} and \cite{zhang2019interpreting} proposed two algorithms that extract DTs from the weights of a DNN. The former method produces a soft DT trained by stochastic gradient descent using the predictions of a neural network and its learned filters to make hierarchical decisions on where to split the data and how to create the paths from the root to the leaves. The latter, which is designed only for image classification tasks, aims at explaining an underlying CNN semantically, meaning that the nodes of the tree should correspond to parts of the objects that can be named. Nodes near the root should correspond to parts shared by many images (such as the presence of four legs in images showing animals) whereas the nodes close to the leaves should represent characteristics of minority images (a peculiar characteristic of each animal). To build such DTs, the network's filters are forced to represent object parts by a special modification of the loss function. The DT is then built on the part/filter pairs recursively on an image by image basis.\\

In regard to the pedagogical methods, Rule Extraction From Neural Network Ensemble (REFNE) \cite{zhou2003extracting} extracts symbolic rules from instances generated by neural network ensembles. The algorithm randomly selects a categorical attribute and checks if there is a value satisfying the condition that all the instances possessing such a value fall into the same class. If the condition is satisfied, a rule is created with the value as antecedent; otherwise, the algorithm selects another categorical attribute and examines all the combinations of the two attributes. When all the categorical attributes are analysed, continuous attributes are considered and the process terminates when no more rules can be created. Rules are limited to only three antecedents. Continuous attributes are discretised and a fidelity evaluation mechanism checks that this process does not compromise the relationship between the attribute and the output classes. An alternative method to extract IF-THEN rules from neural network ensembles, called C4.5Rule-PANE \cite{zhou2003medical}, uses the C4.5 rule induction algorithm. After a neural network ensemble was trained on a dataset, the original labels of the training dataset are replaced by those predicted by the ensemble. Subsequently, C4.5Rule-PANE extracts a ruleset from the modified dataset to mimic the inferential process of the ensemble.
The DecText method \cite{boz2002extracting} also extracts high fidelity DTs from a DNN. It sorts input instances by increasing the value of a feature and split an input dataset by placing the cutpoint at the midpoint of the range. Then, four splitting algorithms check the partitions such created and choose the best ones according to four criteria. The first, SetZero, chooses the most discriminative features of the target variable. The other three, SSE, ClassDiff and Fidelity, respectively select the feature which maximizes the possibility that a single class dominating each partition is created, the quality of the partition and the fidelity between the DNN and the tree. According to the authors, these algorithms have comparable, if not better, prediction performance that a tree extraction technique based on entropy gain split. TREPAN \cite{craven1994using, craven1996extracting} induces a DT that, like DecText, maintain a high level of fidelity to a DNN while being comprehensible and accurate. It queries an underlying network to determine the predicted class of each instance and selects the splits for each node of the tree by using the `gain ratio criterion' and by considering the previously selected splits that lie on the path from the root to that node as constraints.
Tree Regularization \cite{wu2018beyond} consists of a penalty function of the parameters of a differentiable DNN which favours models whose decision boundaries can be approximated by small binary DTs. It finds a binary DT that accurately reproduces the network's prediction and measures its complexity as the `average decision path length'. It then maps the parameter vector of each candidate network to an estimate of the average-path-length and chooses the shortest one.
Word Importance Scores \cite{murdoch2017automatic} visualizes the importance of specific inputs for determining the output of a LSTM. By searching for consistently important phrases and calculating their importance scores, the method extracts simple phrase patterns consisting of one to five words. To concretely validate these patterns, they are inputted to a rule-based classifier which approximates the performance of the original LSTM.
Iterative Rule Knowledge Distillation \cite{hu2016harnessing} and Symbolic Logic Integration \cite{tran2017unsupervised} are the only ante-hoc methods producing rule-based explanations for DNNs. The former combines DNNs with declarative first-logic rules to allow integrating human knowledge and intentions into the networks via an iterative distillation procedure that transfers the structured information of logic rules into the weights of DNNs. This is achieved by forcing the network to emulate the predictions of a rule-based teacher model and evolving both models throughout the training.
The latter instead encodes symbolic knowledge in an unsupervised neural network by converting background knowledge, in the form of propositional IF-THEN rules and first-order logic formulas, into confidence rules which can be represented in a Restricted Boltzmann Machine.

\subsubsection{Textual and numerical explanations}
Textual and numerical explanations are jointly less used than other methods for explainability. 
InterpNET \cite{barratt2017interpnet} makes use of a DNN to generate natural language explanations of classifications done by external models and produces statements like ``This is an image of a Red Winged Blackbird because...''. The explanations are built upon the activation values of this network.
Most-Weighted-Path, Most-Weighted-Combination and Maximum-Frequency-Difference \cite{garcia2019human} are three methods for explainability that generates textual statements.
Most-Weighted-Path starts from the output neuron and selects the corresponding input passing, layer-by-layer, via the neuron connected with the highest weight. Then, it auto-generates a natural language explanation indicating the most relevant feature for predicting the output category. Most-Weighted-Combination works similarly, but it selects the two most-weighted input features. Maximum-Frequency-Difference retrieves from the training dataset the most similar cases for each instance, then perform the difference between the percentages of cases that have the same output and those with different output. The explanation is generated according to the input with the highest difference and it is a statement like ``the smart kitchen estimates that you are sad because you are eating chocolate, which is 50\% more frequent in this emotional state than people in other emotional states''.
Rationales \cite{lei2016rationalizing} is a method for justifying the predictions made by DNNs in natural language processing tasks, such as sentiment analysis, by extracting pieces of the input text as justifications, or rationales. These rationales must contain words that are interpretable and lead to the same prediction as to the entire input text. These words are selected via the combination of a `rationale generator' function, which works as a tagging model assigning to each word a binary tag saying whether it must be included in the rationale, and an `encoder' function that maps a string of words to a target class.
Relevance and Discriminative Loss \cite{hendricks2016generating,hendricks2018grounding} generates textual explanations for an image classifier like ``The bird in the photo is a White Pelican because...''. It consists of a CNN that extracts visual features from the images, such as colours and object parts, and two LSTMs which produce a description of each image conditioned on visual features. The training process aims at reducing two loss functions, called respectively `Relevance' and `Discriminative', which assure that the generated sentences are both image relevant and category-specific. \\

A few methods for explainability produce pure numerical explanations. Concept Activation Vectors (CAVs) \cite{kim2018interpretability} separates the activation values of a hidden layer relative to the instances belonging to a class of interest from those generated by the remaining part of the dataset. Then it trains a binary linear classifier to distinguish the activation values of the two sets and compute directional derivatives on this classifier to measure the sensitivity of the model to changes in inputs towards the class of interest. This is a scalar quantity calculated over the whole dataset for each class.
Probe \cite{alain2017understanding} consists of a linear classifier is fitted to a single feature learned by each layer of a DNN to predict the original classes. The numeric explanations consist of the predictions made by the Probes. Singular Vector Canonical Correlation Analysis (SVCCA) \cite{raghu2017svcca} returns the correlation matrix of the neurons' activation vectors, calculated over the entire dataset, of two instances of a given DNN trained separately. The first network's instance is obtained at the end of the training process, whereas the latter consists of multiple snapshots of the network during training. Every layer of the first instance is compared to every other layer of the other instance to calculate correlation factors between pairs of layers.
Causal Importance \cite{feraud2002methodology} is calculated by summing up the variations in the output when the values of a variable are perturbed instance by instance whilst all the other variables are kept fixed. Firstly, the algorithm suppresses the irrelevant variables by measuring their predictive importance via a metric based on the absolute difference between the predictions made by a DNN with the original and the perturbed variables. The network is then trained with the relevant variables only and data are clustered according to their hidden layer representation. This is done by training an unsupervised Kohonen map on the matrix containing the activation values of each pair neuron/input instance. Finally, causal importance is measured on a cluster by cluster basis, meaning that it is calculated only on the instances belonging to the cluster under analysis.
\cite{framling1996explaining} proposed to measure the Contextual Importance and Contextual Utility of input on the output variable. The former metric is the ratio between the range of the output values covered by varying a variable throughout its own range of values and the whole output space. For example, a neural network was trained to predict the price of a car over a set of variable, among which there is the engine size. By varying the engine size alone, the price varies only within a certain range. Contextual Utility represents the position of the actual output within Contextual Importance. So the price of cars with big engines, produced by the same manufacturer, are towards the upper end of the manufacturer's price range. The ideal values for these two metrics are domain-dependent. They are designed just to help end-users understand where each variable and instance lie within the input space, but they can be used to generate rule-based or textual explanations by structuring the domain in intermediate concepts which attach a positive or negative outlook to certain subsets of the output space. A certain price range can be deemed very good for a manufacturer and all its cars in that range can be considered as a deal.
Finally, REcurrent LEXicon NETwork (RELEXNET) \cite{clos2017towards} combines the transparency of lexicon-based classifiers with the accuracy of recurrent neural networks. Lexicons are linguistic tools for classification and feature extraction which take the form of a list of terms weighted by their strength of association with a given class. RELEXNET uses lexicons as inputs of a naive gated recurrent neural network which returns the probability that the input belongs or not to a certain class.

\subsubsection{Mixed explanations}
To help end-user interpret a model, some scholars have proposed the use of multiple types of output formats for explanations.
The Attention Alignment method \cite{kim2018textual} produces explanations in the form of attention maps. These maps highlight parts of a scene that mattered to a control DNN utilized in self-driving cars in combination with a perception DNN. The perception DNN combines the data received from cameras and other sensors, like radars and infrared, to `understand' the environment and to generate manoeuvring commands like steering angles or braking. The control DNN is trained to identify the presence of specific objects, such as road signs, and obstacles like pedestrians and bikers, that influence the output of the perception network. Attention Alignment consists of an attention-based video-to-text algorithm that produces textual explanations of the model predictions such as ``The car heads down the street because it is clear.'' 
Similarly, Pointing and Justification Model (PJ-X) \cite{park2018multimodal} and Image Caption Generation with Attention Mechanism \cite{xuk2015show} are two multi-modal methods for explainability, designed for VQA tasks, that provides joint textual rationale generation and attention-map visualization. The attention-maps are extracted from a CNN, which performs the object recognition in images, whereas the textual justifications are produced by an LSTM network as image captions. The word(s) in the caption related to the attention region is underlined.
According to \cite{park2018multimodal}, the two explanations support each other in achieving high quality. The visual explanations help to generate better textual explanations which lead to better visual pointing.
Image Caption Generation with Attention Mechanism is based on two algorithms: (I) a `soft' deterministic attention mechanism trainable by standard back-propagation methods and (II) a `hard' stochastic attention mechanism trainable by maximizing an approximate variational lower bound. The word(s) in the caption related to the attention region is/are underlined.
\cite{mayr2018regular} and \cite{omlin1996extraction} proposed two methods for replacing a DNN with a deterministic finite automaton that can be visualized as a graph where each node represent a cluster of values in the output space and the edges represent the presence of shared patterns in a network's internal layers between these clusters.
Lastly, the method in \cite{tamajka2019transforming} uses prototypes to explain the prediction of a new instance. The prototypes are selected according to the activation values of hidden neurons related to training data. For a new observation, it is possible to foresee and justify its prediction by identifying the three most similar training samples based on cosine distance of their hidden activation values.\\

\subsection{Model-specific methods for explainability related to rule-based systems}
\begin{figure}[!ht]
\begin{minipage}{\textwidth}
\centering
  \begin{subfigure}[b]{.32\linewidth}
  \centering
  \subcaptionbox{BRL \cite{letham2015interpretable}}{\includegraphics[scale=0.31]{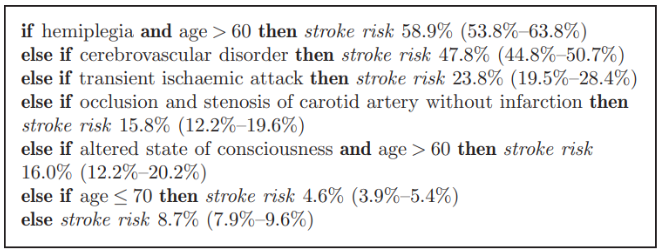}}
  \newline
  \subcaptionbox{ExpliClas \cite{alonso2019explainable}}{\includegraphics[scale=0.42]{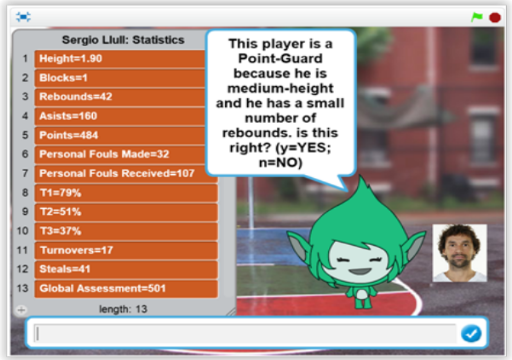}}
  \end{subfigure}
    \begin{subfigure}[b]{.32\linewidth}
  \centering
  \subcaptionbox{Fingrams \cite{pancho2013fingrams}}{\includegraphics[scale=0.36]{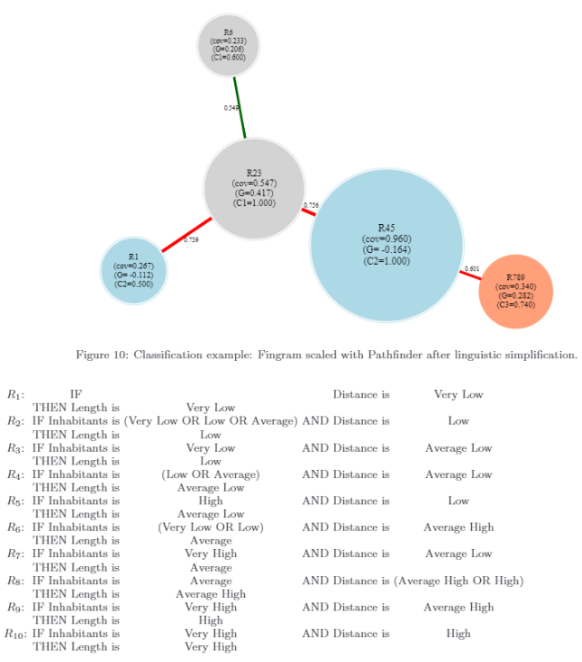}}
  \end{subfigure}
  \begin{subfigure}[b]{.31\linewidth}
  \centering
  \subcaptionbox{Interpretable Decision Set \cite{lakkaraju2016interpretable}}{\includegraphics[scale=0.33]{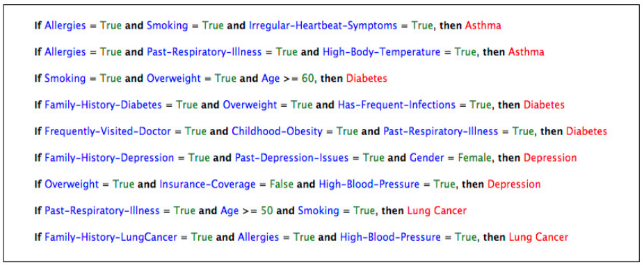}}
  \newline
  \subcaptionbox{Fuzzy Inference Systems \cite{keneni2019evolving}}{\includegraphics[scale=0.38]{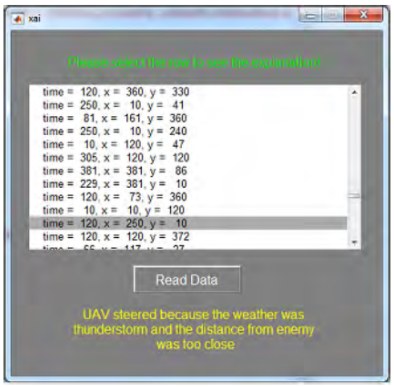}}
  \end{subfigure}
  \caption{Examples of explanations generated by methods for explainability for rule-based inference systems.}
  \label{fig:rule_based_approaches}
\end{minipage}
\end{figure}

Explainable Artificial Intelligence was ignited by the interpretability problem of machine learning, in particular of the deep learning models. However, the problem of explainability exists even before the advent of neural networks.
Many rules-based learning approaches already existed and the majority of these were interpreted with ante-hoc methods that act during the model training stage to make it naturally explainable (see table \ref{tab:rule-based} and figure \ref{fig:rule_based_approaches}).
An Ant Colony Optimization (ACO) algorithm was proposed in \cite{otero2016improving} to create interpretable rules. It follows a sequential covering strategy, one-rule-at-a-time or also known as separate-and-conquer, to generate unordered sets of IF-THEN classification rules which can be inspected individually and independently from the others, thus they are easier to be interpreted. At each step, ACO creates a new unordered set of rules and compare it with those of previous iterations. If the new set contains fewer rules or has a better prediction accuracy, it replaces the previous one. Experiments run on thirty-two publicly available datasets showed that ACO gave the best results in terms of both predictive accuracy and model size, outperforming state-of-the-art rule induction algorithms with statistically significant differences.
Another method based on an ACO algorithm, called AntMinter+ \cite{verbeke2011building}, constructs monotonic rulesets by allowing the inclusion of domain knowledge via the definition of a directed acyclic graph representing the solution space. The nodes at the same depth in the graph represent the splitting values related to an input variable; the edges represent which values of the following variable can be reached from a node. AntMinter+ uses an iterative max-min ant system to construct a set of IF-THEN rules, starting from an empty set. A rule represents a path from the start to the end nodes of the graph. The algorithm stops adding rules when either a predefined percentage of training points is covered or when the addition of new rules does not improve the performance of the classifier. AntMinter+ can be combined with a non-linear SVM, in a method called Active Learning Based Approach (ALBA), to generate comprehensible and accurate rule-based models. 
Exception Directed Acyclic Graphs (EDAGs) \cite{gaines1996transforming} is an empirical induction tool that generates rules from the knowledge base of expert systems to create comprehensible knowledge structures in the form of graphs where nodes are premises, some of which have attached conclusions, leaves are conclusions and edges represent exceptions to some node. The `meaning' of each node can be easily determined by following its path back to the root and by inspecting its child nodes, whilst the rest of the graph is irrelevant.\\

The Interpretable Decision Set \cite{lakkaraju2016interpretable} and the Bayesian Rule Lists (BRL) \cite{letham2012building,letham2013interpretable,letham2015interpretable} are two methods that creates unordered sets of IF-THEN rules. Interpretable Decision Set is based on an objective function that simultaneously optimizes accuracy and interpretability by learning short and non-overlapping rules that cover the whole feature space and pay attention to small but important classes. BRL produces a posterior multinomial distribution over permutations of rules, starting from a large set of possible rules, to assess the probability of predicting a certain label from the selected rules. The prior is the Dirichlet distribution and the permutation that maximises the posterior is included in the final decision set. The Bayesian Rule Sets (BRS) method \cite{wang2016bayesian,wang2017bayesian} is similar to BRL but it uses different probabilities, with the posterior as a Bernoulli distribution, and the prior a Beta distribution whose parameters can be adjusted by end-users to guide BRS toward more interpretable solutions by specifying the desired balance between size and length of rules.
First Order Combined Learner (FOCL) \cite{pazzani1997comprehensible} inductively constructs a set of rules in terms of predicates used to describe examples. Each clause body consists of a conjunction of predicates that cover some positive and no negative examples. The rules are displayed in a tree where the nodes are the predicates, the edge are the conjunctions and the leaves are the conclusions.
Non-monotonic argumentation-based approaches for increasing explainability and dealing with conflictual information were proposed in \cite{rizzo2019inferential,rizzo2018qualitative, zeng2018building}.
They are based upon the concepts of defeasible arguments, in the form of rules, each composed by a set of premises, an inference rule and a conclusion as well as the notion of attacks between arguments to model conflictuality and the retraction of a final inference. Argumentation-based approaches are believed to have a higher explainability as the notions of arguments and conflictuality are common to the way human reason.
Four methods based on fuzzy reasoning to generate interpretable sets of rules that clearly show the dependencies between inputs and outputs were presented in \cite{ishibuchi2007analysis, jin2000fuzzy,pierrard2018learning, wang2011building}. 
Both \cite{ishibuchi2007analysis, wang2011building} examine the interpretability-accuracy tradeoff in fuzzy rule-based classifiers. A multiobjective fuzzy Genetics-Based Machine Learning (GBML) algorithm, analyzed by \cite{ishibuchi2007analysis}, is implemented in the framework of evolutionary multiobjective optimization (EMO) and consists of a hybrid version of Michigan and Pittsburgh approaches. Each fuzzy rule is represented by its antecedent fuzzy sets as an integer string of fixed length and the resulting fuzzy rule-based classifier, consisting of a set of fuzzy rules, is represented as a concatenated integer string of variable length. Multi-Objective Evolutionary Algorithms based Interpretable Fuzzy (MOEAIF) \cite{wang2011building} consists instead of a fuzzy rule-based model engineered to classify gene expression data from microarray technologies. GBML and MOEAIF maximize the accuracy of rule sets, measured by the number of correctly classified training pattern, and minimize their complexity, measured by the number of fuzzy rules and/or the total number of antecedent conditions of fuzzy rules. 
The method in \cite{jin2000fuzzy} is based on a five-step algorithm. First, it generates fuzzy rules that cover the extrema directly from data. Second, it checks rule similarity to delete the redundant and inconsistent rules. Third, it optimizes the rule structure using genetic algorithms based on a local performance index. Fourth, it performs further training of the rule parameters using gradient-based learning method and deletes the inactive rules. Last, it improves interpretability by using regularization.
The method presented in \cite{pierrard2018learning} generates fuzzy rules by starting from a set of relations and properties, selected by an expert, of an input dataset. It then extracts the most relevant ones employing a frequent itemset mining algorithm. The authors do not provide a specific metric for evaluating the relevancy of a relation, but they suggest using ``measures like the number of relations and properties in the antecedent or the value of their support''.
Interpretable Classification Rule Mining (ICRM) \cite{cano2013interpretable} consists of a three-step evolutionary programming algorithm producing comprehensible IF-THEN classification rules, where comprehensibility is achieved by minimizing the number of rules and conditions. First, it creates a pool of rules composed of a single attribute-value comparison. Second, it utilizes evolutionary processes, designed to use only relevant attributes which are to discriminate a class from the others and improve the accuracy of the ruleset, based on the Iterative Rule Learning (IRL) genetic algorithm (also known as the Pittsburgh approach). IRL returns a rule per output class with the exception of one class which is set as default. The third step optimizes the accuracy of the classifier by maximising the product of sensitivity and specificity.
Linear Programming Relaxation \cite{malioutov2017learning, su2016interpretable} is a method for learning two-level Boolean rules in conjunctive normal form (AND-of-ORs) or disjunctive normal form (OR-of-ANDs) as a type of human-interpretable classification model. A  first version uses a generalization of a linear programming relaxation from one level to two-level rules whose objective function is a weighted combination of the total number of errors and features used in the rule. In a second version, the 0-1 classification error is replaced with the Hamming distance between the current rule and the closest rule that correctly classifies a sample instance. The main advantages for explainability of the Hamming distance is that it avoids identical clauses in the ruleset, thus repetitions, by training each clause with a different subset of input instances.\\

All the above methods were intrinsically ante-hoc, but other methods exist for post-hoc explainability.
For example, Mycin \cite{shortliffe1975computer}, probably the first method for explainability ever developed, is a backward chaining expert system based upon a knowledge-based of IF-THEN rules composed by an expert, a database of the context set of facts that satisfy the condition part of the rules, and an inference engine that interpret the rules. 
It also includes a natural language interface that allows end-users to interact with the system independently of the expert by asking English questions, and the system can respond to them by using its inference engine and performing the reasoning involved in composing an answer to them. In details, it searches for facts that match the condition part of the productions that match the action part of the question. This method allows the system to explain its reasoning and final inferences by using AND/OR trees created during the production system reasoning process, thus showing an element of explainability.
Similarly, the Sugeno-type fuzzy inference system proposed in \cite{keneni2019evolving} consists of an explicit declarative knowledge representation of rules, which are fired at the same time by a given input, and produce a final inference.
Besides this, the system includes an explanatory tool which shows a numerical representation of the input variables, the set of co-fired rules and an English statement exposing the reasoning process. In an example taken from the application to an Unmanned Aerial Vehicle (UAV) sent on a fight mission, is a statement like ``UAV aborted the mission because the weather was a thunderstorm and the distance from the enemy was too close''.
Another method, the Fuzzy Inference-Grams (Fingrams)  \cite{pancho2013fingrams} produces inference maps of sets of fuzzy rules which graphically depict the interaction between co-fired rules and gives support to detect redundant or inconsistent rules as well as it identifies the most significant ones, by using network scaling methods that simplify the maps while maintaining their most important relations. Fingrams also quantifies the comprehensibility of the ruleset, measured as the proportion of the co-fired rules. The assumption is that the larger the number of rules which are co-fired for a given input, the smaller the comprehensibility of the ruleset.
ExpliClas \cite{alonso2019explainable} is a visual interface designed to explain, in an instance-based manner, rule-based classifiers (such as those algorithms extracting DTs from data, like C4.5 or CART, for instance) with visual and textual explanations. The rules are graphically displayed as DTs and a natural language generation approach returns textual explanations of the fired rules.
ExpliClas was tested on in the context of recognising the role of basketball players from some physical characteristics and game statistics. An example of textual explanation produced in this case is ``The player is a Point-Guard because he is medium-height and he has a small number of rebounds.'' and it is accompanied by a graph of the DT structure and bar-charts of the player's information.

\subsection{Other model-specific methods for explainability}
\begin{figure}[!ht]
\begin{minipage}{\textwidth}
\centering
  \begin{subfigure}[b]{.49\linewidth}
  \centering
  \subcaptionbox{Decision tree extraction \cite{ferri2002ensemble}}{\includegraphics[scale=0.48]{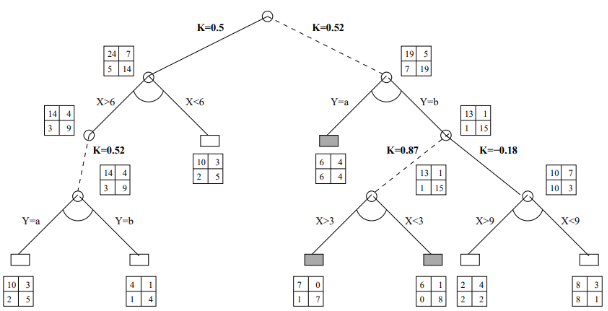}}
  \newline
  \subcaptionbox{Self-Organizing Maps \cite{hamel2006visualization}}{\includegraphics[scale=0.48]{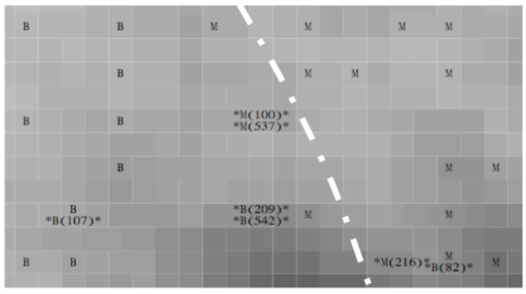}}
  \end{subfigure}
  \begin{subfigure}[b]{.49\linewidth}
  \centering
  \subcaptionbox{Nomograms \cite{movzina2004nomograms}}{\includegraphics[scale=0.48]{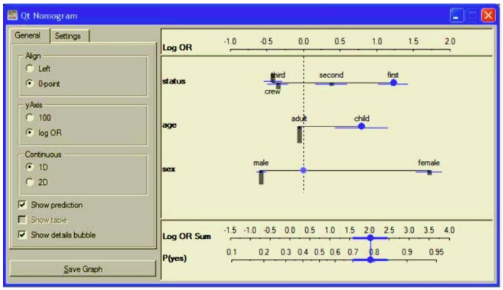}}
  \newline
  \subcaptionbox{PSA \cite{timmer2017two}}{\includegraphics[scale=0.48]{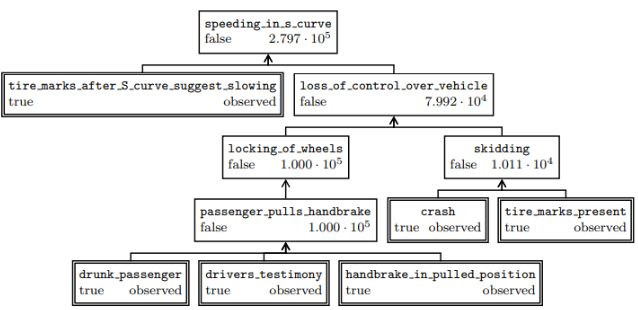}}
  \end{subfigure}
  \caption{Examples of explanations generated by methods specifically designed to interpret ensembles (a), Support Vector Machines (b-c) and bayesian networks (d).}
  \label{fig:other_model_sepcific}
\end{minipage}
\end{figure}

Scholars proposed other methods for explainability that are not strictly based on neural networks or rule-based classifiers (see table \ref{tab:data-driven} and figure \ref{fig:other_model_sepcific}). 

\subsubsection{Ensembles.} Some methods were designed to interpret the logic followed by ensembles. \cite{tolomei2017interpretable} introduced an algorithm to tweak the input features to change the output of a tree ensemble classifier. It modifies a variable (or a set of variable) of an input instance by applying a linear shift, capped to a global tolerance value, until all the trees in the ensemble assign it to another target class. The delta between the original and the tweaked value represents the `effort', or `tweaking cost', required to move the instance into the target class and provides a measure of the sensitivity of the model to changes to that particular feature(s). This information can be used to rank the variables according to their tweaking costs and inform end-users on how a particular instance must be modified to change its output label and at what cost this can be achieved.
Three methods for extracting a single DT from ensemble models, and for generating global explanations were presented in \cite{alonso2018explainable, ferri2002ensemble, van2007seeing}. 
In detail, \cite{ferri2002ensemble} uses the solution obtained from combining the several hypotheses (or models) of the ensemble as an oracle, and it selects the single hypothesis that is most similar to the oracle. The similarity is measured according to three formal metrics: `$\theta - measure$' which determines the probability that both classifiers agree, `$\kappa - measure$' which assesses the probability that two classifiers agree by chance and `$Q-measure$' which assigns values between 0 and 1 to classifiers that correctly predict the same input instances and values between -1 and 0 to classifiers that commit errors on different instances.
Instead, \cite{van2007seeing} proposed to generate a tree from the ensemble by using a divide-and-conquer algorithm analogous to C4.5. 
Similarly, \cite{alonso2018explainable} combined an opaque learning algorithm (random forest), with a more transparent and inherently interpretable algorithm (decision tree).
On one hand, the opaque algorithm represents the `oracle' which search for the most relevant output. On the other hand, a natural language generation approach is aimed at composing a textual explanation for this output which is the interpretation of the inference process carried out by the correspondent decision tree, if the output of both the learning algorithms coincides.
\cite{andrzejak2013interpretable} proposed instead a method to efficiently merge a set DTs, each trained independently on distributed data, into a single tree to overcome the lack of interpretability of the distributed models. The algorithm consists of three steps. First, each DT is converted into a ruleset where each rule replicates a path from the root to a leaf and defines a region in the output space. All the regions are disjoint and they cover the entire space. In the second phase, the regions are combined by using a line sweep algorithm which sorts the limits of each region and merges overlapping and adjacent ones. Finally, a DT is extracted from the regions with an algorithm that mimics the C5.0 and uses the values in the regions as examples. \\

A similar approach is at the basis of the Factorized Asymptotic Bayesian (FAB) inference method \cite{hara2018making} which consists of a bayesian model selection algorithm that simplified and optimized a tree ensemble.
FAB estimates the model's parameters and the optimal number of regions of the input space (ensemble methods often splits the input space into a huge number of regions) to derive a simplified model with appropriate complexity and prediction accuracy.
inTrees \cite{deng2018interpreting} extracts, prunes and selects rules from a tree ensemble. The algorithm starts from the set of all the rules in the ensemble and excludes those covering a small number of instances. At each iteration, the algorithm selects the rule with the minimum error and shorter condition, then it removes the instances satisfying this rule from the dataset and it updates the initial ruleset according to the instances left, by discarding rules that at this stage cover just a few, if not none, instances and recalculating their error.
Discriminative Patterns \cite{gao2017interpretable} aims at interpreting a random forest model that classifies sentences according to their contents by extracting a ruleset that enables interpretation and gains insight of useful information in texts which corresponds to discriminative sequential patterns of words, or sentences that determine the predicted class.
Tree Space Prototypes (TSP) \cite{tan2016tree} selects prototypes from a training dataset to explain the prediction made be ensembles of DTs and gradient boosted tree models on a new observation. 
To measure the similarity between the new instance and the prototypes, the authors proposed a metric based on the weighted average of the number of trees in the ensemble that assigns the points to the same output class to quantify the contribution of the predictions made by each DT to the overall prediction. By following the path root-to-leaf of the most relevant DT, it is possible to determine the values of the features deemed important by the tree for predicting the class of the new instance and select a prototype having the same values.

\subsubsection{Support Vector Machines.} ExtractRule \cite{fung2005rule} converts hyperplane-based linear classifiers, like SVMs, into a set of non-overlapping symbolic rules which are human-understandable because they display, in a compact format, the inferential process of the underlying classifier. An example of a rule extracted from a classifier trained to distinguish between malign and benign tumors is ``$(Cell\ Size \leq 3) \wedge (Bare\ Nuclei \leq 1) \wedge (Normal\ Nucleoli \leq 7) \implies mass\ is\ benign$''. Each rule can be seen as a hypercube in the multidimensional space generated by the input variables with edges parallel to the axis. To define these hypercubes, each iteration of this algorithm is formulated as one of two possible optimization problems. The first formulation seeks to maximize the volume covered by each rule whereas the second formulation maximizes the number of samples covered.
Important support vectors and Border classification \cite{barbella2009understanding} are two methods for providing insight into local classifications produced by a SVM. The former reports the most influential support vectors for the final classification of a particular data instance, thus determining the most similar instances to the test point that belong to the same class. As in the previous methods based on prototypes, presenting this subset helps users understand the model by leveraging on the human ability to induce principles from a few examples. The latter determines which features of a testing data instance would need to be altered (and by how much) to be classified on the separating surface between the two classes, thus providing, as in the feature tweaking method, a measure of the cost required to change a model's prediction.
SVM+Prototypes \cite{nunez2002rule} is based on a clustering algorithm to detect the prototype vectors for each class, after the decision function is determined to employ a SVM. These vectors are combined with the support vectors using geometric methods to define ellipsoids in the input space, which are later transferred to IF-THEN rules as the mathematical equations that defined the ellipsoids, so a rule looks like ``If $AX_1^2+BX_2^2+CX_1X_2+DX_1+EX_2+F \leq G$ Then $Class_1$''.
Weighted Linear Classifier \cite{caragea2003towards} generates weighted linear SVM classifiers or random hyperplanes to obtain models whose accuracy is comparable to that of a non-linear SVM classifier and whose results can be readily visualized by projecting them on separating hyperplanes and decision surfaces. These projections are considered as a sort of explanations.
A method based on Self-Organizing Maps (SOM) used to visualise SVMs was proposed in \cite{hamel2006visualization}. 
It is a specific unsupervised network aimed at investigating a high-dimensional space of data for a cluster of points by projecting these clusters onto a 2-dimensional map, but trying to preserve their topologies.
Thus, it allows visualising both data and the SVM models, providing an overall overview of the support vector decision surface which is not possible with other visualization approaches.  
\cite{jakulin2005nomograms,cho2008nonlinear,movzina2004nomograms} introduced a method for automatically generating nomograms as the graphical tool for visual explanations of the inferential mechanisms of SVM and na\"{i}ve bayesian-driven models. A nomogram is a two-dimensional diagram designed to allow approximating graphical computation of mathematical functions by showing a set of scales, one for each variable (dependent and independent) in an equation. By drawing a line connecting specific values of all the scales related to the independent variables, it is possible to calculate the value of the dependent variable from the intersection point between the line and the variable's scale. The advantages for explainability of nomograms are simplicity of presentation and clear display of the effects of individual attribute values.

\subsubsection{Bayesian and hierachical networks.} Explaining Bayesian network Inferences (EBI) \cite{yap2008explaining} produces DT rules to show how the variables of a bayesian network interact to make predictions. In detail, EBI explains the value of a target node in terms of the influential nodes in the target’s Markov blanket which include the target’s parents, children and the children’s other parents. Working backwards from the target node, EBI shows the derivation of each intermediate node and explains how missing and erroneous values are compensated by displaying these causal relationships in a DT hierarchy.
\cite{vlek2016method} instead proposed an explanation method for understanding bayesian networks in terms of scenarios. Narrative approaches to reasoning with legal evidence, for instance, are based on the formulation of alternative scenarios which are subsequently compared according to two aspects: the relations with the evidence and the quality that depends on the completeness, internal consistency and plausibility of the scenario itself. The aim is to explain the content of the bayesian network by reporting which scenarios were modelled and evaluating their evidential support and quality.
Probabilistically Supported Arguments (PSA) \cite{timmer2017two} is a two-phase method for extracting probabilistically explanatory supported arguments from a bayesian network. In the first phase, a support graph is constructed from a bayesian network representing the structure for a particular variable of interest. In the second phase, given a set of observations, arguments are built from that support graph. To do so, the algorithm defines a logical language and a set of rules built from the support graph by following its edges and nodes. The parents of a node are the rule conditions, the node itself is the rule's outcome. Only the parents supported by pieces of evidence are considered. Then, the ASPIC+ framework for structured argumentation is instantiated. Arguments can attack each other on the conclusion variable and defeat can be based on the inferential strength of the arguments which can be computed with two types of measures: `incremental measures' which assign a number to the weight of the evidence (the Likelihood Ratio is an example of these measures) and `absolute measures' which assign strength based on posterior probability, such as the posterior for instance. 
Such arguments can help interpret and explain the relationship between hypotheses and evidence that is modelled in the Bayesian network.
Contribution propagation \cite{Landecker2013interpreting} is a per-instance method for hierarchical networks that explain which components of the input were responsible (and to what degree) for its classification. The central idea is that a node was important to the classification if it was important to its parents, and its parents were important to the classification. The contribution of each input component is visualized as heat-maps.

\subsection{Self-explainable and interpretable methods}
\begin{figure}[!ht]
\begin{minipage}{\textwidth}
\centering
  \begin{subfigure}[b]{.28\linewidth}
  \centering
  \subcaptionbox{BCM \cite{kim2014bayesian}}{\includegraphics[scale=0.33]{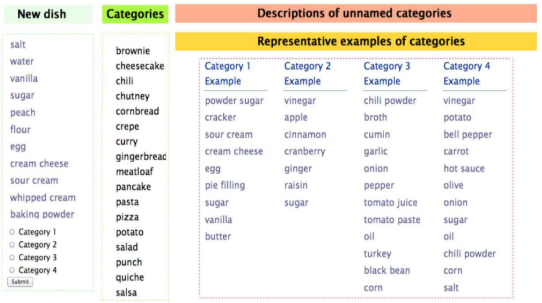}}
  \newline
  \subcaptionbox{SLIM \cite{ustun2014supersparse}}{\includegraphics[scale=0.37]{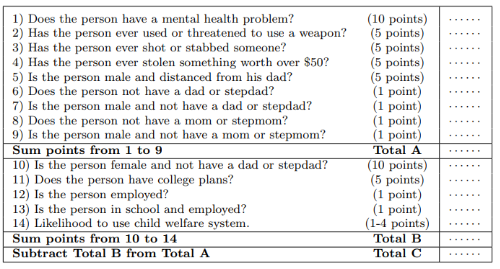}}
  \end{subfigure}
    \begin{subfigure}[b]{.33\linewidth}
  \centering
  \subcaptionbox{PSDD \cite{liang2017towards}}{\includegraphics[scale=0.38]{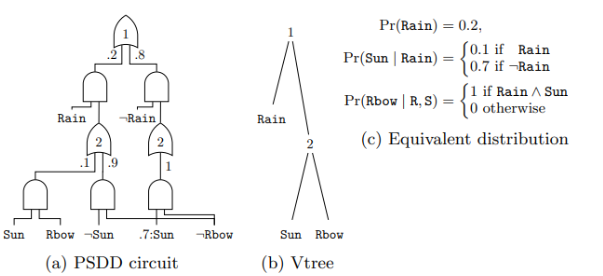}}
   \newline
  \subcaptionbox{GA$^2$Ms \cite{caruana2015intelligible}}{\includegraphics[scale=0.38]{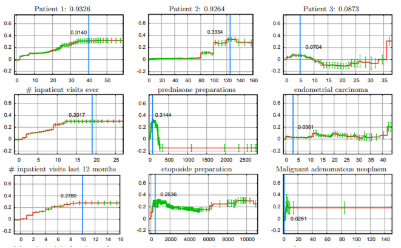}}
  \end{subfigure}
  \begin{subfigure}[b]{.31\linewidth}
  \centering
  \subcaptionbox{MGM \cite{kim2015mind}}{\includegraphics[scale=0.36]{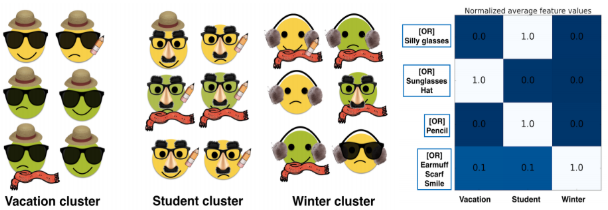}}
  \newline
  \subcaptionbox{Unsupervised Interpretable Word Sense Disambiguation \cite{panchenko2017unsupervised}}{\includegraphics[scale=0.33]{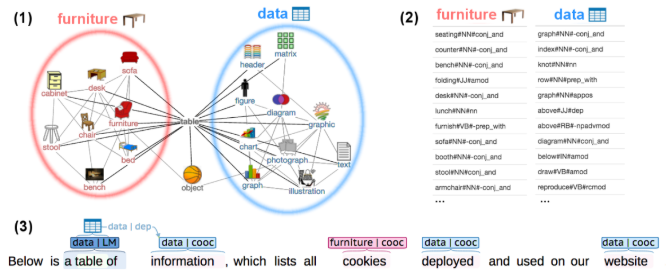}}
  \end{subfigure}
  \caption{Examples of ante-hoc methods for explainability designed to generate self-explainable models.}
  \label{fig:self_explainable}
\end{minipage}
\end{figure}

Naturally interpretable models, sometimes referred to as `white-box models', are `ante-hoc' (see table \ref{tab:white-box} and figure \ref{fig:self_explainable}). Their output format depends on their architecture and inputs format.
Bayesian Case Model (BCM) \cite{kim2014bayesian} is a method for explainability for bayesian case-based reasoning, prototype classification and clustering. BCM learns prototypes, corresponding to the observations that best represent clusters in a dataset, by performing joint inference on cluster labels, prototypes and important features.
Gaussian Process Regression (GPR) \cite{caywood2017gaussian} is a powerful, but amenable to analysis, data-driven model. GPR is a non-parametric regression algorithm, meaning that it does not make any assumption on the estimator function as linear and logistic regression algorithms do, robust to missing data and interpretable because the weights assigned to each feature provide a measure of its relevance.
Generalized Additive Models \cite{lou2012intelligible} and their extension with pairwise interactions (GA$^2$Ms) \cite{caruana2015intelligible, lou2013accurate} are linear combinations of simple models, called `shape functions', trained on a single feature (GAMs) or up to two features (GA$^2$Ms). Their simple structure allows end-user to easily understand the contribution of individual features to the predictions and to visualize them, together with the shape functions, with bar- and line-charts.
Oblique Treed Sparse Additive Models (OT-SpAMs) \cite{wang2015trading} are instances of region-specific predictive models. They divide feature spaces into regions with sparse oblique tree splitting and assign local sparse additive experts to individual regions. Transparent Generalized Additive Model Tree (TGAMT) \cite{fahner2018developing} was proposed as an explainable and transparent method that uses a CART-like greedy recursive search to grow the tree.
Multi-Run Subtree Encapsulation, which comes from the genetic programming (GP) realm, was proposed in \cite{howard2018explainable} as a way to generate simpler tree-based GP programs. If the tree contains sub-trees of different makeup but evaluating the same vector of results, they are to be considered as the same sub-tree. This reduces, according to the authors, the complexity of the entire tree structure and the resulting expressions, in favour of explainability.\\

Probabilistic Sentential Decision Diagrams (PSDD) \cite{liang2017towards} can be described as circuit representations where each parameter represents a conditional probability of deciding the input variables and each node is either a logical AND gate with two inputs or a logical OR gate with an arbitrary number of inputs. The PSDD structure can be visualized as an easily-interpretable binary tree.
Mind the Gap Model (MGM) \cite{kim2015mind} is a method for interpretable feature extraction and selection. The goal is to split the observation into clusters while returning the list of dimensions that are important for distinguishing them. The results are presented as a mix of numbers, which are the relevance values of each dimension, texts and graphs that represent the dimensions themselves. For example, in a classification problem of images representing the four seasons, MGM returns samples of images belonging to each class (spring, summer, autumn and winter) together with the list of their relevant features (like snow, sun and flowers) and the relevance values of each feature per target class (snow has a high relevance value for the class `winter').
Supersparse Linear Integer Model (SLIM) \cite{ustun2014supersparse} generates a scoring system from an input dataset by assigning a score to each variable that contributes to the prediction. These scores are multiplied by a set of coefficients inferred from the training dataset and then added, subtracted, and/or multiplied to make a prediction. Scores are generated by minimising the 0-1 loss to reach a high level of accuracy and to produce a classifier that is robust to outliers by applying a $\ell_0-penalty$ to encourage a high level of sparsity and a set of interpretability constraints which restricts coefficients to a user-defined set of meaningful and intuitive values.
Eventually, Unsupervised Interpretable Word Sense Disambiguation \cite{panchenko2017unsupervised} produces interpretable word sense disambiguation models that create clusters, or inventories, of words. For example, an inventory can be the collection of all the words related to `furniture' (such as table, chair and bed). The words are clustered according to their co-occurrence and relative position in a text, where close words are assumed to be highly correlated, and their syntactic dependency extracted from the Stanford Dependencies (representation of grammatical relations between words in a sentence.)
The resulting word groups can be interpreted at three levels: (1) word sense inventory where each sense of the word under analysis is displayed as a separate network-graph where the nodes are the semantically related words and the edges represent their semantic relationships. For example, the word `table' is connected to two networks corresponding to `furniture' and `data' senses; (2) sense feature representation characterized by a list of sparse features (which consists of words) ordered by relevance; (3) results of disambiguation in context by highlighting the most discriminative words that caused the prediction. Words like `cookie' and `website' indicate that `table' refers to a collection of data and not as a piece of furniture.\\

\section{Evaluation of methods for explainability}\label{xaievaluation}
\begin{figure}[!ht]
\centering
  \includegraphics[scale=0.42]{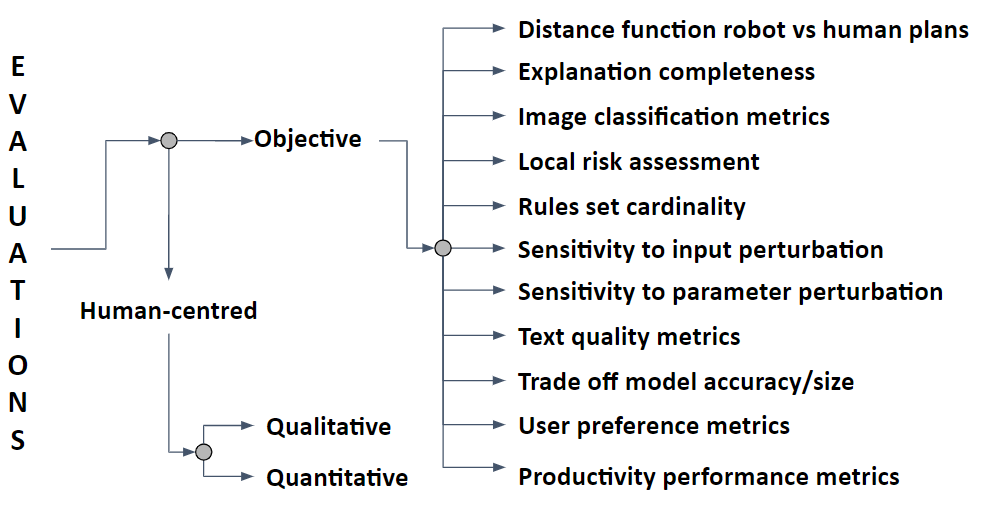}
  \newline
  \includegraphics[scale=0.52]{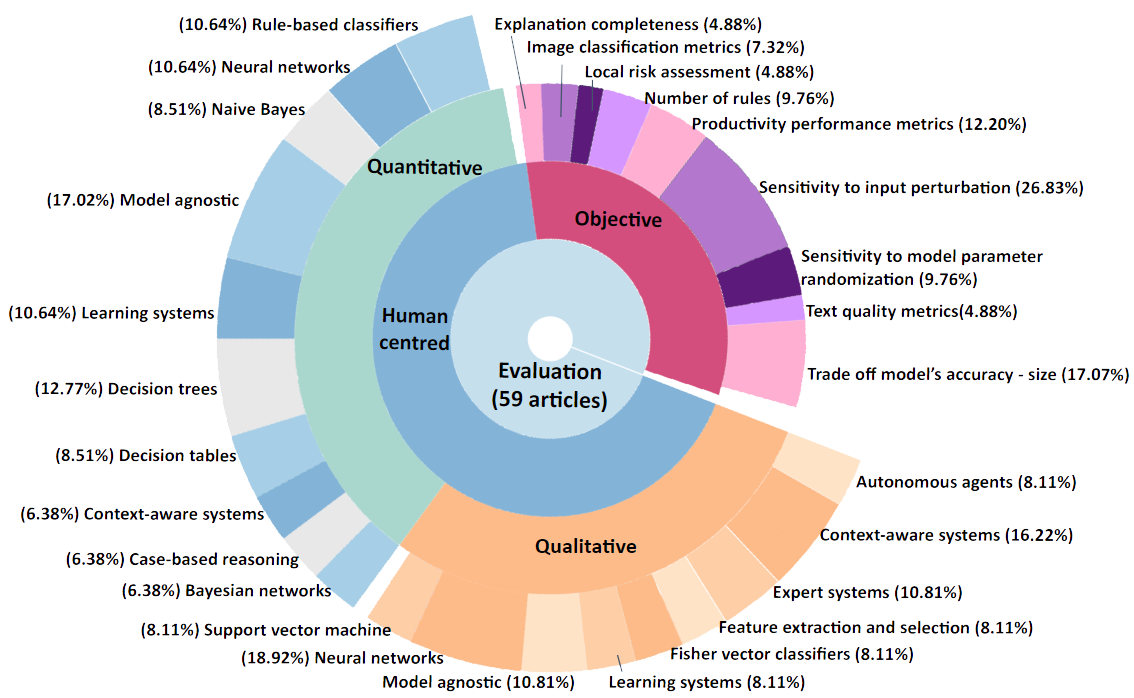}
  \caption{Classification of the approaches to evaluate methods for explainability (up) and distribution of the relative scientific studies across categories (down).}
  \label{fig:evaluation_tree}
\end{figure}

The proposal of many methods for explainability pushed authors to focus also on their evaluation. Different evaluation metrics were proposed and found in the literature as well as different types of evaluation were conducted, as summarised in figure \ref{fig:evaluation_tree}.
A thorough review of these studies led to the identification of two main ways to evaluate methods for explainability, as shown in figure \ref{fig:evaluation_tree}.

\begin{itemize}
\item {\verb|objective evaluations|} - it includes research studies that employed objective metrics and automated approaches to evaluate methods for explainability;
\item {\verb|human-centred evaluations|} - it contains those studies that evaluated methods for explainability with a human-in-the-loop approach by involving end-users and exploited their feedback or informed judgement.
\end{itemize}

The same dual categorisation system was suggested in \cite{bibal2016interpretability}, but they named the two classes \textit{heuristic-based} and \textit{user-based} metrics. The former includes quantitative measures which consist of mathematical entities such as, for example, the size of models \cite{dam2018explainable, freitas2014comprehensible, gacto2011interpretability, garcia2009study, otero2016improving}. 
This is a simple explainability indicator and it is based upon the assumption that the bigger the size of a model, the harder it becomes for the users to understand it. However, this assumption was proved false. One of the outcomes of the human-centred evaluation study conducted in \cite{freitas2014comprehensible} was that users found some larger models to be more comprehensible than some smaller models because larger models provided more classification-relevant information and users are unlikely to accept weaker, simpler models when the underlying modelled concept is believed to be complex.
An alternative categorisation was presented in \cite{preece2018asking} with three classes: \textit{application-grounded}, \textit{functionally-grounded} and \textit{human-grounded} evaluation metrics where functionally-grounded and human-grounded metrics respectively corresponds to the heuristic-based and user-based metrics proposed in \cite{bibal2016interpretability}. Application-grounded metrics assess the quality of machine-produced explanations of data-driven models by comparing the increase in productivity of a few users of these models when following these explanations instead of those produced by human engineers, as done in \cite{dam2018explainable, preece2018asking}. Because they involve humans, they are merged into the human-grounded metrics.

\subsection{Objective evaluations}\label{formal_evaluation}
Scholars proposed several metrics to evaluate, formally and objectively, the methods for explainability, listed in table \ref{tab:criteria}. In the scientific literature, there is consensus that simpler learning techniques, such as linear regression and DTs, can lead to more transparent inferences than more complex techniques, such as neural network, as they are intrinsically self-interpretable \cite{dam2018explainable}.
However, these simpler techniques usually do not lead to the construction of models with the same level of accuracy than those induced by more complex learning techniques. The interpretability of these models depends on many factors such as the learning algorithm, the learning architecture and its configuration (hyper-parameters).\\

A few sale performance indicators were utilized as a formal metric to assess the increase in productivity of the sales department when two methods of explainability, Explain and Ime \cite{robnik2008explaining,robnik2018explanation}, were applied to a complex real-world business problem of business-to-business sales forecasting \cite{bohanec2017decision, bohanec2017explaining}. The system was tested for a long period in a real-world company and the sale performance indicators were monitored. The indicators showed that the forecasts based on the Explain and Ime explanations outperformed initial sales forecasts, which supported the hypothesis that data-driven explanations better facilitate unbiased decision-making than the mental models of sale experts based upon their previous experience.
Two quantitative evaluation metrics to assess the interpretability of methods generating visual explanations of a neural network trained to classify images were presented in \cite{zhang2018visual}. 
The first metric, \textit{filter interpretability}, considers six types of semantics for CNN filters that must be annotated by humans on testing images at the pixel level: objects, parts, scenes, textures, materials, and colours. The metric measures the intersection areas between these annotations and the distribution of the activation values of a filter over a heat-map. If there are overlapping areas, it can be said that the filter represents these semantic concepts. The second metric, \textit{location instability}, checks if a CNN locates the relevant parts of the same object, shown in different images, at an almost constant relative distance, as the distances between the parts of an object must be almost invariant.
\cite{barratt2017interpnet} proposed to use three automated quantitative metrics, designed to assess the quality of text documents, to evaluate textual explanations automatically generated by methods for explainability: \textit{BiLingual Evaluation Understudy (BLEU)} that assesses the similarity of sentences based on the average percentage of n-gram matches, \textit{Automatic NT Translation Metric (METEOR)} that evaluates semantically the similarity between words of sentences by using pre-trained word embeddings and \textit{Consensus-based Image Description Evaluation (CIDEr)} that compares sentences generated by neural networks to reference explanations written by humans by counting `Term Frequency–Inverse Document Frequency' weighted n-grams.
A general evaluation metric for post-hoc methods for explainability was presented in \cite{laugel2019dangers} based on the risk of generating unjustified `counterfactual examples' which are instances that do not depend on previous knowledge but are artefacts of the classifier. This might happen when a model must predict an area not covered by the training set. The algorithm that generates these examples applies the minimal perturbation that changes the predicted classes of observation in such a way that it is still connected to the input data and avoid the construction of examples representing situations that are neither feasible nor logical. The explanations of the predictions made by an underlying model for these observations would not make sense and would not help the understanding of the model's logic. \\

Only a few scholars carried out formal comparisons, based on heuristic-based metrics, between different methods for explainability to evaluate their strengths and weaknesses. The methodologies utilized for the comparisons are (see also table \ref{tab:comparison-methods}):
\begin{itemize}
\item {\verb|sensitivity to input perturbation|} - some features of the input dataset are removed, masked or altered and the explanations generated by a method for explainability from the model trained on both the original and modified inputs are compared; 
\item {\verb|sensitivity to model parameter randomization|} - the outputs a method for explainability generated from a trained model and another model of the same architecture with some or all parameters replaced by random values are compared;
\item {\verb|explanation completeness|} - these approaches check which method generates explanations that describe the inferential process of the underlying model to the highest extent. This consists of capturing the highest number of features of the input that affect the decision process of the model.
\end{itemize}

The vast majority of these scientific articles compared methods for explainability designed to generate visual explanations of the logic followed by neural networks for the classification of either images or texts. All these methods produce maps, like heat-maps or feature maps, and the comparison is carried out by measuring the differences in these maps generated before and after the input or the model's parameters were perturbed. The complete list of the methods that were compared, along with the type of input data that were analysed in these comparisons, is shown in table \ref{tab:visual-comparison}.
\cite{adebayo2018local, adebayo2018sanity} compared the saliency maps generated by various methods for visual explainability to either a randomly initialised untrained network or from a copy of the dataset in which the labels were randomly permutated.
The degree of correlation between the saliency maps was measured by calculating  Spearman Rank Correlation coefficients.
Similarly, \cite{ancona2018towards} proposed to vary input images by occluding with zero-valued pixels their portions sharing the same relevance level, according to the saliency maps generated by four gradient-based attribution methods. The sensitivity of the four methods to this input perturbation was assessed with a formal metric, \textit{Sensitivity-n}, which quantifies the variation in the output caused when features sharing the same relevance level are removed. The results of this analysis showed that Occlusion Sensitivity (see Section \ref{newmethods}) is the method that identifies the few most important features, in respect with the other methods, because it suffers the faster variations in the output when the most relevant pixels are removed.
Layer-wise Relevance Propagation (LRP) and Sensitivity Analysis were tested in \cite{arras2016explaining, arras2017relevant, samek2017evaluating, samek2017explainable} by removing important words from input documents in text classification problems \cite{arras2016explaining, arras2017relevant} or replacing the most relevant pixels (in case of images) by randomly sampling new pixel values from a uniform distribution \cite{samek2017evaluating, samek2017explainable}. The metric used in this study assesses the differences in the model's classification accuracy between the original and the perturbed input points when fed into the model. Both studies showed that LRP qualitatively and quantitatively provides a better explanation of what made a DNN reach a specific classification prediction. \cite{binder2016analyzing} used the same evaluation approach of \cite{samek2017evaluating, samek2017explainable} to compare LRP with Occlusion Sensitivity and Sensitivity Analysis. LRP and Occlusion Sensitivity performed better than Sensitivity Analysis, seconding the findings of \cite{samek2017evaluating, samek2017explainable}.
Input perturbation was also used in \cite{alvarez2018robustness, ghorbani2017interpretation, kindermans2017reliability} to test the robustness of several methods that generate visual explanations of the inferential process of DNNs applied to image classification problems. 
Robustness concerns variations of an explanation provided by a method  with respect to changes in the input leading to that eplanation. Intuitively, if the input being explained is modified slightly-subtly enough  not to change the prediction of the model then the explanation must not change much either \cite{alvarez2018robustness}. On the one hand, \cite{alvarez2018robustness} applied a Gaussian noise to input images and measured the relative changes in the output with respect to these perturbations with the Local Lipschitz Continuity metric. On the other hand, \cite{ghorbani2017interpretation, kindermans2017reliability} added/subtracted a constant shift to the input images. Then, \cite{ghorbani2017interpretation} used two metrics to assess the similarity between the heat-maps generated from the original and the perturbed images: Spearman Rank Correlation coefficients and Top-$\kappa$ intersection which measure the size of the intersection of the $\kappa$ most important features before and after perturbation. \cite{kindermans2017reliability} instead measure the differences in the model's predictions by checking whether the methods for explainability under analysis satisfy the requirement of `input invariance' (see Section \ref{xaiattributes}). These studies show that all tested methods (table \ref{tab:visual-comparison}) are vulnerable even to small perturbations that do not affect the predictions of an underlying model but they  significantly change the heat and saliency maps produced by the explainers. These do not satisfy the `input invariance' requirement \cite{kindermans2017reliability}, that means they do not reflect the sensitivity of a model with respect to input perturbations.
Lastly, \cite{gevrey2003review} compared the completeness of the explanations generated by seven methods for explainability that interpret the logic of  DNNs by calculating the partial derivatives of the output according to the inputs, their perturbation and analysing the network's weights.

\subsection{Human-centred evaluation}\label{humansurvey}
Explanations are effective when they help end-users to build a complete and correct mental representation of the inferential process of a given model. Many scientific articles focused on testing the degree of explainability of one or multiple methods, with a human-in-the-loop approach. These experiments involved human participants of two kinds. On the one hand, people randomly selected from the lay public and without any prior technical/domain knowledge who were asked to interact with one or more explanatory tools and give feedback by filling questionnaires. On the other hand, domain experts who were asked to give informed opinions on the explanations produced by these methods and verify the consistency of the explanations with the domain knowledge.
Examined scientific articles can be clustered into two categories, depending on the nature of questions administered to people (see table \ref{tab:human-evaluation}).
\textit{Qualitative studies} are based upon open-ended questions aimed at achieving deeper insights whereas \textit{quantitative studies} make use of close-ended questions that can be easily analyzed statistically.
The first methods for explainability that were tested by human users are those generating textual explanations for Expert Systems (ES) in the 90's \cite{suermondt1993evaluation, ye1995impact}. These researches aimed at collecting pieces of evidence on whether and how explanations could enhance the user's confidence in the decisions proposed by an ES. Scholars carried out several human-centred evaluations over the years to assess the effects of textual explanations on end-users to other types of systems employing ML models, such as learning systems \cite{aleven2002effective}.
The impact of explanations on the reliability, trust and reliance of end-users on automated systems was explored in \cite{dzindolet2003role}. The participants were presented with photos of the Fort Sill terrain where the presence of a camouflaged soldier was indicated by an automated decision system. Initially, they considered the inference produced by the system to be trustworthy and reliable but, after observing the system making errors, they distrusted even reliable aids, unless an explanation was provided regarding why the system failed. In conclusion, explanations of these errors increased the trust of the participants in the automated system who were asked to estimate their perceived reliability on a 9-point Likert-format scale, ranging from `not very well' to `very well'.
The influence of explanation factors over the capacity of end-users to understand and develop a mental representation of the internal reference process of a model was investigated in \cite{harbers2010guidelines, harbers2010design, poursabzi2018manipulating, tullio2007works}. The explanations produced by the analysed methods for explainability consisted of graphical representation of the most relevant features. \cite{harbers2010guidelines, harbers2010design} showed that users of simulation-based training systems with virtual players prefer short explanations to long ones where length is defined by the number of elements in an explanation. An element can be a goal or an event of the training program. This was tested by showing to the participants four explanation alternatives of different length (they contained either one or two elements), in the form of DTs, for each action of the virtual players and asked to indicate which alternative they considered the most useful for increasing end-user understanding. The influence factors analysed in \cite{poursabzi2018manipulating} were the number of independent variables of a trained linear regression model and the values of these variables for each instance. Some participants were randomly assigned to check either a model that uses only two features or a model that uses eight features. The coefficients of the linear regression model were also presented only to half of the participants. Then, the participants were asked to estimate what would be the output of the model and to correct it in case it was not accurate. As expected, users can easily simulate models with a small number of features whereas, surprisingly, displaying model internal parameters can hamper their ability to notice unusual inputs and correct inaccurate predictions. The method for explainability tested in \cite{tullio2007works} listed the two most relevant features and the prediction of the model. The prediction was coded with a solid colour taken from a scale running from red, representing the most negative possible outcome, to green, the most positive one. Participants were asked to interact with the system for two weeks at the end of which they were interviewed to collect their feedback and to check whether they gained some insight on the logic of the model to be explained by the explanatory system under analysis.
\cite{krause2016interacting} studied the explainability of an interactive interface, called \textit{Prospector}, containing a set of diagnostic tools that allows end-user, via visual and numerical representations, to understand how the features of a dataset affect the prediction of a model overall. Users can also inspect a specific instance to check its prediction and can weak feature values to see how the model responds. A team of data-scientists was asked to interact with this tool to debug a set of models designed to predict if a patient is at risk of developing diabetes by using a database of electronic medical records. 
The human experiment consists of interviewing, at the end of the experiment, the data scientists on whether they feel that it was beneficial for their work.
Methods for evaluating the interpretability of data-driven models with a human-in-the-loop approach were proposed in \cite{lage2018human, allahyari2011user, huysmans2011empirical, luvstrek2014comprehensibility}. The approach proposed in \cite{lage2018human} identifies some \textit{proxies} which consist of other, simpler models inherently more explainable. For example, a DT is inherently more interpretable than DNNs and the former methods can be used to explain the latter. The authors presented to participants a list of the coefficients of the features used by each proxy and a graphical depiction of its structure (in the form of a DT) and they asked them to identify the correct prediction. 
\cite{luvstrek2014comprehensibility} proposed instead to assess the comprehensibility of DTs by asking the participants to perform the following tasks: (I) classify a data-point according to the classification tree, (II) explain the classification of a data-point by pointing out which attributes’ values must be changed or retained to modify the instance's class, (III) validate a part of the classification tree by asking the participant to check whether a statement about the domain is confirmed/rejected by the tree, (IV) discover new knowledge by finding a property (attribute-value pair) that is unusual for instances from one class, (V) rate the classification tree by giving a subjective opinion on the comprehensibility of the tree and, lastly, (VI) compare two classification tree by saying which one is more comprehensible. 
The two studies presented in \cite{allahyari2011user, huysmans2011empirical} analysed the interpretability of predictive models by asking participants to interact with them and fill self-reporting questionnaires. The surveys carried out in \cite{allahyari2011user} aimed at comparing six supervised learning algorithms. These were ranked in order of preference based on the subjective quantification of understandability obtained from the self-reporting questionnaires filled by participants. Pairs of models trained on the same dataset were generated and participants were asked to rate them on a scale where one extreme is `the first model is the most understandable' to the other extreme `the second model is the most understandable' via increasingly positive grades. In the study carried out in \cite{huysmans2011empirical}, participants were required to evaluate the explanations of a credit model, trained to accept or reject loan applications, consisting of IF-THEN rules and displayed as a decision tree. They were asked to predict the model's outcome on a new loan application, answer a few yes/no questions such as ``Does the model accept all the people with an age above 60?'' and rate, for each question, the degree of confidence in the answer on a scale from 1 (Totally not confident) to 5 (Very confident). The authors measured, besides the answer confidence, other two variables about task performance: accuracy, quantified as the percentage of correct answers, and the time in seconds spent to answer the questions.
The effectiveness of why-oriented explanation systems in debugging a na\"{i}ve Bayes learning model for text classification and in context-aware applications were respectively tested in \cite{kulesza2011oriented, kulesza2015principles} and \cite{lim2009assessing,lim2009and}. \cite{kulesza2011oriented, kulesza2015principles} asked participants to train a prototype system, based on a Multinomial na\"{i}ve Bayes classifier, that can learn from users how to automatically classify emails by manually moving a few of them from the inbox into an appropriate folder. The system was subsequently run over a new set of messages, some of which were wrongly classified. The participants had to debug the system by asking `why' questions via an interactive explanation tool producing textual answers and by giving two types of feedback: some participants could label the most relevant feature (words) whereas the others could only provide more labelled instances (moving more messages to the appropriate folders). At the end of the session, the participants filled a questionnaire to express their opinions on the prototype. 
In the experiment run in \cite{lim2009assessing,lim2009and}, participants were invited to interact with a model that predicts whether a person is doing physical exercise or not, based on the body temperature and the pace. They were shown with some examples of inputs and outputs accompanied by graphical (in the form of decision trees) and textual explanations on the logic followed by the model. Half of the participants were presented with \underline{why} explanations, such as \textit{``Output classified as Not Exercising, because Body Temperature $<$ 37 and Pace $<$ 3''} whereas the other half were presented with \underline{why not} explanations, such as \textit{``Output \underline{not} classified as Exercising, because Pace $<$ 3, but \underline{not} Body Temperature $>$ 37''}. Subsequently, the participants had to fill two questionnaires to check their understanding by asking questions how the system works and to give feedback on the explanations in terms of understandability, trust and usefulness. Both questionnaires contained a mix of open and close questions, where the close ones consisted of a 7-point Likert scale.
A qualitative evaluation of the interpretability of the Mind the Gap Model (MGM) method for explainability was gathered in \cite{kim2015mind}. MGM clustered the data of an input dataset according to the most relevant features. The participants were presented with the raw data and the data clustered with MGM and k-means and were asked to write a 2-3 sentence executive summary of each data representation within 5 minutes. They all found impossible to summarise the raw data, not being able to complete the task in the given amount of time, but they managed to do so on the data with clustered MGM and k-means. 
To test Bayesian Case Model (BCM), \cite{kim2014bayesian} asked the participants to assign sixteen recipes, described only by a set of ingredients, to the right category (so a recipe of cookies had to be classified under `cookie') and then they counted how many of them were correctly classified. BCM was compared with Latent Dirichlet Allocation (LDA), a clustering approach based on extracting similar characteristics in the data.
The need for XAI in Intelligent Tutoring Systems (ITS) was explored in \cite{putnam2019exploring}. The participants in the study were asked to use an ITS that provided tools to explore and explain an algorithm solving constraint satisfaction problems in an interactive simulation. The participants were instructed by the exploration tool with a textual hint message. They could select to have the hint explained by the explanatory tool which was also designed to solicit their suggestions on the explanations they would like to see for each hint by presenting them a checkbox list with the following options: `why the system gave this hint', `how the system chose this hint', `some other explanation about this hint' (followed by an open-text field) and `I do not want an explanation for this hint.'\\

Many scholars proposed human-centred evaluation approaches for methods for explainability generating visual explanations. The participants selected in the study in \cite{stock2018convnets} were presented with whole images misclassified by a DNN and the visual explanations generated by LIME and MMD-critic. For example, a photo of a Jeep with a red-cross was wrongly classified as an ambulance and the visual explanations show the red-cross with the rest of the image greyed out. Participants were asked to say whether the class predicted by the model was nonetheless relevant where the possible answers to questions like `Is the label Red-Cross relevant?' were `yes' and `no'.
A similar experiment was carried out in \cite{bau2017network} to test GAN Dissection. Participants were presented with images reporting highlighted patches showing the most highly-activating regions for each unit at each intermediate convolutional layer of a DNN. Each layer was aligned with a semantic and were given labels across a range of objects, parts, scenes, textures, materials, and colours. For example, if a DNN was trained to recognize a list of object in input images, like flowers and cars, the semantic consists of this list and images showing flowers were labelled `flower`. The participants were asked to say if the highlighted patches were pertinent to the label by answering yes/no questions.
The capacity of Anchors and LIME in helping end-users forecasting the predictions of an image classifier was tested in \cite{ribeiro2018anchors}. Participants were asked to predict the output class assigned by the classifier to ten random test instances before and ten instances after seeing two rounds of explanations generated by either Anchors or LIME.
A few scholars conducted human-centred studies to test the interpretability of the heat-maps generated with LRP. \cite{lapuschkin2016analyzing,srinivasan2017interpretable} applied it respectively to test models built with Fisher vector classifiers for object recognition in images and to SVMs, trained on videos, to understand and interpret action recognition and to check whether LRP allows identifying in which point of the video the important action happens. 
By visually inspecting the heat-maps, the authors of the two studies could show a possible weakness of the underlying classifiers by looking at the regions highlighted in the heat-maps and examining whether they were relevant for the recognition of an object (or at least part of it) in images and of the areas of video frames showing the actions performed in the video. 
Similarly, \cite{sturm2016interpretable} employed LRP with DNNs for electroencephalography (EEG) data analysis. The predictions of the trained DNNs are transformed with LRP, for every trial, into heat-maps indicating the relevance of each data point. 
The relevance information can be plotted as a scalp topography that can be visually inspected by experts to check if there are neurophysiologically plausible patterns in the EEG data. 
\cite{ding2017visualizing} used LRP for computing the contribution of contextual words to arbitrary hidden states in the attention-based encoder-decoder framework of neural machine translation (NMT).
As per the previous studies, the authors checked if the translation made by the NMT (Japanese-English) were right or wrong and what types of errors were made more frequently.
The participants in \cite{spinner2019explainer} were asked to interact with explAIner which showed them explanations generated by LRP and Saliency Analysis of both a simple and a complex network trained on the MNIST dataset and, in the meantime, to communicate their thoughts and actions by `thinking aloud'. The sessions were audio-recorded and screen-captured. At the end of the sessions, participants were also interviewed to provide qualitative feedback on the overall experience.
Another method for explainability producing visual explanations as maps, GradCam, was applied to multivariate time series from photovoltaic power plants that were fed into a neural network to forecast the energy production of these plants in different weather conditions \cite{assaf2019explainable}. GradCam was used to explain which features, such as environmental temperature, wind bearing or humidity, or any combinations of these features were responsible, at different time intervals, for a given prediction. The results showed that GradCam was able to visualise the network’s attention over the time dimension and the features of multivariate time series data.
\cite{erhan2009visualizing} compared other three methods, namely Activation Maximization, Sampling Unit and Linear Combination, designed to produce explanations as heat-maps of the most relevant features of the input images. Activation Maximization consists of selecting the input features that maximise the activation of a single hidden neuron. Sampling Unit consists of setting the value of a neuron to one and calculating the probability with which each sample is assigned to a class. Lastly, Linear Combination consists of choosing the largest weights of the connections between neurons of two adjacent layers. The authors did not use any objective measure to compare these methods but  a qualitative visual inspection of the heat-maps and  the connections between  them.

\section{Final remarks and recommendations}\label{conclusions}
Research on methods to explain the inner logic either of a learning algorithm, a model induced from it, or a knowledge-based approach for inference is now generally recognized as a core area of AI and is referred to as eXplainable Artificial Intelligence (XAI). 
Note that other common terms exist, such as `Interpretable Machine Learning', but with XAI we would like to emphasise the wider applicability of this emerging and fascinating field of research.
Several scientific studies are published every year, with many workshops and conferences organised around the world. to propose novel methods and disseminate findings. Although this has led to the production of an abundance of knowledge, unfortunately, this is very scattered. 
This systematic review attempted to fill this gap by organising this vast knowledge in a structured and hierarchical way. Some scholars already tried to classify scientific studies but, given a large amount of literature, they decided to focus only on a specific aspect of explainability. To the best of our knowledge, this is the first attempt to review a wider body of literature that has led to the definition of four clusters: (I) reviews focused on specific aspects of XAI, (II) the theories and notions related to the concept of explainability, (III) the methods aimed at explaining the inferential process of data-driven and knowledge-based modelling approaches, and (IV) the ways to evaluate the methods for explainability.
Many studies within XAI have focused on improving the quality and widening the variety of explanations for several types of learning approaches with data. 
Since the early '80s and '90s, with research concerned only with textual explanations, to nowadays, scholars have been targeting new explanation formats such as visual aids, rules, numbers and different combinations of these.
For each of these formats, scholars designed, deployed and tested several solutions, such as saliency masks, attention maps, heat-maps, feature maps, as well as graphs, rules sets, trees and dialogues. These advances were aimed at meeting the needs of different types of end-users operating in various fields, such as laypeople, doctors and lawyers, and adapting explanations to their domains of application.
Additionally, scholars widened their research horizons by incorporating the knowledge developed in other fields, like Psychology, Philosophy and Social Sciences, into the design of the novel methods for explainability. The goal was to improve the structure, efficiency and efficacy on people's understanding of automatically generated explanations. All this research has produced many definitions of explainability and identified several notions related to it, such as interpretability, understandability, comprehensibility and justifiability, just to mention a few. Coupled to these notions, different objective metrics have also been produced for their measurement.
Despite the large number and variety of methods and metrics for explainability proposed so far, there are still important scientific issues that must be tackled. 
Firstly, there is no agreement among scholars on what an explanation exactly is and which are the salient properties that should be considered to make it effective and understandable for end-users, in particular non-experts. 
Secondly, the construct of explainability is a concept borrowed from Psychology, since it is strictly connected to humans, and it is also linked to other constructs such as trust, transparency and privacy.
Thirdly, this concept has been invoked in several fields, such as Physics, Mathematics, Social Sciences and Medicine.
All this make its formalisation and operationalisation a non-trivial research task.
This holds true for every explanation format, either textual, visual,  numerical. The same can be said for rule-based explanations, in particular for those that are generated after a model has been induced by employing deep-learning neural networks.  
In accordance with \cite{adadi2018peeking}, we believe that scholars have produced enough material that can be used to construct a generally applicable framework for XAI. This would guide the advancement of novel, end-to-end methods for explainability, rather than keep creating isolated methods that remain only fragments of a broad solution which should also be flexible enough to adapt to various contexts, fields of application and type of end-users. 
Additionally, the ultimate scope of an explanation is to help end-users build a complete and correct mental model of the inferential process of either a learning algorithm or a knowledge-based system and to promote trust for its outputs.
An area for future research is the involvement of humans, as final users of artificial explanations, since their role has not been sufficiently studied in the creation and exploitation of existing explainability methods \cite{adadi2018peeking}. 
To support this research direction, we recommend exploiting knowledge and experiences belonging to the field of Human-Computer Interaction and its advances for the development of interactive explanatory interfaces \cite{lawless2019artificial}.
This should always take into consideration the existing trade-off between the dimensions of models accuracy and their interpretability/explainability which are currently inversely correlated.
One possible suggestion is the use of methods that take advantage of modern learning techniques, to maximise the former dimension, and reasoning approaches to optimise the latter dimension.
The assumption is that integrating connectionist and symbolic paradigms is the most efficient way to produce meaningful and precise explanations. 
Advances on these two paradigms are immense, however, their intersection is under exploration. For example, on one hand, a school of thought suggests to firstly train accurate models from data and then wrap them with a reasoning layer \cite{bride2018towards}.
This layer, for instance, can be produced by exploiting advances in defeasible reasoning and argumentation \cite{rizzo2018qualitative,rizzo2019inferential,zeng2018building} making use of knowledge-bases constructed with a human-in-the-loop approach.
On the other hand, another direction is to promote the use of neuro-symbolic learning and reasoning in parallel, each one informing the other at all stages of model construction and evaluation \cite{garcez2015neural}. 
Eventually, another interesting, novel and under-explored direction for future scholars concerns the development of structured formats of explanations. These formats should consider all the elements and notions related to explainability, that can be trained with connectionist paradigms from data. \\

\begin{figure}[!ht]
\begin{minipage}{\textwidth}
\centering
  \subcaptionbox{}
    {\includegraphics[scale=0.54, align=c]{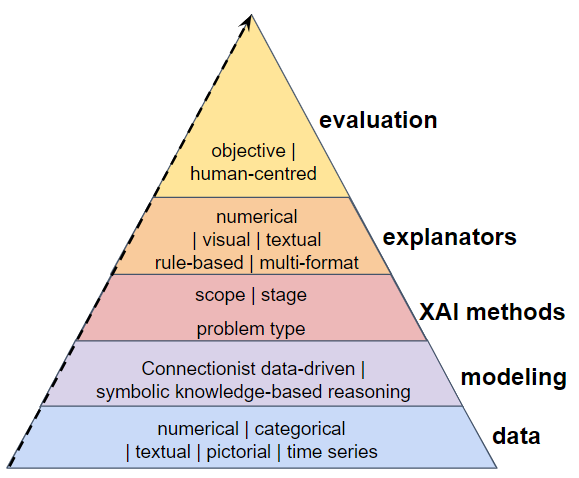}}
  \subcaptionbox{}
    {\includegraphics[scale=0.54, align=c]{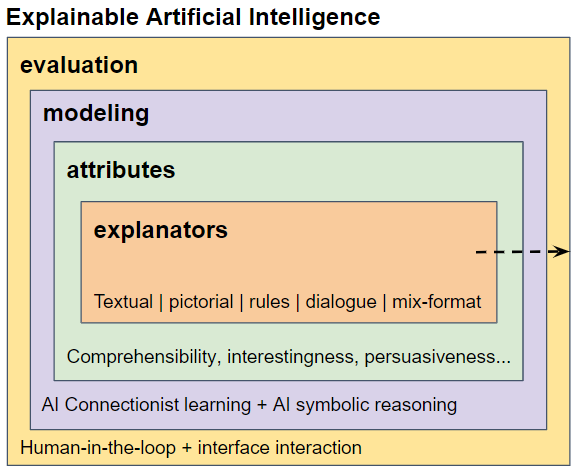}}
  \caption{State of the art (a) and envisioned (b) frameworks for eXplainable Artificial Intelligence.}
  \label{fig:xai_frameworks}
\end{minipage}
\end{figure}

\newpage
In summary, an high-level structure of a the current state-of-the-are in XAI is depicted in figure \ref{fig:xai_frameworks} (part a). On one hand, here, emphasis has been placed on the sequence of research activities currently and often performed by several scholars, their dependencies and order. This sequence usually starts from input data that is then used for modeling purposes, employing connectionist data-driven learning or symbolic reasoning knowledge-based paradigms. 
After a model has been formed, then an XAI methods is applied for its analysis, knowledge discovery, supporting its interpretability. This phase provide the end-users of these models with one or more explanators for the purpose of its explainability. 
Eventually, very few scholars have proposed approaches for evaluating such layer of explainability, either proposing formal, objective metrics or involving human-centred evaluation with model designers and end-users.
On the other end, what we believe is an ideal framework for XAI is depicted  in figure \ref{fig:xai_frameworks} (part b). 
Here, the main focus should be on the explanators, which is what end-users will ultimately interact with. The development of explanators should be designed by taking into account the multiple attributes that are linked to the psychological construct of explainability. Subsequently, scholars can focus on the modeling phase, preferrably using both connectionist and symbolic paradigms from Artificial Intelligence. This will allow to develop models that are both robust in terms of accuracy but also intrinsically interpretable during all the stages of development.
Eventually, the last phase should focus on the evaluation of explainability of such models with a human-in-the-loop approach, involving both designers and end-users, and the development of interactive interface for supporting model interpretability and inference explainability.\\





\newpage
\bibliography{reference}

\newpage
\appendix


\section{Appendix}
\begin{table}[h!]
\footnotesize
  \caption{Classification of the systematic and literature review articles on explainable artificial intelligence and machine learning interpretability}
  \label{tab:surveys}
  \begin{tabular}{m{3.8cm} m{4.2cm} m{4.3cm}}
    \hline
    Category & Subcategory & Reference\\
    \hline
    Application fields & Finance, military and transportation & \cite{adadi2018peeking}\\
    Application fields & Law & \cite{adadi2018peeking, rudin2014algorithms,rudin2019stop}\\
    Application fields & Healthcare & \cite{fellous2019explainable,freitas2010importance,gordon2019explainable,holzinger2016interactive,ram2019importance,rudin2014algorithms,rudin2019stop}\\
    Application fields & Human-computer interaction & \cite{lawless2019artificial}\\
    Approaches - data-driven & Bayesian networks & \cite{lacave2002review}\\
    Approaches - data-driven & Neural networks & \cite{andrews1995survey,arras2019evaluating,bengio2013representation,choo2018visual,craven1992visualizing,dam2018explainable,hailesilassie2016rule,hoffman2018explaining,jacobsson2005rule,mathews2019explainable,montavon2017methods,samek2019towards,zhang2018visual}\\
    Approaches - data-driven & Support vector machines & \cite{backhaus2014classification,martens2007comprehensible}\\
    Approaches - knowledge-driven & Expert systems & \cite{swartout1993explanation}\\
    Approaches - knowledge-driven & Intelligent systems & \cite{gregor1999explanations}\\
    Approaches - knowledge-driven & Recommender systems & \cite{papadimitriou2012generalized}\\
    Theories and concepts & \emph{See expanded subcategories in Sec. \ref{notions}} & \cite{abdul2018trends,bibal2016interpretability,byrne2019counterfactuals,doran2017does,dovsilovic2018explainable,lipton2018mythos, miller2017explainable, paez2019pragmatic, sormo2005explanation}\\
    Output formats & Rule & \cite{andrews1995survey, besold2015towards, bonacina2017automated, fernandez2019evolutionary, garcez2015neural, guillaume2001designing, hailesilassie2016rule, jacobsson2005rule, lisboa2013interpretability, martens2007comprehensible}\\
    Output formats & Textual & \cite{belle2017logic, mencar2018paving, swartout1993explanation}\\
    Output formats & Visual & \cite{choo2018visual, craven1992visualizing,  lisboa2013interpretability, zhang2018visual}\\
    Problem Types & Classification & \cite{backhaus2014classification,choo2018visual, freitas2010importance,guillaume2001designing, hailesilassie2016rule,martens2011performance, otte2013safe,zhang2018visual}\\
    Problem Types & Regression & \cite{otte2013safe}\\
    Generic reviews &  -  & \cite{adadi2018peeking, biran2017explanation,cui2019integrative,gade2019explainable,guidotti2018survey,xu2019explainable}\\
  \hline
\end{tabular}
\end{table}
\vspace{-2mm}

\begin{table}[h!]
\footnotesize
    \caption{Model agnostic methods for explainability generating numerical explanations, classified according to the output format, stage (AH: Ante-hoc, PH: Post-hoc), type of problems (C: Classification, R: Regression), scope (G: Global, L: Local) and input data (NC: Numerical / Categorical, P: Pictorials, T: Textual, TS: Time Series) of the  model.}
    \label{tab:model-agnostic-numerical}
    \begin{tabular}{m{4.2cm} m{2.5cm} m{0.5cm} m{0.5cm} m{0.5cm} m{0.6cm} m{0.8cm} m{0.6cm}}
    \hline
    Method for explainability & Authors & Ref & Year & Stage & Scope & Problem & Input\\
    \hline
    Distill-and-Compare &  \citeauthor{tan2018distill} & \cite{tan2018distill} & \citeyear{tan2018distill} & PH & G & C / R & NC\\
    Explain and Ime & \citeauthor{robnik2008explaining,robnik2018explanation} & \cite{robnik2008explaining,robnik2018explanation} & \citeyear{robnik2008explaining,robnik2018explanation} & PH & L & C & NC\\
    Feature contribution &  \citeauthor{Kononenko10anefficient,kononenko2013explanation,vstrumbelj2009explaining} &  \cite{Kononenko10anefficient,kononenko2013explanation,vstrumbelj2009explaining} &  \citeyear{Kononenko10anefficient,kononenko2013explanation,vstrumbelj2009explaining} & PH & L & C / R & NC\\
    Feature contribution & \citeauthor{vstrumbelj2008towards,vstrumbelj2010explanation} & \cite{vstrumbelj2008towards,vstrumbelj2010explanation} & \citeyear{vstrumbelj2008towards,vstrumbelj2010explanation} & PH & G & C / R & NC\\
    Feature Importance &  \citeauthor{henelius2014peek} &  \cite{henelius2014peek} & \citeyear{henelius2014peek}& PH & G & C & NC\\
    Feature perturbation &  \citeauthor{vstrumbelj2014explaining} & \cite{vstrumbelj2014explaining} &  \citeyear{vstrumbelj2014explaining} & PH & G & C / R & NC\\
    Global Sensitivity Analysis (GSA) &  \citeauthor{cortez2011opening,cortez2013using} &  \cite{cortez2011opening,cortez2013using} &  \citeyear{cortez2011opening,cortez2013using} & PH & G & C / R & NC\\
    Gradient Feature Auditing (GFA) &  \citeauthor{adler2018auditing} &  \cite{adler2018auditing} &  \citeyear{adler2018auditing} & PH & G & C / R & NC\\
    Influence functions & \citeauthor{koh2017understanding} & \cite{koh2017understanding} & \citeyear{koh2017understanding} & PH & G & C & P\\
    Monotone Influence Measures & \citeauthor{sliwinski2017characterization} & \cite{sliwinski2017characterization} &  \citeyear{sliwinski2017characterization} & PH & L & C & P\\
    Quantitative Input Influence (QII) functions &  \citeauthor{datta2016algorithmic} &  \cite{datta2016algorithmic} &  \citeyear{datta2016algorithmic} & PH & G & C & NC\\
    SHapley Additive exPlanations (SHAP) &  \citeauthor{lundberg2017unified} &  \cite{lundberg2017unified} &  \citeyear{lundberg2017unified} & PH & G & C & P\\
    \hline
\end{tabular}
\end{table}

\begin{table}[h!]
\centering
\footnotesize
    \caption{Model agnostic methods for explainability generating mixed explanations, classified according to the output format, stage (AH: Ante-hoc, PH: Post-hoc), type of problems (C: Classification, R: Regression), scope (G: Global, L: Local) and input data (NC: Numerical / Categorical, P: Pictorials, T: Textual, TS: Time Series) of the underlying model.}
    \label{tab:model-agnostic-mixed}
    \begin{tabular}{m{4.2cm} m{2.4cm} m{0.5cm} m{0.5cm} m{0.5cm} m{0.6cm} m{0.8cm} m{0.8cm}}
    \hline
    Method for explainability & Authors & Ref & Year & Stage & Scope & Problem & Input\\
    \hline
    Bayesian Teaching & \citeauthor{yang2017explainable} & \cite{yang2017explainable} & \citeyear{yang2017explainable} & PH & L & C / R & NC\\
    Evasion-Prone Samples Selection & \citeauthor{liu2018adversarial} & \cite{liu2018adversarial} & \citeyear{liu2018adversarial} & PH & G & C & T\\
    ExplAIner & \citeauthor{spinner2019explainer} & \cite{spinner2019explainer} & \citeyear{spinner2019explainer} & PH & G & C / R & P; NC; TS\\
    Functional ANOVA decomposition, Variable Interaction Network graph &  \citeauthor{hooker2004discovering} &  \cite{hooker2004discovering} &  \citeyear{hooker2004discovering} & PH & G & C / R & NC\\
    Justification Narratives & \citeauthor{biran2014justification} & \cite{biran2014justification} &  \citeyear{biran2014justification} & PH & G & C & NC\\
    Local Interpretable Model-Agnostic Explanations (LIME) &  \citeauthor{ribeiro2016model} &  \cite{ribeiro2016model,ribeiro2016should} & \citeyear{ribeiro2016model} & PH & L & C & P; T\\
    Maximum Mean Discrepancy (MMD)-critic & \citeauthor{kim2016examples} & \cite{kim2016examples} & \citeyear{kim2016examples} & PH & L & C & P\\
    Neighborhood-Based Explanations & \citeauthor{caruana1999case} & \cite{caruana1999case} & \citeyear{caruana1999case} & PH & L & C & NC\\
    Pertinent negatives & \citeauthor{dhurandhar2018explanations} & \cite{dhurandhar2018explanations} & \citeyear{dhurandhar2018explanations} & PH & L & C & P; NC\\
    Rivelo & \citeauthor{tamagnini2017interpreting} & \cite{tamagnini2017interpreting} & \citeyear{tamagnini2017interpreting} & PH & L & C & T\\
    Sequential Bayesian Quadrature & \citeauthor{khanna2019interpreting} & \cite{khanna2019interpreting} &  \citeyear{khanna2019interpreting} & PH & L & C & P; NC\\
    Set Cover Optimization (SCO) & \citeauthor{bien2011prototype} & \cite{bien2011prototype} & \citeyear{bien2011prototype} & PH & L & C & P; NC\\
    \hline
\end{tabular}
\end{table}

\begin{table}[h!]
\footnotesize
    \caption{Model agnostic methods for explainability generating rule-based explanations, classified according to the output format, stage (AH: Ante-hoc, PH: Post-hoc), type of problems (C: Classification, R: Regression), the scope (G: Global, L: Local) and input data (NC: Numerical/Categorical, P: Pictorials, T: Textual, TS: Time Series) of the  model.}
    \label{tab:model-agnostic-rule}
    \begin{tabular}{m{4.3cm} m{2.1cm} m{0.5cm} m{0.7cm} m{0.5cm} m{0.6cm} m{0.8cm} m{0.6cm}}
    \hline
    Method for explainability & Authors & Ref & Year & Stage & Scope & Problem & Input\\
    \hline
    Anchors & \citeauthor{ribeiro2018anchors} & \cite{ribeiro2018anchors} & \citeyear{ribeiro2018anchors} & PH & G / L & C & T\\
    Automated Reasoning & \citeauthor{bride2018towards} & \cite{bride2018towards} & \citeyear{bride2018towards} & PH & G & C & NC\\
    Genetic Rule EXtraction (G-REX) & \citeauthor{johansson2004accuracy} & \cite{johansson2004accuracy,johansson2004truth} & \citeyear{johansson2004accuracy} & PH & G & C / R & NC\\
    Model Extraction &  \citeauthor{bastani2017interpretability} &  \cite{bastani2017interpretability} &  \citeyear{bastani2017interpretability} & PH & G & C / R & NC\\
    Partition Aware Local Model (PALM) &  \citeauthor{krishnan2017palm} &  \cite{krishnan2017palm} &  \citeyear{krishnan2017palm} & PH & G & C / R & NC\\
    \hline
\end{tabular}
\end{table}

\begin{table}[h!]
\footnotesize
    \caption{Model agnostic methods for explainability generating visual explanations, classified according to output format, stage (AH: Ante-hoc, PH: Post-hoc), type of problems (C: Classification, R: Regression), the scope (G: Global, L: Local) and input data (NC: Numerical/Categorical, P: Pictorials, T: Textual, TS: Time Series) of the underlying model.}
    \label{tab:model-agnostic-visual}
    \begin{tabular}{m{4.5cm} m{2.2cm} m{0.5cm} m{0.5cm} m{0.5cm} m{0.5cm} m{0.7cm} m{0.7cm}}
    \hline
    Method for explainability & Authors & Ref & Year & Stage & Scope & Problem & Input\\
    \hline
    Class Signatures & \citeauthor{krause2016using} & \cite{krause2016using} & \citeyear{krause2016using} & PH & G & C / R & NC\\
    ExplainD &  \citeauthor{poulin2006visual} &  \cite{poulin2006visual} &  \citeyear{poulin2006visual} & PH & G & C & NC\\
    Explanation Graph based on perturbed input element order &  \citeauthor{alvarez2017causal} &  \cite{alvarez2017causal} &  \citeyear{alvarez2017causal} & PH & L & C & T\\
    Image Perturbation &  \citeauthor{fong2017interpretable} &  \cite{fong2017interpretable} &  \citeyear{fong2017interpretable} & PH & L & C & P\\
    Individual Conditional Expectation  & \citeauthor{goldstein2015peeking} & \cite{goldstein2015peeking} & \citeyear{goldstein2015peeking} & PH & G & C / R & NC\\
    iVisClassifier & \citeauthor{choo2010ivisclassifier} & \cite{choo2010ivisclassifier} & \citeyear{choo2010ivisclassifier} & PH & G & C & NC\\
    Layer-wise Relevance Propagation  &  \citeauthor{bach2015pixel} &  \cite{bach2015pixel} &  \citeyear{bach2015pixel} & PH & L & C & P\\
    Manifold &  \citeauthor{zhang2019manifold} & \cite{zhang2019manifold} & \citeyear{zhang2019manifold} & PH & G & C / R & NC\\
    MLCube Explorer &  \citeauthor{kahng2016visual} &  \cite{kahng2016visual} &  \citeyear{kahng2016visual} & PH & G & C & NC\\
    Partial Importance and Individual Conditional Importance  plots based on Shapley feature importance & \citeauthor{casalicchio2018visualizing} & \cite{casalicchio2018visualizing} & \citeyear{casalicchio2018visualizing} & PH & G & C / R & NC\\
    Restricted Support Region Set (RSRS) Detection &  \citeauthor{liu2012has} &  \cite{liu2012has} &  \citeyear{liu2012has} & PH & L & C & T\\
    Saliency Detection & \citeauthor{dabkowski2017real} & \cite{dabkowski2017real} & \citeyear{dabkowski2017real} & PH & L & C & P\\
    Sensitivity analysis & \citeauthor{baehrens2010explain} & \cite{baehrens2010explain} &  \citeyear{baehrens2010explain} & PH & L & C & P; NC\\
    Spectral Relevance Analysis (SpRAy) & \citeauthor{lapuschkin2019unmasking} & \cite{lapuschkin2019unmasking} & \citeyear{lapuschkin2019unmasking} & PH & G & C & P\\
    Worst-case perturbations & \citeauthor{goodfellow2015explaining} & \cite{goodfellow2015explaining} & \citeyear{goodfellow2015explaining} & PH & L & C & P\\
    \hline
\end{tabular}
\end{table}

\begin{table}[htbp]
\footnotesize
    \caption{Methods for explainability for neural networks generating visual explanations, classified according to the stage (AH: Ante-hoc, PH: Post-hoc), type of problems (C: Classification, R: Regression), scope (G: Global, L: Local) and input data (NC: Numerical/Categorical, P: Pictorials, T: Textual, TS: Time Series).}
    \label{tab:neural-networks-visual}
    \begin{tabular}{m{3.9cm} m{2.7cm} m{0.5cm} m{0.5cm} m{0.5cm} m{0.6cm} m{0.8cm} m{1.05cm}}
    \hline
    Method for explainability & Authors & Ref & Year & Stage & Scope & Problem & Input\\
    \hline
    Activation maps &  \citeauthor{hamidi2019interactive} &  \cite{hamidi2019interactive} &  \citeyear{hamidi2019interactive} & PH & L & C & P\\
    Activation Maximization &  \citeauthor{erhan2010understanding,nguyen2016multifaceted} &  \cite{erhan2010understanding,nguyen2016multifaceted,nguyen2016synthesizing} &  \citeyear{erhan2010understanding,nguyen2016multifaceted} & PH & L & C & P\\
    ActiVis &  \citeauthor{kahng2018cti} &  \cite{kahng2018cti} &  \citeyear{kahng2018cti} & PH & G & C / R & NC\\
    And-Or Graph (AOG) &  \citeauthor{zhang2017growing} &  \cite{zhang2017growing} &  \citeyear{zhang2017growing} & PH & G & C & P\\
    Cell Activation Values & \citeauthor{karpathy2015visualizing} & \cite{karpathy2015visualizing} &  \citeyear{karpathy2015visualizing} & PH & G / L & C & T\\
    CLass-Enhanced Attentive Response (CLEAR) & \citeauthor{kumar2017explaining} & \cite{kumar2017explaining} &  \citeyear{kumar2017explaining} & PH & L & C & NC\\
    Cnn-Inte & \citeauthor{liu2018interpretable} & \cite{liu2018interpretable} & \citeyear{liu2018interpretable} & PH & G & C & P\\
    Compositionality & \citeauthor{li2016visualizing} & \cite{li2016visualizing} & \citeyear{li2016visualizing} & PH & L & C & T\\
    Data-flow graphs & \citeauthor{wongsuphasawat2018visualizing} & \cite{wongsuphasawat2018visualizing} &  \citeyear{wongsuphasawat2018visualizing} & PH & G & C / R & P; NC; T\\
    Deep Learning Important FeaTures (DeepLIFT) & \citeauthor{shrikumar2017learning} & \cite{shrikumar2017learning} &  \citeyear{shrikumar2017learning} & PH & L & C & P; NC\\
    Deep View (DV) &  \citeauthor{zhong2017evolutionary} & \cite{zhong2017evolutionary} & \citeyear{zhong2017evolutionary} & PH & G & C / R & P\\
    Deep Visualization Toolbox & \citeauthor{yosinski2015understanding} & \cite{yosinski2015understanding} &  \citeyear{yosinski2015understanding} & PH & G & C & P\\
    Deep-Taylor Decomposition & \citeauthor{montavon2017explaining} & \cite{montavon2017explaining} & \citeyear{montavon2017explaining} & PH & G & C & P\\
    DeepResolve & \citeauthor{liu2017visualizing} & \cite{liu2017visualizing} & \citeyear{liu2017visualizing} & PH & G & C & NC\\
    Explanatory Graph &  \citeauthor{zhang2018interpreting} &  \cite{zhang2018interpreting} &  \citeyear{zhang2018interpreting} & PH & G & C & P\\
    Feature maps &  \citeauthor{zhang2018interpretable} &  \cite{zhang2018interpretable} &  \citeyear{zhang2018interpretable} & AH & L & C & P\\
    Generative Adversarial Network (GAN) Dissection &  \citeauthor{bau2019gandissect} &  \cite{bau2019gandissect} &  \citeyear{bau2019gandissect} & PH & L & C & P\\
    GradCam &  \citeauthor{selvaraju2017grad} & \cite{selvaraju2017grad} & \citeyear{selvaraju2017grad} & PH & L & C & P\\
    Guided BackProp and Occlusion &  \citeauthor{goyal2016towards} &  \cite{goyal2016towards} &  \citeyear{goyal2016towards} & PH & L & C & P\\
    Guided Feature Inversion &  \citeauthor{du2018towards} &  \cite{du2018towards} &  \citeyear{du2018towards}& PH & L & C & P\\
    Hidden Activity Visualization &  \citeauthor{rauber2017visualizing} &  \cite{rauber2017visualizing} &  \citeyear{rauber2017visualizing} & PH & G & C & P\\
    Important Neurons and Patches & \citeauthor{lengerich2017towards} &  \cite{lengerich2017towards} &  \citeyear{lengerich2017towards} & PH & G & C & P\\
    iNNvestigate &  \citeauthor{alber2019innvestigate} &  \cite{alber2019innvestigate} & \citeyear{alber2019innvestigate} & PH & L & C & P\\
    Integrated Gradients &  \citeauthor{sundararajan2017axiomatic} &  \cite{sundararajan2017axiomatic} &  \citeyear{sundararajan2017axiomatic} & PH & L & C & P\\
    Inverting Representations  &  \citeauthor{mahendran2015understanding} & \cite{mahendran2015understanding} &  \citeyear{mahendran2015understanding} & PH & L & C & P\\
    LRP w/ Relevance Conservation & \citeauthor{arras2017explaining} & \cite{arras2017explaining} &  \citeyear{arras2017explaining} & PH & L & C & T\\
    LRP w/ Local Renormalization Layers &  \citeauthor{binder2016layer} & \cite{binder2016layer} &  \citeyear{binder2016layer} & PH & L & C & P\\
    LSTMVis &  \citeauthor{strobelt2018lstmvis} & \cite{strobelt2018lstmvis} & \citeyear{strobelt2018lstmvis} & PH & G / L & C & T\\
    N$^2$VIS & \citeauthor{streeter2001nvis} & \cite{streeter2001nvis} & \citeyear{streeter2001nvis} & PH & G & C / R & NC\\
    Net2Vec &  \citeauthor{fong2018net2vec} & \cite{fong2018net2vec} & \citeyear{fong2018net2vec} & PH & G & C & P\\
    Neural Network and CBR Twin-systems & \citeauthor{kenny2019twin} & \cite{kenny2019twin} &  \citeyear{kenny2019twin} & PH & L & C & P\\
    OcclusionSensitivity &  \citeauthor{zeiler2014visualizing} &  \cite{zeiler2014visualizing} &  \citeyear{zeiler2014visualizing} & PH & G & C & P\\
    OpenBox &  \citeauthor{chu2018exact} &  \cite{chu2018exact} &  \citeyear{chu2018exact} & PH & G & C & P; NC\\
    PatternNet, PatternAttribution &  \citeauthor{kindermans2018learning} &  \cite{kindermans2018learning} &  \citeyear{kindermans2018learning} & PH & L & C & P\\
    Prediction Difference Analysis &  \citeauthor{zintgraf2017visualizing} &  \cite{zintgraf2017visualizing} &  \citeyear{zintgraf2017visualizing} & PH & L & C & P\\
    Principal Component Analysis & \citeauthor{aubry2015understanding} &  \cite{aubry2015understanding} &  \citeyear{aubry2015understanding} & PH & G & C & P\\
    Receptive Fields &  \citeauthor{he2017deep} & \cite{he2017deep} &  \citeyear{he2017deep} & PH & G & C & P\\
    Relevant Features Selection &  \citeauthor{mogrovejo2019visual} &  \cite{mogrovejo2019visual} &  \citeyear{mogrovejo2019visual} & PH & L & C & P\\
    Saliency maps &  \citeauthor{olah2018building} & \cite{olah2018building} &  \citeyear{olah2018building} & PH & G / L & C & P\\
    Saliency maps &  \citeauthor{simonyan2014deep} & \cite{simonyan2014deep} &  \citeyear{simonyan2014deep} & PH & L & C & P\\
    Seq2seq-Vis &  \citeauthor{strobelt2018s} &  \cite{strobelt2018s} &  \citeyear{strobelt2018s} & PH & L & C & T\\
    SmoothGrad &  \citeauthor{smilkov2017smoothgrad} & \cite{smilkov2017smoothgrad} &  \citeyear{smilkov2017smoothgrad} & PH & L & C & P\\
    Stacking w/ Auxiliary Features (SWAF) & \citeauthor{rajani2017using} &  \cite{rajani2017using} &  \citeyear{rajani2017using} & PH & L & C & P\\
    Symbolic Graph Reasoning (SGR) & \citeauthor{liang2018symbolic} &  \cite{liang2018symbolic} &  \citeyear{liang2018symbolic} & AH & G & C / R & P\\
    t-SNE maps & \citeauthor{zahavy2016graying} & \cite{zahavy2016graying} &  \citeyear{zahavy2016graying} & PH & G & C & NC\\
    TreeView & \citeauthor{thiagarajan2016treeview} & \cite{thiagarajan2016treeview} & \citeyear{thiagarajan2016treeview} & PH & G & C & P\\
    \hline
\end{tabular}
\end{table}

\begin{table}[htbp]
\footnotesize
    \caption{Methods for explainability for neural networks generating rule-based explanations, classified according to the stage (AH: Ante-hoc, PH: Post-hoc), type of problems (C: Classification, R: Regression), scope (G: Global, L: Local) and input data (NC: Numerical/Categorical, P: Pictorials, T: Textual, TS: Time Series) of the underlying model.}
    \label{tab:neural-networks-rule}
    \begin{tabular}{m{4.2cm} m{2cm} m{0.5cm} m{0.5cm} m{0.5cm} m{0.6cm} m{0.8cm} m{1.1cm}}
    \hline
    Method for explainability & Authors & Ref & Year & Stage & Scope & Problem & Input\\
    \hline
    C4.5Rule-PANE &  \citeauthor{zhou2003medical} &  \cite{zhou2003medical} &  \citeyear{zhou2003medical} & PH & L & C / R & NC\\
    DecText &  \citeauthor{boz2002extracting} &  \cite{boz2002extracting} &  \citeyear{boz2002extracting} & PH & G & C & NC\\
    Discretized Interpretable Multi Layer Perceptrons (DIMLP) &  \citeauthor{bologna2017characterization,bologna2018rule} &  \cite{bologna2017characterization,bologna1998symbolic,bologna2018rule, bologna2018comparison} &  \citeyear{bologna2017characterization,bologna1998symbolic,bologna2018rule} & PH & G / L & C & P; NC; T\\
    Discretizing Hidden Unit Activation Values by Clustering &  \citeauthor{setiono1995understanding} &  \cite{setiono1995understanding} &  \citeyear{setiono1995understanding} & PH & G & C & NC\\
    DT extraction &  \citeauthor{frosst2017distilling,zhang2019interpreting} &  \cite{frosst2017distilling,zhang2019interpreting} &  \citeyear{frosst2017distilling,zhang2019interpreting} & PH & G & C & P\\
    Interval Propagation &  \citeauthor{palade2001interpretation} &  \cite{palade2001interpretation} &  \citeyear{palade2001interpretation} & PH & G & C & NC\\
    Iterative Rule Knowledge Distillation &  \citeauthor{hu2016harnessing} &  \cite{hu2016harnessing} &  \citeyear{hu2016harnessing} & AH & G & C & T\\
    Neural Network Knowledge eXtraction (NNKX) &  \citeauthor{bondarenko2017classification} &  \cite{bondarenko2017classification} &  \citeyear{bondarenko2017classification} & PH & G & C & NC\\
    Rule Extraction From Neural network Ensemble (REFNE) &  \citeauthor{zhou2003extracting} &  \cite{zhou2003extracting} &  \citeyear{zhou2003extracting} & PH & G & C / R & NC\\
    Rule Extraction from Neural Network using Classified and Misclassified data (RxNCM) &  \citeauthor{biswas2017rule} &  \cite{biswas2017rule} &  \citeyear{biswas2017rule} & PH & G & C & NC\\
    Rule Extraction by Reverse Engineering (RxREN) &  \citeauthor{augasta2012reverse} &  \cite{augasta2012reverse} &  \citeyear{augasta2012reverse} & AH & G & C / R & NC\\
    Symbolic logic integration &  \citeauthor{tran2017unsupervised} &  \cite{tran2017unsupervised} &  \citeyear{tran2017unsupervised} & AH & G & C / R & NC\\
    Symbolic rules &  \citeauthor{garcez2001symbolic} &  \cite{garcez2001symbolic} &  \citeyear{garcez2001symbolic} & PH & G & C / R & NC\\
    Tree Regularization &  \citeauthor{wu2018beyond} &  \cite{wu2018beyond} &  \citeyear{wu2018beyond} & AH & G & C & NC\\
    Trepan &  \citeauthor{craven1994using, craven1996extracting} &  \cite{craven1994using, craven1996extracting} &  \citeyear{craven1994using, craven1996extracting} & PH & G & C / R & NC\\
    Validity Interval Analysis (VIA) &  \citeauthor{thrun1995extracting} &  \cite{thrun1995extracting} &  \citeyear{thrun1995extracting} & PH & G & C / R & TS\\
    Word Importance Scores &  \citeauthor{murdoch2017automatic} &  \cite{murdoch2017automatic} &  \citeyear{murdoch2017automatic} & PH & G & C & T\\
    \hline
\end{tabular}
\end{table}

\begin{table}[h!]
\footnotesize
    \caption{Methods for explainability for neural networks generating textual and numerical explanations, classified according to the output format (N: Numerical, T: Textual), stage (AH: Ante-hoc, PH: Post-hoc), type of problems (C: Classification, R: Regression), scope (G: Global, L: Local) and input data (NC: Numerical/Categorical, P: Pictorials, T: Textual, TS: Time Series) of the underlying model.}
    \label{tab:neural-networks-textual-numeric}
    \begin{tabular}{m{3.8cm} m{2.1cm} m{0.5cm} m{0.5cm} m{0.85cm} m{0.5cm} m{0.6cm} m{0.7cm} m{0.5cm}}
    \hline
    Method for explainability & Authors & Ref & Year & Output Format & Stage & Scope & Problem & Input\\
    \hline
    Causal Importance &  \citeauthor{feraud2002methodology} &  \cite{feraud2002methodology} &  \citeyear{feraud2002methodology} & N & PH & G & C & NC\\
    Concept Activation Vectors &  \citeauthor{kim2018interpretability} &  \cite{kim2018interpretability} &  \citeyear{kim2018interpretability} & N & PH & G & C & P\\
    Contextual Importance, Utility &  \citeauthor{framling1996explaining} &  \cite{framling1996explaining} &  \citeyear{framling1996explaining} & N & PH & G / L & C & NC\\
    InterpNET &  \citeauthor{barratt2017interpnet} &  \cite{barratt2017interpnet} &  \citeyear{barratt2017interpnet} & T & PH & L & C & P\\
    Most-Weighted-Path, Most-Weighted-Combination, Maximum-Frequency-Difference &  \citeauthor{garcia2019human} &  \cite{garcia2019human} &  \citeyear{garcia2019human} & T & PH & L & C & TS\\
    Probes &  \citeauthor{alain2017understanding} &  \cite{alain2017understanding} &  \citeyear{alain2017understanding} & N & PH & G & C & P\\
    Rationales &  \citeauthor{lei2016rationalizing} &  \cite{lei2016rationalizing} &  \citeyear{lei2016rationalizing} & T & PH & L & C & T\\
    REcurrent LEXicon NETwork (RELEXNET) &  \citeauthor{clos2017towards} &  \cite{clos2017towards} &  \citeyear{clos2017towards} & N & AH & G & C & T\\
    Relevance, Discriminative Loss &  \citeauthor{hendricks2016generating} &  \cite{hendricks2016generating,hendricks2018grounding} &  \citeyear{hendricks2016generating,hendricks2018grounding} & T & PH & L & C & P\\
    Singular Vector Canonical Correlation Analysis (SVCCA) &  \citeauthor{raghu2017svcca} &  \cite{raghu2017svcca} &  \citeyear{raghu2017svcca} & N & PH & G & C / R & P\\
    \hline
\end{tabular}
\end{table}

\begin{table}[h!]
\footnotesize
    \caption{Methods for explainability for neural networks generating mixed explanations, classified according to the stage (AH: Ante-hoc, PH: Post-hoc), type of problems (C: Classification, R: Regression), scope (G: Global, L: Local) and input data (NC: Numerical/Categorical, P: Pictorials, T: Textual, TS: Time Series) of the underlying model.}
    \label{tab:neural-networks-mixed}
    \begin{tabular}{m{4.5cm} m{2cm} m{0.5cm} m{0.5cm} m{0.5cm} m{0.6cm} m{0.8cm} m{0.8cm}}
    \hline
    Method for explainability & Authors & Ref & Year & Stage & Scope & Problem & Input\\
    \hline
    Activation values of hidden neurons &  \citeauthor{tamajka2019transforming} &  \cite{tamajka2019transforming} &  \citeyear{tamajka2019transforming} & AH & L & C & P\\
    Attention Alignment &  \citeauthor{kim2018textual} &  \cite{kim2018textual} &  \citeyear{kim2018textual} & PH & L & C & P\\
    Deterministic Finite Automaton &  \citeauthor{mayr2018regular} &  \cite{mayr2018regular} &  \citeyear{mayr2018regular} & PH & L & C & NC\\
    Deterministic Finite-state Automata (DFAs) &  \citeauthor{omlin1996extraction} &  \cite{omlin1996extraction} &  \citeyear{omlin1996extraction} & PH & G & C & NC\\
    Image Caption Generation with Attention Mechanism &  \citeauthor{xuk2015show} &  \cite{xuk2015show} &  \citeyear{xuk2015show} & PH & L & C & P\\
    Pointing and Justification Model (PJ-X) &  \citeauthor{park2018multimodal} &  \cite{park2018multimodal} &  \citeyear{park2018multimodal} & AH & L & C & P\\
    \hline
\end{tabular}
\end{table}

\begin{table}[htbp]
\footnotesize
    \caption{Methods for explainability for rule-based construction approaches, classified according to the output format (M: Mixed, N: Numerical, R: Rules, T: Textual, N: Numerical), stage (AH: Ante-hoc, PH: Post-hoc), type of problems (C: Classification, R: Regression), scope (G: Global, L: Local) and input data (NC: Numerical/Categorical, P: Pictorials, T: Textual, TS: Time Series) of the underlying model.}
    \label{tab:rule-based}
    \begin{tabular}{m{3.5cm} m{2cm} m{0.5cm} m{0.5cm} m{0.85cm} m{0.5cm} m{0.6cm} m{0.8cm} m{0.6cm}}
    \hline
    Method for explainability & Authors & Ref & Year & Output Format & Stage & Scope & Problem & Input\\
    \hline
    Ant Colony Optimization (ACO) &  \citeauthor{otero2016improving} &  \cite{otero2016improving} &  \citeyear{otero2016improving} & R & AH & G & C & NC\\
    AntMiner+ and ALBA &  \citeauthor{verbeke2011building} &  \cite{verbeke2011building} &  \citeyear{verbeke2011building} & R & AH & G & C & NC\\
    Argumentation &  \citeauthor{rizzo2019inferential} &  \cite{rizzo2019inferential,rizzo2018qualitative} &  \citeyear{rizzo2019inferential} & R & AH & G & C & NC\\
    Argumentation &  \citeauthor{zeng2018building} &  \cite{zeng2018building} &  \citeyear{zeng2018building} & R & AH & G & C / R & P\\
    Bayesian Rule Lists (BRL) &  \citeauthor{letham2012building} &  \cite{letham2012building,letham2013interpretable,letham2015interpretable} &  \citeyear{letham2012building,letham2013interpretable,letham2015interpretable} & R & AH & G & C & NC\\
    Bayesian Rule Sets (BRS) &  \citeauthor{wang2016bayesian} &  \cite{wang2016bayesian,wang2017bayesian} &  \citeyear{wang2016bayesian,wang2017bayesian} & R & AH & G & C & NC\\
    Interpretable Decision Set &  \citeauthor{lakkaraju2016interpretable} &  \cite{lakkaraju2016interpretable} &  \citeyear{lakkaraju2016interpretable} & R & AH & G & C & NC\\
    Exception Directed Acyclic Graphs (EDAGs) &  \citeauthor{gaines1996transforming} &  \cite{gaines1996transforming} &  \citeyear{gaines1996transforming} & M & AH & G & C & NC\\
    ExpliClas &  \citeauthor{alonso2019explainable} &  \cite{alonso2019explainable} &  \citeyear{alonso2019explainable} & M & PH & L & C & NC\\
    First Order Combined Learner (FOCL) &  \citeauthor{pazzani1997comprehensible} &  \cite{pazzani1997comprehensible} &  \citeyear{pazzani1997comprehensible} & R & AH & G & C & NC\\
    Fuzzy logic &  \citeauthor{pierrard2018learning} &  \cite{pierrard2018learning} &  \citeyear{pierrard2018learning} & R & AH & L & C & NC\\
    Fuzzy system &  \citeauthor{jin2000fuzzy} &  \cite{jin2000fuzzy} &  \citeyear{jin2000fuzzy} & R & AH & G & C & NC\\
    Fuzzy Inference-Grams (Fingrams) &  \citeauthor{pancho2013fingrams} &  \cite{pancho2013fingrams} &  \citeyear{pancho2013fingrams} & V & PH & G & C & NC\\
    Fuzzy Inference Systems &  \citeauthor{keneni2019evolving} &  \cite{keneni2019evolving} &  \citeyear{keneni2019evolving} & T & PH & L & C & TS\\
    Genetics-Based Machine Learning (GBML) &  \citeauthor{ishibuchi2007analysis} &  \cite{ishibuchi2007analysis} &  \citeyear{ishibuchi2007analysis} & R & AH & G & C & NC\\
    Interpretable Classification Rule Mining (ICRM) &  \citeauthor{cano2013interpretable} &  \cite{cano2013interpretable} &  \citeyear{cano2013interpretable} & R & AH & G & C & NC\\
    Linear Programming Relaxation &  \citeauthor{malioutov2017learning, su2016interpretable} &  \cite{malioutov2017learning, su2016interpretable} &  \citeyear{malioutov2017learning, su2016interpretable} & R & AH & G & C & NC\\
    Multi-Objective Evolutionary Algorithms based Interpretable Fuzzy (MOEAIF) &  \citeauthor{wang2011building} &  \cite{wang2011building} &  \citeyear{wang2011building} & R & AH & G & C & NC\\
    Mycin &  \citeauthor{shortliffe1975computer} &  \cite{shortliffe1975computer} &  \citeyear{shortliffe1975computer} & T & PH & L & C & NC\\
    \hline
\end{tabular}
\end{table}

\begin{table}[htbp]
\footnotesize
    \caption{Methods for explainability for data-driven approaches, classified according to the output format (M: Mixed, N: Numerical, R: Rules, T: Textual, N: Numerical), stage (AH: Ante-hoc, PH: Post-hoc), type of problems (C: Classification, R: Regression), scope (G: Global, L: Local) and input data (NC: Numerical/Categorical, P: Pictorial, T: Textual, TS: Time Series).}
    \label{tab:data-driven}
    \begin{tabular}{m{2.3cm} m{1.4cm} m{0.5cm} m{0.5cm} m{1.5cm} m{0.85cm} m{0.5cm} m{0.6cm} m{0.8cm} m{0.8cm}}
    \hline
    Method for \newline explainability & Authors & Ref & Year & Construction approach & Output Format & Stage & Scope & Problem & Input\\
    \hline
    Contribution Propagation &  \citeauthor{Landecker2013interpreting} &  \cite{Landecker2013interpreting} &  \citeyear{Landecker2013interpreting} & Hierarchical networks & V & PH & L & C & P\\
    DT extraction &  \citeauthor{andrzejak2013interpretable} &  \cite{andrzejak2013interpretable} &  \citeyear{andrzejak2013interpretable} & Distributed DTs & R & PH & G & C / R & NC\\
    DT extraction &  \citeauthor{ferri2002ensemble, van2007seeing} &  \cite{ferri2002ensemble,van2007seeing} &  \citeyear{ferri2002ensemble,van2007seeing} & Ensembles & R & PH & G & C & NC\\
    DT extraction &  \citeauthor{alonso2018explainable} &  \cite{alonso2018explainable} &  \citeyear{alonso2018explainable} & Ensembles & T & PH & L & C & NC\\
    Discriminative Patterns &  \citeauthor{gao2017interpretable} &  \cite{gao2017interpretable} &  \citeyear{gao2017interpretable}  & Ensembles & T & PH & G & C & T\\
    Explaining Bayesian network Inferences (EBI) &  \citeauthor{yap2008explaining} &  \cite{yap2008explaining} &  \citeyear{yap2008explaining} & Bayesian networks & R & PH & G & C & NC\\
    ExtractRule &  \citeauthor{fung2005rule} &  \cite{fung2005rule} &  \citeyear{fung2005rule} & Hyperplane-Based Linear Classifiers & R & PH & G & C & P; NC\\
    Factorized Asymptotic Bayesian (FAB) inference  &  \citeauthor{hara2018making} &  \cite{hara2018making} &  \citeyear{hara2018making}  & Ensembles & R & PH & G & C & NC\\
    Feature Tweaking &  \citeauthor{tolomei2017interpretable} &  \cite{tolomei2017interpretable} &  \citeyear{tolomei2017interpretable} & Ensembles & N & PH & L & C & NC\\
    Important Support Vectors and Border Classification &  \citeauthor{barbella2009understanding} &  \cite{barbella2009understanding} &  \citeyear{barbella2009understanding} & SVM & N & PH & L & C & NC\\
    inTrees &  \citeauthor{deng2018interpreting} &  \cite{deng2018interpreting} &  \citeyear{deng2018interpreting} &  Ensembles & R & PH & G & C / R & NC\\
    Nomograms &  \citeauthor{jakulin2005nomograms} &  \cite{jakulin2005nomograms} &  \citeyear{jakulin2005nomograms} & SVM & V & PH & G & C & NC\\
    Nomograms &  \citeauthor{movzina2004nomograms} &  \cite{movzina2004nomograms} &  \citeyear{movzina2004nomograms} & Na\"{i}ve Bayes & V & PH & G & C & NC\\
    Probabilistically Supported Arguments &  \citeauthor{timmer2017two} &  \cite{timmer2017two} &  \citeyear{timmer2017two} & Bayesian networks & M & PH & G & C & NC\\
    Scenarios &  \citeauthor{vlek2016method} &  \cite{vlek2016method} &  \citeyear{vlek2016method} & Bayesian networks & T & PH & L & C & NC\\
    Self-Organizing Maps &  \citeauthor{hamel2006visualization} &  \cite{hamel2006visualization} &  \citeyear{hamel2006visualization} & SVM & V & PH & G & C & NC\\
    SVM+Prototypes &  \citeauthor{nunez2002rule} &  \cite{nunez2002rule} &  \citeyear{nunez2002rule} & SVM & M & PH & G & C & NC\\
    Tree Space Prototypes &  \citeauthor{tan2016tree} &  \cite{tan2016tree} &  \citeyear{tan2016tree} & Ensembles & M & PH & L & C & NC\\
    Visualization for RIsk Factor Analysis (VRIFA) &  \citeauthor{cho2008nonlinear} &  \cite{cho2008nonlinear} &  \citeyear{cho2008nonlinear} & SVM & V & PH & G & C & NC\\
    Weighted Linear Classifier &  \citeauthor{caragea2003towards} &  \cite{caragea2003towards} &  \citeyear{caragea2003towards} & SVM & N & PH & G & C & NC\\
    \hline
\end{tabular}
\end{table}

\begin{table}[htbp]
\footnotesize
    \caption{Methods for explainability generating white-box models, classified according to the output format (M: Mixed, N: Numerical, R: Rules, T: Textual, N: Numerical), stage (AH: Ante-hoc, PH: Post-hoc), type of problems (C: Classification, R: Regression), scope (G: Global, L: Local) and input data (NC: Numerical/Categorical, P: Pictorial, T: Textual, TS: Time Series) of the underlying model.}
    \label{tab:white-box}
    \begin{tabular}{m{2cm} m{1.5cm} m{0.5cm} m{0.5cm} m{1.5cm} m{0.85cm} m{0.5cm} m{0.6cm} m{0.8cm} m{0.7cm}}
    \hline
    Method for \newline explainability & Authors & Ref & Year & Construction \newline approach & Output Format & Stage & Scope & Problem & Input\\
    \hline
    Bayesian Case Model (BCM) &  \citeauthor{kim2014bayesian} &  \cite{kim2014bayesian} &  \citeyear{kim2014bayesian} & Bayesian case-based reasoning & M & AH & G & C & P; T\\
    Gaussian Process Regression (GPR) &  \citeauthor{caywood2017gaussian} &  \cite{caywood2017gaussian} &  \citeyear{caywood2017gaussian} & Gaussian Process Regression & N & AH & G & R & TS\\
    General Additive Models (GAMs) &  \citeauthor{lou2012intelligible,lou2013accurate} &  \cite{lou2012intelligible,lou2013accurate} &  \citeyear{lou2012intelligible} & Additive models & M & AH & G & C / R & NC\\
    GAMs with pairwise interactions (GA$^2$Ms) &  \citeauthor{caruana2015intelligible,lou2013accurate} &  \cite{caruana2015intelligible} &  \citeyear{caruana2015intelligible} & Additive models & M & AH & G & C / R & NC\\
    Mind the Gap Model (MGM) &  \citeauthor{kim2015mind} &  \cite{kim2015mind} &  \citeyear{kim2015mind} & Mind the Gap Model (MGM) & M & AH & G & C & P; NC; T\\
    Multi-Run Subtree Encapsulation &  \citeauthor{howard2018explainable} &  \cite{howard2018explainable} &  \citeyear{howard2018explainable} & Tree-based Genetic Programming & M & AH & G & C & NC\\
    Oblique Treed Sparse Additive Models (OT-SpAMs) &  \citeauthor{wang2015trading} &  \cite{wang2015trading} &  \citeyear{wang2015trading} & Additive models & N & AH & G & C & NC\\
    Probabilistic Sentential Decision Diagrams (PSDD) &  \citeauthor{liang2017towards} &  \cite{liang2017towards} &  \citeyear{liang2017towards} & Probabilistic sentential decision diagrams & R & AH & G & C / R & NC\\
    Supersparse Linear Integer Model (SLIM) &  \citeauthor{ustun2014supersparse} &  \cite{ustun2014supersparse} &  \citeyear{ustun2014supersparse} & Scoring systems & N & AH & G & C & NC\\
    Unsupervised Interpretable Word Sense Disambiguation &  \citeauthor{panchenko2017unsupervised} &  \cite{panchenko2017unsupervised} &  \citeyear{panchenko2017unsupervised} & UIWSD & V & AH & G & C & T\\
    Transparent Generalized Additive Model Tree (TGAMT) &  \citeauthor{fahner2018developing} &  \cite{fahner2018developing} &  \citeyear{fahner2018developing} & Additive models & R & AH & G & C & NC\\
    \hline
\end{tabular}
\end{table}

\begin{table}[htbp]
\footnotesize
  \caption{List and classification of the scientific articles proposing human-centered approaches to evaluate methods for explainability, classified according to the construction approach, the type of measurement employed (qualitative or quantitative), and the format of their output explanation.}
  \label{tab:human-evaluation}
  \begin{tabular}{m{1.8cm} m{5cm} m{2cm} m{3cm}}
    \hline
    Measure type & Construction approach & Output format & Reference\\
    \hline
    Qualitative & Autonomous agents & Textual & \cite{dzindolet2003role}\\
    Qualitative & Context-aware systems & Textual & \cite{lim2009assessing, lim2009and}\\
    Qualitative & Context-aware systems & Visual & \cite{lim2009assessing}\\
    Qualitative & Data clustering & Mixed &  \cite{kim2015mind}\\
    Qualitative & Expert Systems & Textual & \cite{suermondt1993evaluation, ye1995impact}\\
    Qualitative & Learning Systems & Textual & \cite{putnam2019exploring}\\
    Qualitative & ML (Model agnostic) & Visual & \cite{krause2016interacting, tullio2007works}\\
    Qualitative & Fisher vector classifiers & Visual & \cite{lapuschkin2016analyzing}\\
    Qualitative & Support Vector Machine & Visual & \cite{srinivasan2017interpretable}\\
    Qualitative & Neural networks & Visual & \cite{assaf2019explainable, erhan2009visualizing, ding2017visualizing,sturm2016interpretable}\\
    Quantitative & Case-based Reasoning & Mixed & \cite{kim2014bayesian}\\
    Quantitative & Context-aware systems & Textual & \cite{lim2009and}\\
    Quantitative & Neural networks & Visual & \cite{stock2018convnets} \\
    Quantitative & Neural networks & Visual & \cite{bau2017network} \\
    Quantitative & Decision trees & Mixed & \cite{luvstrek2014comprehensibility}\\
    Quantitative & Kernel-based neural networks & Visual & \cite{hansen2011visual} \\
    Quantitative & Learning Systems & Textual &  \cite{aleven2002effective,harbers2010guidelines, harbers2010design}\\
    Quantitative & ML (Model agnostic) & Mixed & \cite{lage2018human}\\
    Quantitative & ML (Model agnostic) & Rules & \cite{ribeiro2018anchors}\\
    Quantitative & ML (Model agnostic) & Mixed &  \cite{poursabzi2018manipulating}\\
    Quantitative & ML (Model agnostic) & Visual & \cite{spinner2019explainer}\\
    Quantitative & Na\"{i}ve Bayes & Textual & \cite{kulesza2011oriented, kulesza2015principles}\\
    Quantitative & Rule-based classifier & Mixed & \cite{allahyari2011user, huysmans2011empirical}\\
    Quantitative & Decision trees & Mixed & \cite{allahyari2011user, huysmans2011empirical}\\
    Quantitative & Decision tables & Mixed & \cite{huysmans2011empirical}\\
  \hline
\end{tabular}
\end{table}

\begin{table}[htbp]
\small
  \caption{Classification of the scientific articles proposing comparative approaches to evaluate methods for explainability, classified according to the methodology followed to carry out the comparison task.}
  \label{tab:comparison-methods}
  \begin{tabular}{m{6cm} m{6cm}}
    \hline
    Comparison approach & Reference\\
    \hline
    Sensitivity to input perturbation &\cite{adebayo2018local, adebayo2018sanity, alvarez2018robustness, ancona2018towards, arras2016explaining, binder2016analyzing, ghorbani2017interpretation, kindermans2017reliability, samek2017explainable, samek2017evaluating} \\
    Sensitivity to model parameter randomization & \cite{adebayo2018local, adebayo2018sanity, erhan2009visualizing}\\
    Explanation completeness & \cite{gevrey2003review}\\
  \hline
\end{tabular}
\end{table}

\begin{table}[htbp]
\small
  \caption{List of the methods for explainability generating visual explanations, such as heat-maps, whose degree of explainability is evaluated by comparison (listed in the fourth column). These comparisons were carried out over different types of input data (listed in the third column).}
  \label{tab:visual-comparison}
  \begin{tabular}{m{4.9cm} m{1cm} m{1.5cm} m{4.4cm}}
    \hline
    Method for explainability (references) & Acronym & Input type & Compared with (references)\\
    \hline
    \hline
    Deep-Taylor Decomposition \cite{montavon2017explaining} & DTD & Pictorial & GBP, IG, SA, PM in \cite{kindermans2017reliability}\\
    \hline
    DeepLIFT \cite{shrikumar2017learning} & DLT & Pictorial & IG, LRP in \cite{ancona2018towards}\newline IG, SA in \cite{ghorbani2017interpretation}\\
    \hline
    Gradient*Input \cite{shrikumar2017learning} & GI & Pictorial & GC, GBP and SG \cite{adebayo2018local} \newline GC, GBP, IG, SG \cite{adebayo2018sanity} \newline IG, LIME, OS, SM, SHAP in \cite{alvarez2018robustness}\\
    \hline
    GrandCam \cite{selvaraju2017grad} & GC & Pictorial & GI, GBP, SG \cite{adebayo2018local} \newline GI, GBP, IG, SG \cite{adebayo2018sanity}\\
    \hline
    Guided BackProp \cite{goyal2016towards} & GBP & Pictorial & GI, GC, SG \cite{adebayo2018local} \newline GI, GC, IG, SG \cite{adebayo2018sanity} \newline DTD, IG, SA, PM in \cite{kindermans2017reliability} \\
    \hline
    Integrated Gradients \cite{sundararajan2017axiomatic} & IG & Pictorial & GI, GC, GBP, SG \cite{adebayo2018sanity} \newline LRP, LIME, OS, SM, SHAP \cite{alvarez2018robustness} \newline DLT and LRP in \cite{ancona2018towards} \newline DLT, SA in \cite{ghorbani2017interpretation} \newline DTD, GBP, SA, PM in \cite{kindermans2017reliability} \\
    \hline
    Layer-wise Relevance Propagation \cite{bach2015pixel} & LRP & Pictorial & IG, LIME, OS, SM, SHAP in  \cite{alvarez2018robustness} \newline DLT, IG in \cite{ancona2018towards} \newline SA in \cite{binder2016analyzing, samek2017evaluating, samek2017explainable} \newline OS in \cite{binder2016analyzing}\\
    \hline
    Local Interpretable Model-Agnostic Explanations \cite{ribeiro2016should} & LIME & Pictorial \newline Numerical / Categorical & GI, IG, OS, SM, SHAP in \cite{alvarez2018robustness}\\
    \hline
    Occlusion Sensitivity \cite{zeiler2014visualizing} & OS & Pictorial & GI, IG, LIME, SM, SHAP in \cite{alvarez2018robustness} \newline LRP in \cite{binder2016analyzing} \\
    \hline
    Sensitivity Analysis \cite{baehrens2010explain} & SA & Pictorial & LRP in \cite{binder2016analyzing} \newline DLT, IG in \cite{ghorbani2017interpretation} \newline DTD, GBP, IG, PM in \cite{kindermans2017reliability} \newline LRP in \cite{samek2017evaluating, samek2017explainable}\\
    \hline
    PatternNet \cite{kindermans2018learning} & PM & Pictorial & DTD, GBP, IG, SA in \cite{kindermans2017reliability} \\
    \hline
    Saliency Maps \cite{simonyan2014deep} & SM & Pictorial & GI, IG, LIME, OS, SHAP in \cite{alvarez2018robustness}\\
    \hline
    SHapley Additive exPlanations \cite{lundberg2017unified} & SHAP & Pictorial & GI, IG, LIME, OS, SM in \cite{alvarez2018robustness}\\
    \hline
    SmoothGrad \cite{smilkov2017smoothgrad} & SG & Pictorial & GI, GC, GBP \cite{adebayo2018local} \newline GI, GC, GBP and IG \cite{adebayo2018sanity}\\
    \hline
    Layer-wise Relevance Propagation  \cite{bach2015pixel} & LRP & Textual & SA in \cite{arras2016explaining,samek2017explainable} \\
    \hline
    Sensitivity Analysis methods \cite{baehrens2010explain} & SA & Textual & LRP in \cite{arras2016explaining,samek2017explainable} \\
  \hline
\end{tabular}
\end{table}

\end{document}